\documentclass[10pt,twocolumn,letterpaper]{article}

\usepackage{cvpr}      

\usepackage[dvipsnames]{xcolor}

\definecolor{cvprblue}{rgb}{0.21,0.49,0.74}
\usepackage[pagebackref,breaklinks,colorlinks,citecolor=cvprblue]{hyperref}
\usepackage{multirow}
\usepackage{lipsum}

\def\paperID{11063} 
\def\confName{CVPR}
\def\confYear{2024}

\title{
Towards a Simultaneous and Granular Identity-Expression Control in Personalized Face Generation
}

\author{Renshuai Liu$^{1,2}$, Bowen Ma$^2$, Wei Zhang$^2$,\\
Zhipeng Hu$^2$, Changjie Fan$^2$, Tangjie Lv$^2$, Yu Ding$^2\ast$, Xuan Cheng$^1\ast$\\
$^1$School of Informatics, Xiamen University \\
$^2$Virtual Human Group, Netease Fuxi AI Lab\\}

\begin{document}
\twocolumn[{
\renewcommand\twocolumn[1][]{#1}
\maketitle
\begin{center}
    \captionsetup{type=figure}
    \includegraphics[width=0.98\textwidth]{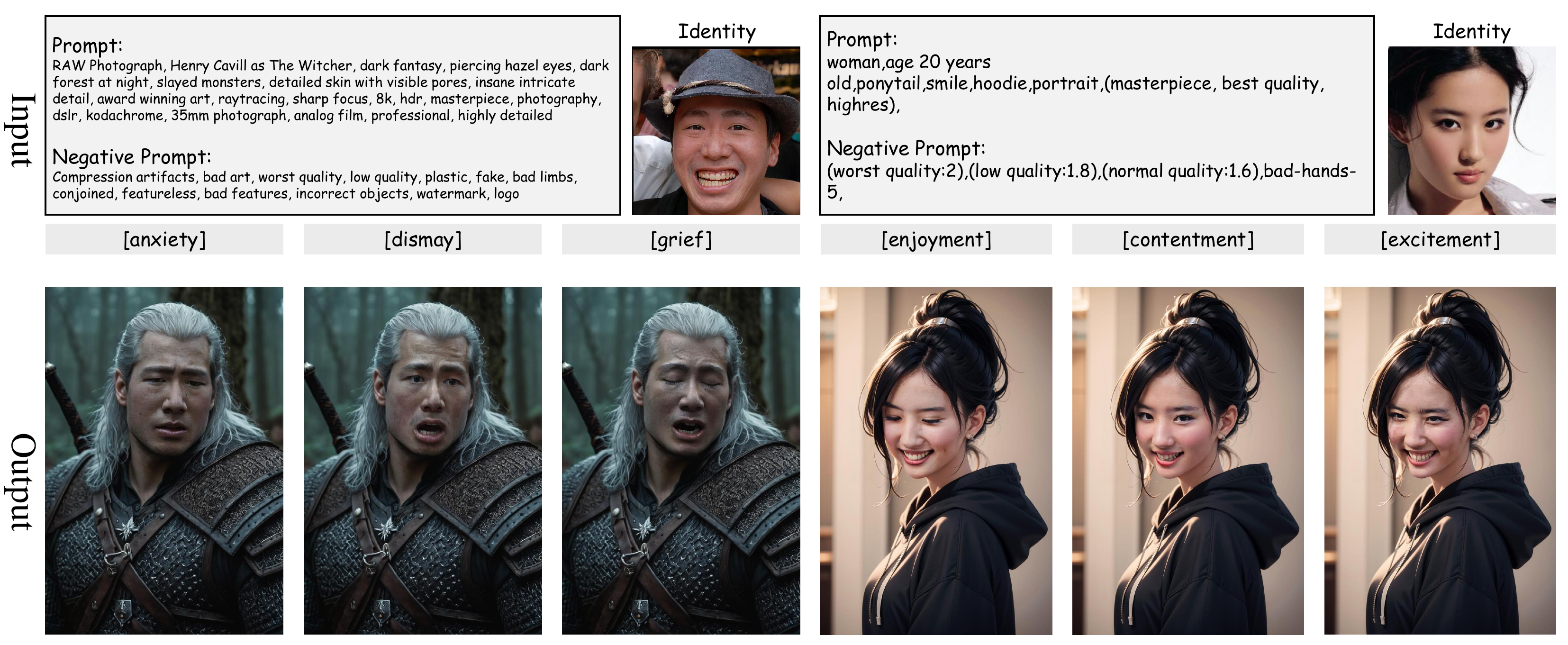}
    \captionof{figure}{The proposed framework takes three inputs: a prompt describing the background, a selfie photo uploaded by the user, and a text related to the fine-grained expression labels. The generated faces well match the inputted triples and exhibit fine-grained expression synthesis.}
    \label{fig:banner}
\end{center}
}]

\newcommand\blfootnote[1]{%
\begingroup
\renewcommand\thefootnote{}\footnote{#1}%
\addtocounter{footnote}{-1}%
\endgroup
}
\blfootnote{Work done when Renshuai Liu was an intern at Netease Fuxi AI Lab}
\blfootnote{$\ast$ Yu Ding and Xuan Cheng are co-Corresponding authors.}

\begin{abstract}
In human-centric content generation, the pre-trained text-to-image models struggle to produce user-wanted portrait images, which retain the identity of individuals while exhibiting diverse expressions. This paper introduces our efforts towards personalized face generation. To this end, we propose a novel multi-modal face generation framework, capable of simultaneous identity-expression control and more fine-grained expression synthesis. Our expression control is so sophisticated that it can be specialized by the fine-grained emotional vocabulary. We devise a novel diffusion model that can undertake the task of simultaneously face swapping and reenactment. Due to the entanglement of identity and expression, it's nontrivial to separately and precisely control them in one framework, thus has not been explored yet. To overcome this, we propose several innovative designs in the conditional diffusion model, including balancing identity and expression encoder, improved midpoint sampling, and explicitly background conditioning. Extensive experiments have demonstrated the controllability and scalability of the proposed framework, in comparison with state-of-the-art text-to-image, face swapping, and face reenactment methods.

\vspace{-8pt}
\end{abstract}

\section{Introduction}
\label{sec:intro}


The research community has been striving to improve controllability in the generation of facial images tailored to user preferences. A common practice in controllable generation and manipulation is to use different modalities as conditioning in a face generator model, such as texts \cite{TediGAN2021, StyleCLIP2021, StyleT2I2022, Text2FaceGAN2019, FacesLaCarte2021}, reference images \cite{SimSwap2020, FaceShifter2019, ReenactGAN2018, FSGAN2019}, segmentation masks \cite{EncodingInStyle2021, MaskGAN2020, SC-FEGAN2019} and audios \cite{Wave2lip2020, PCAVS2021, MakeltTalk2020}. 


Although these methods have realized the ability to control the local features and global attributes in a face, the simultaneous control of identity and expression in a specific background has not been fully explored, which involves three important high-level attributes (i.e. identity, expression, and background) to determine a face image. Since identity and expression are highly entangled, it's challenging to separately and precisely control them in a unified framework. Additionally, in existing generation or manipulation methods, the granularity of expression control remains at a coarse level, often limited to the commonly used seven or eight labels, e.g. ``surprise'', ``happiness'', ``anger'' etc. These labels struggle to cover the entire emotional space sufficiently in the open world.

\begin{figure}
\centering
\includegraphics[width=0.48\textwidth]{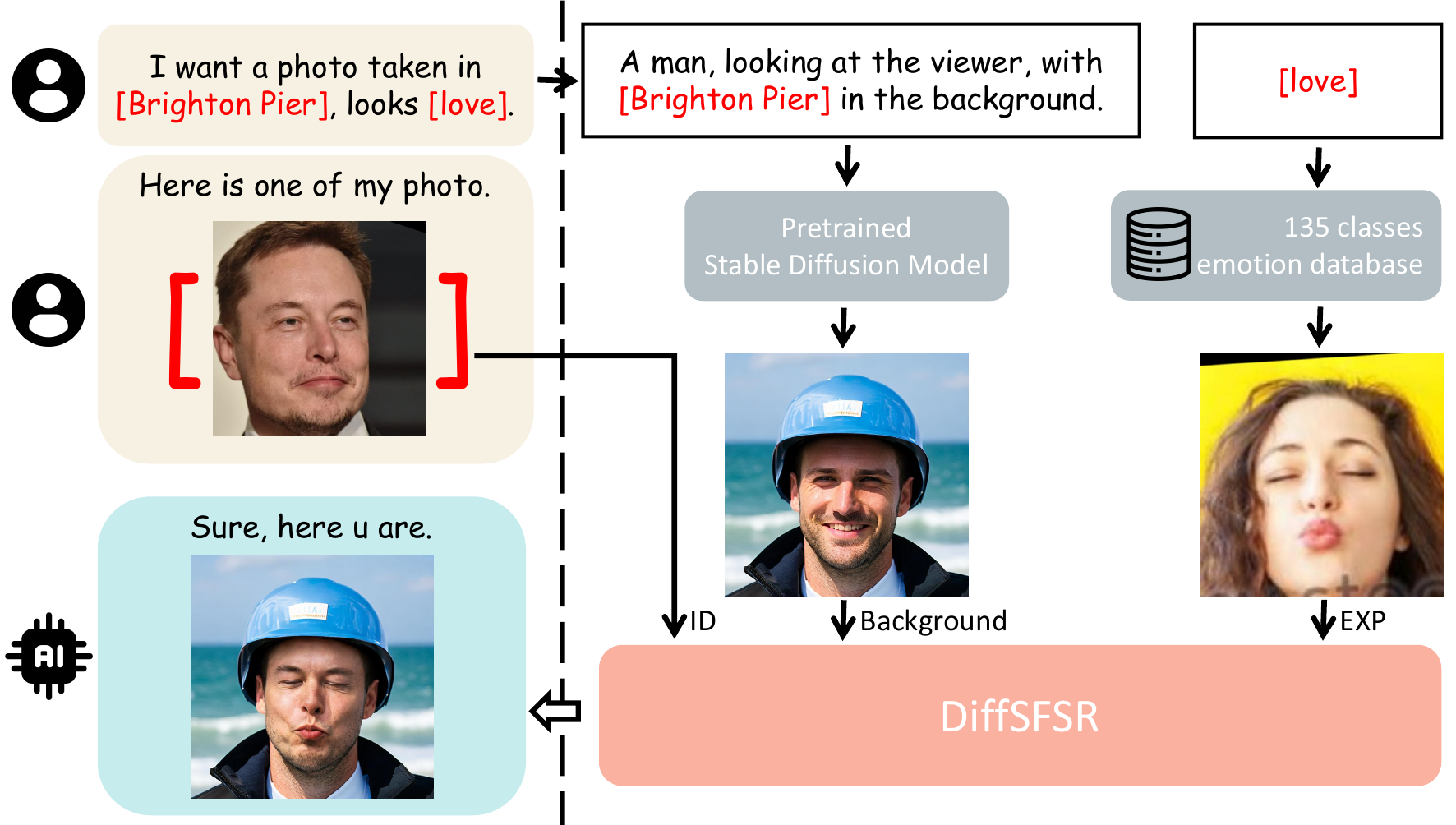}
\vspace{-15pt}
\caption{Overview of the proposed face generation framework.}
\vspace{-8pt}
\label{fig:T2I}
\end{figure}

To tackle these issues, this paper proposes a novel framework that can \emph{simultaneously control identity, expression, and background from multi-modal inputs}. As shown in Fig. \ref{fig:banner} and \ref{fig:T2I}, the inputs contain three items: 1) a text that describes the scene, and 2) a selfie photo uploaded by the user to provide identity, 3) a text related to the expression labels. Human language can conceptually describe expressions and accurately describe scenes but can't describe identity precisely, while images can be naturally used in identity recognition. On the output side, the generated face will have the same identity as the input selfie photo, show the expression specified by the text, and be placed in the background described in the text \cite{civitai1, civitai2}, as shown in Fig. \ref{fig:banner}. To support \emph{fine-grained expression description}, we employ an expression dictionary of 135 English words \cite{EmoFace135}, e.g. ``amazement'', ``exhilaration'', ``hysteria'' etc., which can more comprehensively describe the emotion domain.

The technical core inside the proposed framework is a novel diffusion model that can conduct \emph{Simultaneous Face Swapping and Reenactment (SFSR)}. Swapping and reenactment, which transfer the identity or expression of the source face to the target face, are two classical face manipulation tasks and have been studied extensively. Meanwhile, SFSR is a relatively new and unexplored task, which aims at separately transferring the identity from the source face, and the expression from another source to one target face, while keeping the background attributes (e.g. face pose, hair, glasses, and surroundings) in the target unchanged. To prepare the two sources and one target for SFSR, the text that describes the scene will be input to a pre-trained text-to-image model (Stable Diffusion) \cite{StableDiffusion} to get the background image, while the text that describes the expression will be used as the search key in the 135-class emotion dataset \cite{EmoFace135, emo135url} to retrieve the expression image. Together with the input identity image, the three images will be used as conditioning in a latent diffusion model \cite{StableDiffusion} to generate the result, which has already exhibited high customizability of various conditions on image generation \cite{ControlNet}.

Based on the foundations of the diffusion model \cite{DDPM2020, StableDiffusion}, we also propose several elaborate designs in SFSR diffusion model. 
1) \emph{Balancing identity and expression encoder}. We develop the identity and expression encoder, which are competitive with each other, to reduce the transfer of residual identity attribute in the expression encoder to the final result.
2) \emph{Improved midpoint sampling}. To achieve both efficiency and accuracy in imposing the identity and expression constraints during training, we propose the improved midpoint sampling, which can generate the face of higher quality with only two times of prediction than the original midpoint sampling \cite{DiffSwap2023}. 
3) \emph{Explicitly background conditioning}. 
We provide background condition in the training phase so that the diffusion model can focus on the generation of face but not background, and get more hints from inputs to recover face pose and lighting. This design is different from previous methods \cite{inpaint2022, DiffSwap2023} that use background image only in inference, and proves to be more effective.  We name the newly designed diffusion model for SFSR task as \emph{DiffSFSR}.
 Finally, the contributions of this paper are summarized as:
\begin{itemize}
    \item A novel face generation framework that achieves simultaneous control of identity and expression, and more fine-grained expression synthesis than state-of-the-art text-to-image methods. 
    \item A novel face manipulation task, simultaneously face swapping and reenactment, which has never been explored by previous methods. This task is also compatible with the traditional separate swapping and reenactment tasks by re-combination of inputs.
    \item Three innovative designs in the conditional diffusion model, including balancing identity and expression encoder, improved midpoint sampling, and explicitly background conditioning, which increase the controllability and image quality. 
    
\end{itemize}

\section{Related Works}

\textbf{Conditional Face Generation.}
Early methods usually use a single modality as conditioning. For example, there has been a surge of text-to-face researches that utilize the pre-trained StyleGAN \cite{StyleGAN2019, StyleGAN2020, StyleGAN2021} and the text encoder, such as TediGAN \cite{TediGAN2021}, StyleCLIP \cite{StyleCLIP2021} and StyleT2I \cite{StyleT2I2022}. Using images as conditioning \cite{SimSwap2020, FaceShifter2019, ReenactGAN2018, FSGAN2019} is also popular in the research community. This kind of methods is also known as face swapping and face reenactment, where the generated face has the same identity or expression with the input face image. The most recent methods begin to use multiple modalities, due to the fact that different modalities are complementary to each other. For example, the tuple of texts and segmentation masks \cite{CollaborativeDM2023, PixelFace+2023} is very popular to control face generation. Our proposed face generation framework also takes as input multiple modalities, text and image.

Our work is also closely related to face swapping and expression reenactment.
The mainstream way to improve visual quality is using GANs \cite{SimSwap2020, SmoothSwap2022, FaceShifter2019, FewShotHeadSwap2022, RegionFaceSwap2022, BlendFace2023, ReenactGAN2018, DeepVideoP2018, GANimation2018, Recycle-GAN2018, CycleGAN2017, StarGAN2018}, which inject the identity or expression features extracted from the source into the face swapping or expression reenactment network, and use multiple losses to ensure semantic consistency and image fidelity. The most recent method \cite{DiffSwap2023} employs a diffusion model, and reformulates the face swapping as a conditional inpainting task. There exist methods \cite{UnifiedFaceTCSVT2022, FSGAN2019, FSGANv22023, UnifiedFaceECCV2022} that combine the two tasks, swapping and reenactment, in a single framework. In their pipeline, a switch operator is usually placed in the facial features transfer stage to switch between swapping and reenactment tasks. The main difference in functionality between these methods and our DiffSFSR is that, either identity or expression, but not both of them, is transferred to the result.

\textbf{Preliminary on Diffusion Models.}\label{sec:dm} The diffusion model (DDPM) \cite{DDPM2020} has been well documented. It contains diffusion and denoising processes. Given a data distribution $\mathbf{x}_0 \sim q(\mathbf{x}_0)$, 
the diffusion process produces a series of intermediate noisy samples $\{\mathbf{x}_t\}$ by continuously adding Gaussian noise $\mathcal{N}$ with variance $\beta_t \in (0,1)$ at timestep $t$: $q(\mathbf{x}_{1:T}|\mathbf{x}_{0})=\prod_{t=1}^{T}q(\mathbf{x}_{t}|\mathbf{x}_{t-1})$ where $q(\mathbf{x}_{t}|\mathbf{x}_{t-1})=\mathcal{N}(\mathbf{x}_{t};\sqrt{1-\beta_{t}}\mathbf{x}_{t-1},\beta_{t}\mathbf{I})$. $x_t$ can be sampled directly from $x_0$, without generating intermediate steps: 

\begin{equation}
\label{eq:4}
q(\mathbf{x}_t\vert \mathbf{x}_0)=\mathcal{N}(\mathbf{x}_t; \sqrt{\Bar{\alpha}_t}\mathbf{x}_0, (1-\Bar{\alpha}_t)\mathbf{I})
\end{equation}
where $\alpha_t = 1-\beta_t$ and $\Bar{\alpha}_t=\prod_{s=1}^{t}\alpha_s$.
When a long increasing sequence $\beta_{1:T}$ is set such that $\Bar{\alpha} \approx 0$, the distribution of $\mathbf{x}_T$ will converge to a standard Gaussian.



The denoising process starts from a Gaussian noise sample $\mathbf{x}_T \sim \mathcal{N}(0, I)$, and denoises $\mathbf{x}_T$ to $\mathbf{x}_0$ by sequentially sampling the posteriors $q(\mathbf{x}_{t-1}|\mathbf{x}_t)$. Based on Bayesian rules, $q(\mathbf{x}_{t-1}|\mathbf{x}_t, \mathbf{x}_0)$ can be derived to: 
\begin{equation}
\label{eq:xt-1xt}
\begin{split}
q(\mathbf{x}_{t-1} \vert \mathbf{x}_{t}, \mathbf{x}_0)=\mathcal{N}(\mathbf{x}_{t-1};\Tilde{\mu}_t(\mathbf{x}_t,\mathbf{x}_0),\Tilde{\beta}_t\mathbf{I}),\quad \quad \quad
\\
\mathbf{x}_{t-1} = \Tilde{\mu}_t(\mathbf{x}_t,\mathbf{x}_0) + \sqrt{\Tilde{\beta}_t}\epsilon,\quad \quad \quad\quad \quad \quad \quad \quad\quad
\\
\text{where  }\Tilde{\mu}_t(\mathbf{x}_t,\mathbf{x}_0)=\frac{\sqrt{\Bar{\alpha}_{t-1}}\beta_t}{1-\Bar{\alpha}_t}\mathbf{x}_0 + \frac{\sqrt{\alpha_{t}}(1-\Bar{\alpha}_{t-1})}{1-\Bar{\alpha}_t}\mathbf{x}_t
\\
\text{and  }\Tilde{\beta}_t=\frac{1-\Bar{\alpha}_{t-1}}{1-\Bar{\alpha}_{t}}\beta_t. 
\end{split}
\end{equation}


Hence, $q(\mathbf{x}_{t-1} \vert \mathbf{x}_{t}, \mathbf{x}_0)$ has no closed-form, and a deep neutral network $p_\theta$ is trained to approximate it.

\begin{figure}
\centering
\includegraphics[width=0.5\textwidth]{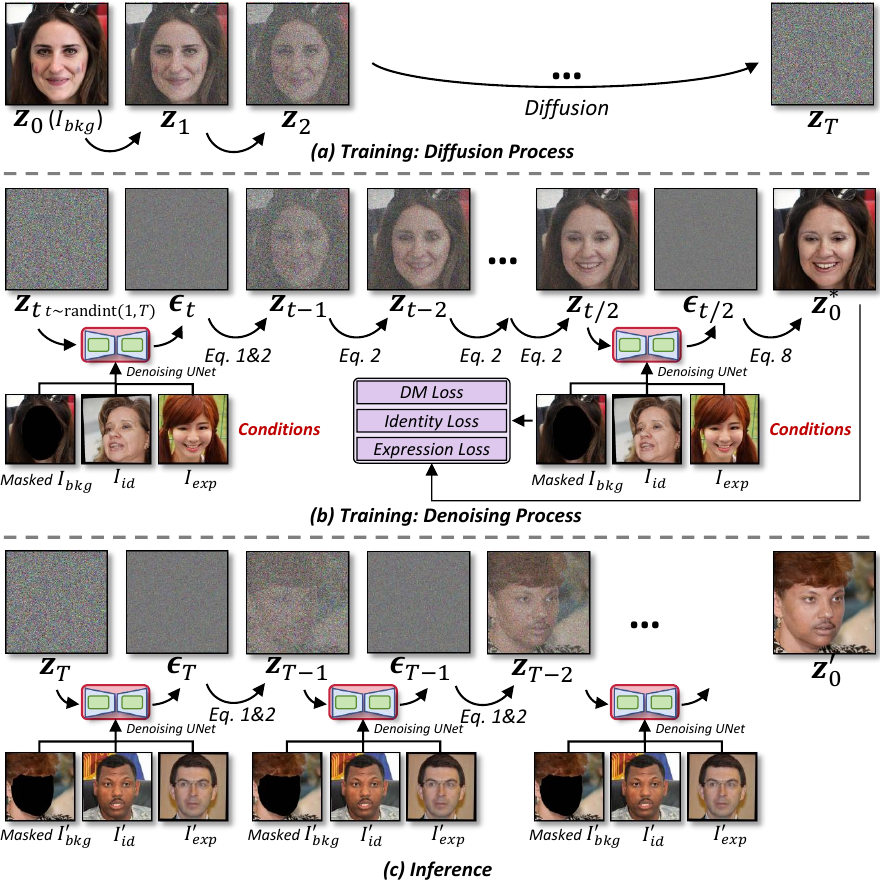} 
\vspace{-14pt}
\caption{Pipeline of DiffSFSR, including training and inference phases. Although the diffusion model is practically trained and tested in the latent space \cite{StableDiffusion}, we illustrate all the processes in the original image space for visualization. 
The transformations between the image space $\mathbf{x}$ and the latent space $\mathbf{z}$ are not illustrated for brevity.}
\vspace{-6pt}
\label{fig:pipeline}
\end{figure}

\section{The Proposed Framework}
\label{sec:methods}

As shown in Fig. \ref{fig:T2I}, the proposed face generation framework contains two main modules: firstly converting the multi-modal inputs into three images, and then generating the face image by SFSR diffusion model from the inputted three images. In the converting module, the inputs contain an identity image $I_{id}$, a text prompt $P_{bkg}$ describing the scene and a text prompt $P_{exp}$ related to the expression label. $P_{bkg}$ is injected to a pre-trained text-to-image diffusion model \cite{StableDiffusion} to obtain the background image $I_{bkg}$. 

$P_{exp}$ is used as the search key in the emotion dataset \cite{EmoFace135} containing 135 emotion categories consisting of 728,946 facial images. According to $P_{exp}$, an expression image $I_{exp}$ is randomly retrieved from the corresponding category. In the DiffSFSR, $I_{bkg}$, $I_{id}$ and $I_{exp}$ are used as conditioning to generate the final result $I_{out}$. 

The pipeline of the DiffSFSR is shown in Fig. \ref{fig:pipeline}, including the training and inference phases. The latent diffusion model \cite{StableDiffusion} is chosen as the backbone due to its high customizability for various conditions. Similar to DDPM \cite{DDPM2020}, the training of the latent diffusion model also consists of the diffusion process and the denoising process. One sample in the training data is the triplet $[I_{bkg}, I_{id}, I_{exp}]$, excluding the ground truth counterpart of $I_{out}$. The input $I_{bkg}$ is firstly embedded to a latent $\mathbf{z}_0$, and then be added with the Gaussian noise in the diffusion process. The denoising process denoises the latent $\mathbf{z}_t$ at random \emph{t}-th timestep to $\mathbf{z}_0^*$, by using two times the UNet conditioned on the masked $I_{bkg}$ and the extracted features from $I_{id}, I_{exp}$. The denoised latent $\mathbf{z}_0^*$ further needs to be transformed back to the image space to generate $\mathbf{x}_0^*$. To better compute the identity loss between $\mathbf{x}_0^*$ and $I_{id}$, and the expression loss between $\mathbf{x}_0^*$ and $I_{exp}$, we adopt the improved midpoint sampling, considering both accuracy and efficiency. In the inference phase, the trained denoising UNet is used multiple times to gradually generate $\mathbf{z}_0^{\prime}$ from a Gaussian noise $\mathbf{z}_T$, which is finally decoded to image $\mathbf{x}_0^{\prime}$.


\subsection{Conditions}

\begin{figure}
\centering
\includegraphics[width=0.47\textwidth]{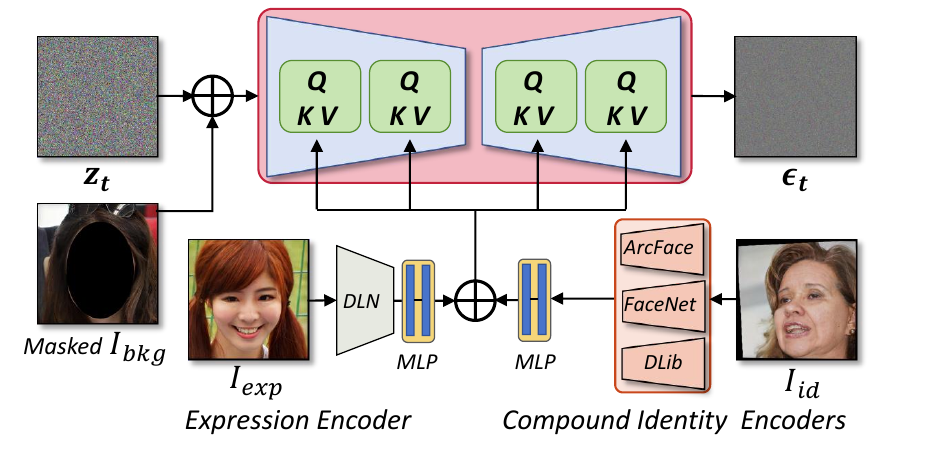} 
\vspace{-6pt}
\caption{The network architecture of the denoising UNet. QKV denotes the cross-attention layer.}
\vspace{-6pt}
\label{fig:condtions}
\end{figure}

The key in the DiffSFSR is how to learn the disentangled representations respectively for $I_{bkg}, I_{id}, I_{exp}$ and then condition the diffusion model on the learned embeddings simultaneously, to enable disentangled and precise control of the targeted face. The network architecture of the denoising UNet, together with the three conditioning embeddings, is shown in Fig. \ref{fig:condtions}.


\textbf{Background.} 
Except for identity and expression, the input $I_{bkg}$ provides all other attributes for the output, e.g. face pose, hair, glasses, lighting and surroundings. We mask the facial region in $I_{bkg}$ with a face parsing method, turning the task from face editing to face inpainting. Then, the masked $I_{bkg}$ is concatenated with the latent $\mathbf{z}_i$ as conditioning for the diffusion model, in both training and inference phases, so that most parts of the background in $I_{out}$ are exactly the same with $I_{bkg}$. 

Our way of preserving background attributes is totally different from the recently proposed diffusion model based face-swapping method, DiffSwap \cite{DiffSwap2023}. In DiffSwap, the background pixels (masked $I_{bkg}$) 
are not explicitly provided during training, but only during the inference phase, which requires the diffusion model to reconstruct the background pixels in the training phase. To ensure global consistency, the reconstruction loss is computed on the whole image. Since the facial region usually occupies less than 50\% of the total image, a significant portion of the network optimization is initially dedicated to the reconstruction of background pixels before the network starts to perform fine-grained generation in the facial region. Hence, the faces generated by DiffSwap may suffer from image blur and low quality, compared with our results. 

Another advantage of explicitly providing background pixels in the training is that the diffusion model is forced to learn to estimate the face pose and lighting from the background pixels, as there exists a strong correlation between face pose, lighting, and the background. To summarize, we provide the masked $I_{bkg}$ as conditioning in the hope that, it can make the diffusion model focus on the generation of face but not the generation of background, and provide more hints for recovering important attributes. We are the first to disentangle background attributes from identity and expression and explicitly make them as conditioning, in the task of face manipulation.

\textbf{Expression.}
Previous methods prefer to use 2D facial landmarks as the expression representation. Human expression contains complicated and subtle facial movements and is closely related to facial texture, such as facial wrinkles and facial action unit activation, thus the 2D facial landmarks are not enough to represent the accurate expression attribute. In order to get a powerful expression representation, we adopt the identity-disentangled and fine-grained expression representation network named DLN \cite{DLN2021} as the encoder. A two-layer MLP is then used for domain transformation and feature shape alignment. After that, the expression embedding is injected into the diffusion model through the cross-attention module.

\textbf{Identity.}
The identity encoder should be competitive with the expression encoder \cite{DLN2021} for balanced conditioning, otherwise, the residual identity attribute in the expression encoder will be accidentally transferred to the result. As the expression is closely entangled with identity, designing a completely identity-ignored expression encoder is still an open problem in the field of face analysis. In the aspect of identity encoder, we apply an identity compound embedding, since a single identity embedding is usually biased \cite{compound2022} and insufficient to balance the expression embedding.
Three state-of-the-art face recognition models \cite{arcface2019, facenet2015, dlib09} are selected to construct the identity compound embedding. Experiments in Sec. \ref{sec:ablation} demonstrate that a single embedding can't compensate for the impact of the residual identity in the expression encoder, while the compound embedding can meet the requirement. Similar to the expression encoder, we use a two-layer MLP to map the identity embedding of different shapes to a uniform dimension and condition them through cross-attention.

\subsection{Training Objective}

The training objective of the diffusion model can be formulated as the Mean Squared Error (MSE) loss: 
\begin{equation}
\label{eq:DMloss}
L_{DM} = \mathbb{E}_{\mathbf{z}_t, \mathbf{C},\epsilon,t}[||\epsilon - \epsilon_\theta (\mathbf{z}_t,\mathbf{C},t) ||_2^{2}],
\end{equation}
where $\mathbf{z}_t$ denotes the noisy latent obtained by adding noise $\epsilon$ to $\mathbf{z}_0$ at \emph{t}-th timestep, $\epsilon_{\theta}$ denotes the denoising UNet learned to predict $\epsilon$, and $\mathbf{C}$ denotes the conditions. The conditions are defined as:
\begin{equation}
\label{eq:conditions}
\mathbf{C} = [M \odot I_{bkg}, \mathcal{E}_{id}(I_{id}), \mathcal{E}_{exp}(I_{exp})],
\end{equation}
where $M$ denotes the binary mask of the facial region, $\mathcal{E}_{id}(\cdot)$ denotes the identity encoder, and $\mathcal{E}_{exp}(\cdot)$ denotes the expression encoder.

We also use the identity loss and the expression loss, following the common practice. The losses are defined as:
\begin{equation}
\label{eq:id}
L_{id} = 1-\text{CosSim}(\mathcal{E}_{id}(I_{id}), \mathcal{E}_{id}(\mathcal{D}_{DM}(\mathbf{z}_0^*))),
\end{equation}
\begin{equation}
\label{eq:exp}
L_{exp} = \text{MSE}(\mathcal{E}_{exp}(I_{exp}), \mathcal{E}_{exp}(\mathcal{D}_{DM}(\mathbf{z}_0^*))),
\end{equation}
where $\text{CosSim}(\cdot, \cdot)$ denotes the cosine similarity function and $\text{MSE}(\cdot, \cdot)$ denotes the MSE function. The final loss function can be formulated as:
\begin{equation}
L = L_{DM} + \lambda_1 L_{id} + \lambda_2 L_{exp},
\end{equation}
where $\lambda_1, \lambda_2$ denote the hyper-parameters.

\subsection{Improved Midpoint Sampling}
\label{sect:IMS}

To compute the identity and expression losses at \emph{t}-th timestep during training, the denoised latent $\mathbf{z}_0^*$ firstly needs to be generated from the noisy latent $\mathbf{z}_t$. In the original diffusion model, e.g. DDPM \cite{DDPM2020}, the generation of $\mathbf{z}_0^*$ requires multiple times sampling on different timesteps, which is unacceptable in the training phase of high-fidelity image generation task. To tackle this issue, DiffSwap \cite{DiffSwap2023} proposes a midpoint sampling method, which can get a coarse $\mathbf{z}_0^*$ with only two steps of sampling. Specifically, in timestep $t$ it firstly estimates $\mathbf{z}_{t_1}$, $t_1=\lfloor \frac{t}{2} \rfloor$ by using the formula:
\begin{equation}
\label{eq:diffswap1}
\mathbf{z}_{t_1} = \frac{\mathbf{z}_t-\sqrt{1-\Bar{\alpha}_t/\Bar{\alpha}_{t_1}}\epsilon_\theta(\mathbf{z}_t,t,\mathbf{C})}{\sqrt{\Bar{\alpha}_t/\Bar{\alpha}_{t_1}}}.
\end{equation}
Then, starting from the estimated $\mathbf{z}_{t_1}$, it predicts the final $\mathbf{z}_0^*$ by using the formula:
\begin{equation}
\label{eq:diffswap2}
\mathbf{z}_0^* = \frac{\mathbf{z}_{t_1}-\sqrt{1-\Bar{\alpha}_{t_1}}\epsilon_\theta(\mathbf{z}_{t_1},t_1,\mathbf{C})}{\sqrt{\Bar{\alpha}_{t_1}}}
\end{equation}
It seems to be an appealing solution to compute the identity and expression losses with only two steps of sampling. We refer readers to the literature \cite{DiffSwap2023} for the detailed derivation of the two formulas in Eq. \ref{eq:diffswap1} and Eq. \ref{eq:diffswap2}.

However, we find that there is an issue in Eq. \ref{eq:diffswap1}. The estimated noise here should be the noise that can convert $\mathbf{z}_{t_1}$ into $\mathbf{z}_t$, but the noise estimated by $\epsilon_\theta(\mathbf{z}_t,t,\mathbf{C})$ is actually the noise that convert $\mathbf{z}_0$ to get $\mathbf{z}_t$. 
For a noisy latent $\mathbf{z}_t$, we can get $\mathbf{z}_{t-1}$ using the following process:
firstly predicts noise $\epsilon$ using the Denoising UNet $\epsilon_\theta(\mathbf{z}_t,t,\mathbf{C})$, then calculates $\mathbf{z}_0$ with the inverse process of Eq. \ref{eq:4}, finally gets $\mathbf{z}_{t-1}$ using Eq. \ref{eq:xt-1xt}. In other words, strictly following the formulas in DDPM, we are only allowed to move to $\mathbf{z}_0$ and then $\mathbf{z}_{t-1}$, starting from $\mathbf{z}_t$. The direct moving from $\mathbf{z}_t$ to $\mathbf{z}_{t_1}$ in DiffSwap is suboptimal, which will degrade the overall performance.

To be more in line with the formulas in DDPM, we propose an improved midpoint sampling method, which also samples within two steps but can reduce the information loss, compared to the original midpoint sampling \cite{DiffSwap2023}. Specially, starting from $\mathbf{z}_t$, we can obtain $\mathbf{z}_0$ and $\mathbf{z}_{t-1}$ using the process introduced above. Then, we can obtain  $\mathbf{z}_{t-2}$ using Eq. \ref{eq:xt-1xt} again, which is an efficient linear transformation without using the Denoising UNet.
Through repeating $t-t_1$ times linear transformation, we can get $\mathbf{z}_{t_1}$ in a more accurate and graceful way than DiffSwap.
Finally, $\mathbf{z}_0^*$ is obtained by using Eq. \ref{eq:diffswap2} on $\mathbf{z}_{t_1}$. 

\section{Experiments}

\textbf{Dataset.} We split the CelebA-HQ dataset \cite{CelebA-HQ2017} into a training set of 29,000 images and a test set of 1,000 images, by random selection. Our diffusion model is trained on the training set of CelebA-HQ and FFHQ \cite{StyleGAN2019}, and evaluated in the test set of CelebA-HQ and FF++ \cite{FaceForensics++2019}. The competitors are evaluated by using their public pre-trained networks or other open-source projects. 

\textbf{Metrics.} The quantitative evaluations are performed in terms of four metrics: identity retrieval accuracy (ID.), expression error (Exp.), pose error (Pose.), and mean squared error (MSE.). For ID., we employ CosFace \cite{CosFace2018} to perform identity retrieval. For Exp., we adopt the expression embedding model \cite{DLN2021} to compute the Euclidean distance between $I_{out}$ and $I_{exp}$. For Pose., we use a pose estimator \cite{PoseEstimation2018} to estimate head pose and compute the Euclidean distance between $I_{out}$ and $I_{bkg}$. MSE. is used to measure the pixel difference between the estimated image and ground truth. It is noteworthy that in the calculation of the metrics in different swapping tasks, the reference images are accordingly changed with the source images.

\textit{Implementation Details.} The network architecture of our DiffSFSR follows the latent diffusion model \cite{StableDiffusion}, which has a $4\times64\times64$ latent space. DiffSFSR is trained from SD-1.4 in $512 \times 512$ resolution with an AdamW optimizer. The hyper-parameters are set as $\lambda_{1} = 0.003$, $\lambda_{2} = 0.01$. 
In the first 100k steps, the learning rate is set to $1e-5$ which decays linearly in the following 100k steps. 8 NVIDIA Tesla A100 GPUs are used to train our diffusion model with a global batch size of 64. In inference time, we apply a PNDM \cite{PNDM2021} sampler with 50 steps, which takes roughly 1 second to generate an image.

\subsection{Fine-grained Expression Controlling Results}
\begin{figure*}
\centering
\includegraphics[width=0.92\textwidth]{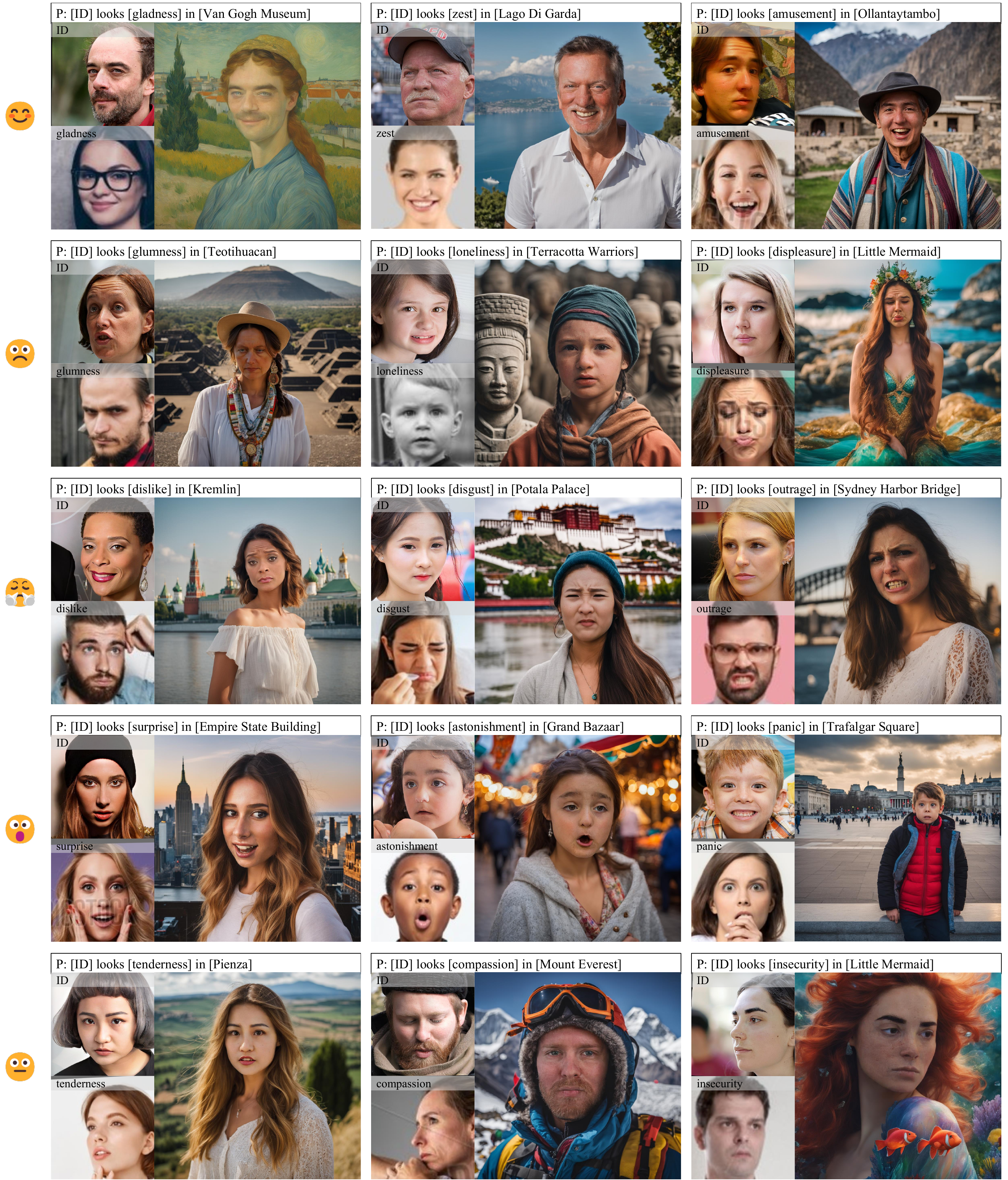} 
\vspace{-10pt}
\caption{A subset of 135 classes expression synthesis samples. Please zoom in for more details.}
\label{fig:p-1}
\end{figure*}

Fig. \ref{fig:p-1} shows samples of fine-grained expression synthesis results. More samples of the full set of 135 expression text labels \cite{EmoFace135}  can be observed in Sec. \ref{sec:allresults}. To the best of our knowledge, \emph{there is currently no face generation or manipulation method in academic or industry, that can reach this level of fine-grained expression control.} As mentioned above, the dataset \cite{EmoFace135} provides 728,946 facial expression images labeled with 135 categories. In our work, a reference expression image is randomly selected from the corresponding category according to the input expression text. As observed in Fig. \ref{fig:p-1} and Sec. \ref{sec:allresults}, the synthesized facial expression is similar to the facial expression in the reference image. Readers can zoom in for more details.

\textbf{User Study.} We conduct a user study to evaluate the quality of the fine-grained expression synthesis, in terms of expression consistency and ID consistency. For each sample, 27 participants were recruited to answer two questions: 1) whether the synthesized face has a consistent ID with the ID image; 2) whether the synthesized face has a consistent expression with the reference expression image. The user study results demonstrate the ability of simultaneous ID and expression preserving with the ID consistency of $95.6\%$ (with a variance of $2.6\%$) and the expression consistency of $90.4\%$ (with a variance of $3.4\%$).

We further conduct another user study to evaluate the expression consistency of the proposed method. For each sample, 30 participants were recruited to score the expression consistency between the synthesized face and the reference expression image (with a minimum score of 1 and a maximum score of 5; 1 refers to very inconsistent and 5 refers to very consistent). Finally, our method achieves a score of $4.08$ (with a variance of $0.81$), indicating its ability to generate very consistent expressions.

In summary, our method is capable of achieving fine-grained expression control while maintaining ID consistency.

\begin{table}[]
\scriptsize
\begin{tabular}{cl|lll|lll}
\hline
\multicolumn{2}{c|}{\multirow{2}{*}{Methods}}                                                              & \multicolumn{3}{c|}{CelebA-HQ}                                                       & \multicolumn{3}{c}{FF++}                                                            \\ \cline{3-8} 
\multicolumn{2}{c|}{}                                                                                      & \multicolumn{1}{c}{ID.$\uparrow$} & \multicolumn{1}{c}{Exp.$\downarrow$} & \multicolumn{1}{c|}{Pose.$\downarrow$}      & \multicolumn{1}{c}{ID.$\uparrow$} & \multicolumn{1}{c}{Exp.$\downarrow$} & \multicolumn{1}{c}{Pose.$\downarrow$}      \\ \hline
\multicolumn{1}{c|}{\multirow{3}{*}{\begin{tabular}[c]{@{}c@{}}Swap\\ EXP\end{tabular}}} & Face2Face       & 87.9                    & 3.25                     & \multicolumn{1}{c|}{-}          & 96.8                    & 2.83                     & \multicolumn{1}{c}{\textbf{-}} \\
\multicolumn{1}{c|}{}                                                                    & StyleHEAT       & 98.2                    & 2.43                     & \multicolumn{1}{c|}{\textbf{-}} & 97.7                    & 2.16                     & \multicolumn{1}{c}{\textbf{-}} \\
\multicolumn{1}{c|}{}                                                                    & DiffSFSR (ours) & \textbf{99.9}           & \textbf{0.58}            & \multicolumn{1}{c|}{-}          & \textbf{98.9}           & \textbf{0.68}            & \multicolumn{1}{c}{\textbf{-}} \\ \hline \hline
\multicolumn{1}{c|}{\multirow{6}{*}{\begin{tabular}[c]{@{}c@{}}Swap\\ ID\end{tabular}}}  & FaceShifter     & 94.5                    & 0.65                     & \textbf{2.13}                    & 95.4                    & 1.10                     & \textbf{1.62}                  \\
\multicolumn{1}{c|}{}                                                                    & SimSwap         & \textbf{98.8}           & 0.93                     & 2.89                            & \textbf{98.0}           & 1.46                     & 2.87                           \\
\multicolumn{1}{c|}{}                                                                    & HifiFace        & 85.1                    & 1.11                     & 3.38                            & 92.4                    & 1.80                     & 3.32                           \\
\multicolumn{1}{c|}{}                                                                    & E4S             & 81.6                    & 2.77                     & 6.99                            & 91.5                    & 2.34                     & 3.96                           \\
\multicolumn{1}{c|}{}                                                                    & DiffSFSR (ours) & 90.8                    & \textbf{0.32}            & 2.59                            & 91.0                    & \textbf{0.49}            & 3.89                           \\ \hline \hline
\multicolumn{1}{c|}{\multirow{2}{*}{\begin{tabular}[c]{@{}c@{}}Swap\\ All\end{tabular}}} & Hybrid Method   & 76.9                    & 2.24                     & 7.07                            & 83.5                    & 2.14                     & 5.67                           \\
\multicolumn{1}{c|}{}                                                                    & DiffSFSR (ours) & \textbf{90.2}           & \textbf{0.55}            & \textbf{6.00}                   & \textbf{90.7}           & \textbf{0.73}            & \textbf{5.19}                  \\ \hline
\end{tabular}
\vspace{-6pt}
\caption{The quantitative results in the three tasks: ``Swap All" denotes SFSR task, ``Swap ID" denotes face swapping task, and ``Swap EXP" denotes face reenactment task.
The scores in ID. and Exp. are scaled up by a factor of 100 for simplicity.}
\vspace{-6pt}
\label{tab:main}
\end{table}


\subsection{Comparisons}
\begin{figure*}
\centering
\includegraphics[width=0.88\textwidth]{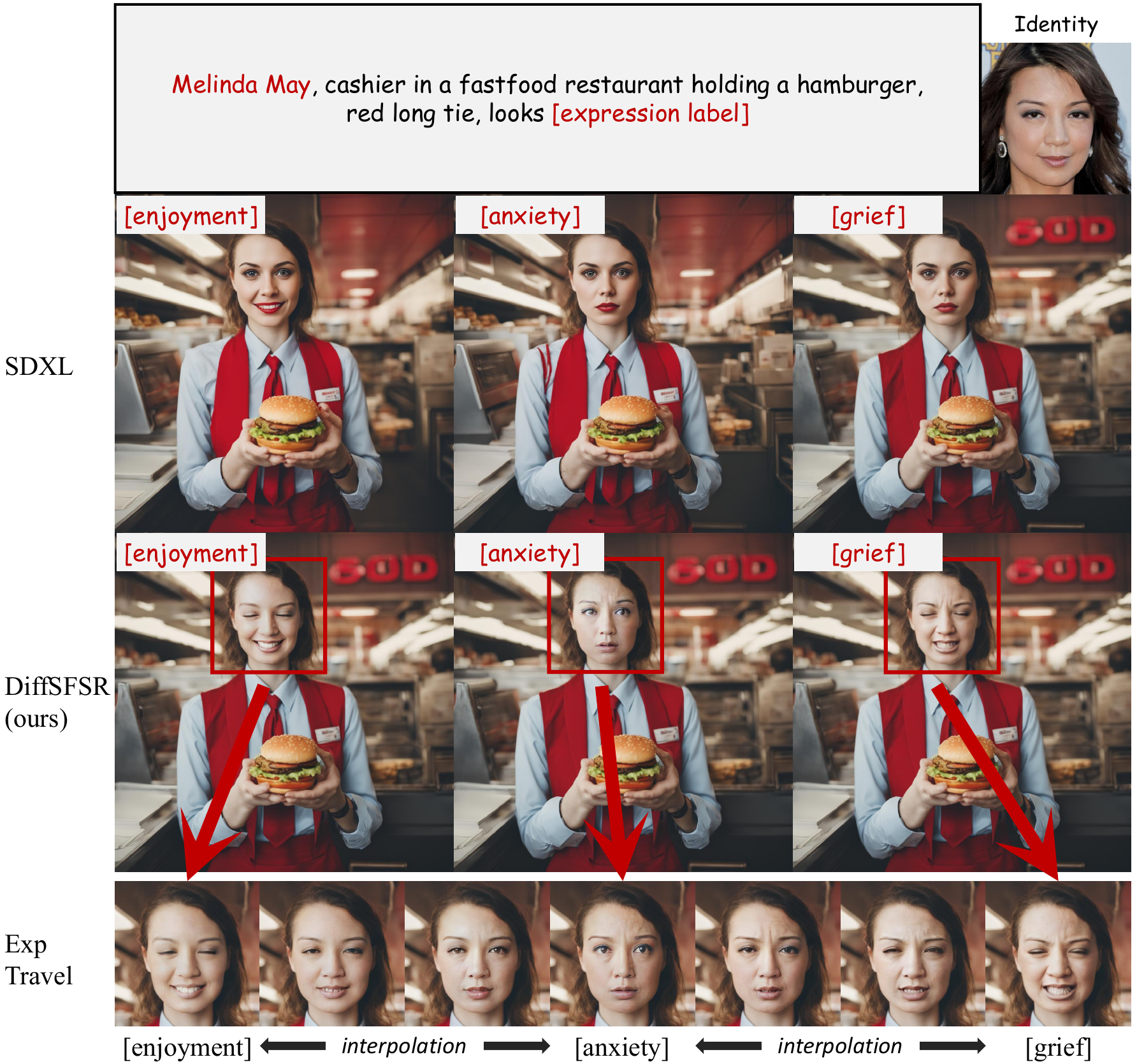} 
\vspace{-7pt}
\caption{Comparison with text-to-image method SDXL, and illustration of expression travel.}
\vspace{-4pt}
\label{fig:T2I_com}
\end{figure*}

\textbf{Comparison with Text-to-Image Method.} We compare our method with the SOTA open-source text-to-image model Stable Diffusion XL (SDXL) \cite{sdxl2023}. As shown in Fig. \ref{fig:T2I_com}, the two methods take as input the same prompt and the same fine-grained expression labels. The additional input to our framework is a portrait of ``Melinda May". Compared with our results, SDXL can't synthesize the accurate expressions corresponding to ``enjoyment'', ``anxiety" or ``grief''. The dilemma faced by most text-to-image methods is that they can only recognize a few limited expression labels. Models like ControlNet \cite{ControlNet} need a large amount of training data and couple the background to the face attributes. Our framework also supports expression travel by interpolating between embeddings to further explore fine-grained expression. 


\textbf{Comparison with Hybrid Methods.} As there is currently no method that can perform SFSR, we construct a hybrid method as a potential competitor by directly combining the best face-swapping method in terms of ID. score, SimSwap\cite{SimSwap2020}, and the best face reenactment method in terms of Exp. score, StyleHEAT\cite{styleheat2022}, based on the quantitative comparisons reported in the part of Swap ID and Swap EXP in Tab. \ref{tab:main}. 

\begin{figure}
\centering
\includegraphics[width=0.47\textwidth]{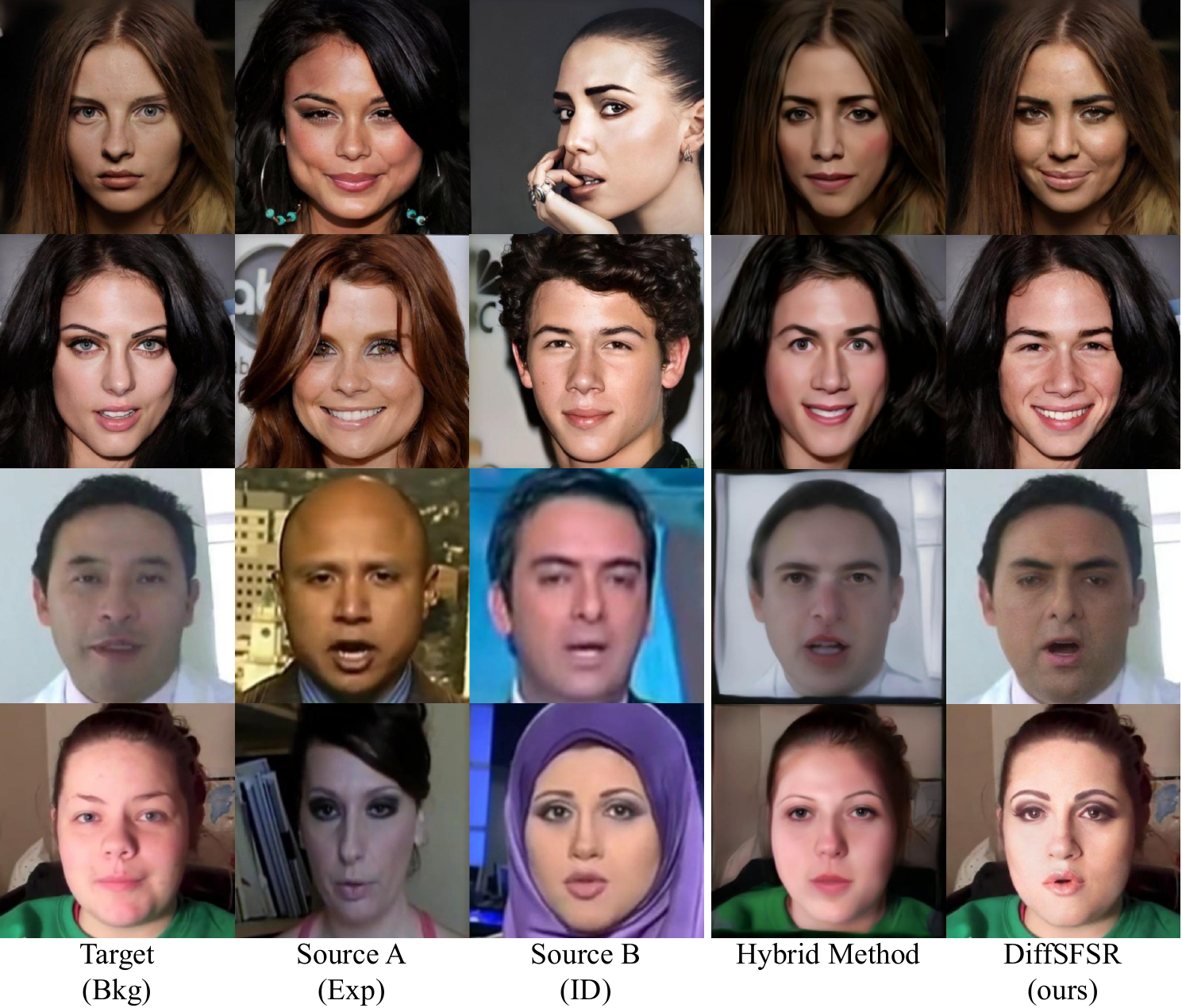} 
\vspace{-8pt}
\caption{Qualitative comparison in SFSR task.}
\vspace{-9pt}
\label{fig:swap_all}
\end{figure}

As shown in the part of Swap All in Tab. \ref{tab:main}, our method outperforms the hybrid method with a significant margin in all metrics. From the qualitative results shown in Fig. \ref{fig:swap_all}, our method can produce more accurate expressions and poses than the hybrid method, due to the powerful ability of expression embedding \cite{DLN2021} and latent diffusion model \cite{StableDiffusion}.

\textbf{Comparison with Face Reenactment Methods.} 
Our DiffSFSR can also conduct the face reenactment task by setting the inputs $I_{bkg}$ and $I_{id}$ as the same image. 
We make a comparison with two SOTA face reenactment methods including Face2Face\cite{face2face2016} and StyleHEAT\cite{styleheat2022}, and use their pre-trained networks. 

\begin{figure}
\centering
\includegraphics[width=0.47\textwidth]{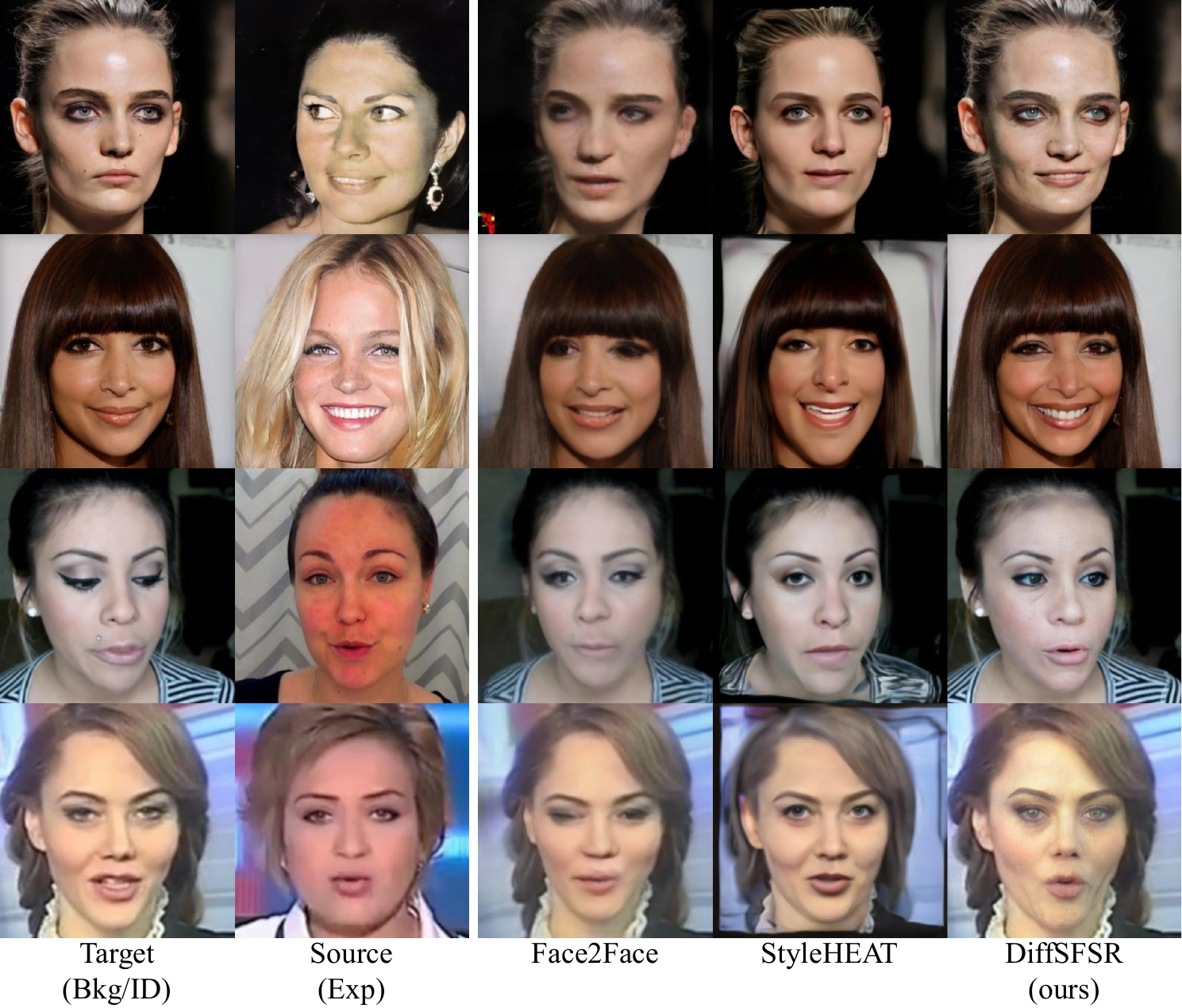} 
\vspace{-8pt}
\caption{Qualitative comparison in face reenactment task.}
\vspace{-8pt}
\label{fig:swap_exp}
\end{figure}

From the statistics in the part of Swap EXP in Tab. \ref{tab:main}, our method outperforms the two competitors in all metrics. The advantage is more obvious in the qualitative comparison shown in Fig. \ref{fig:swap_exp}, where the expression in our results is more similar to the source, and the identity and pose are more similar to the target.

\textbf{Comparison with Face Swapping Methods.} 
Similar to face reenactment, our DiffSFSR can also conduct the face swapping task by setting the inputs $I_{bkg}$ and $I_{exp}$ as the same. We make comparisons with five SOTA face swapping methods including FaceShifter\cite{FaceShifter2019}, SimSwap\cite{SimSwap2020}, HifiFace\cite{HifiFace2021}, E4S\cite{E4S2023} and DiffSwap\cite{DiffSwap2023}. For SimSwap, E4S and DiffSwap, we directly use their public pre-trained networks.
As FaceShifter and HifiFace do not make their codes publicly available, we use the implementations from the open-source community\footnote{\url{https://github.com/richarduuz/Research_Project}}\footnote{\url{https://github.com/xuehy/HiFiFace-pytorch}}. 

\begin{figure}
\centering
\includegraphics[width=0.47\textwidth]{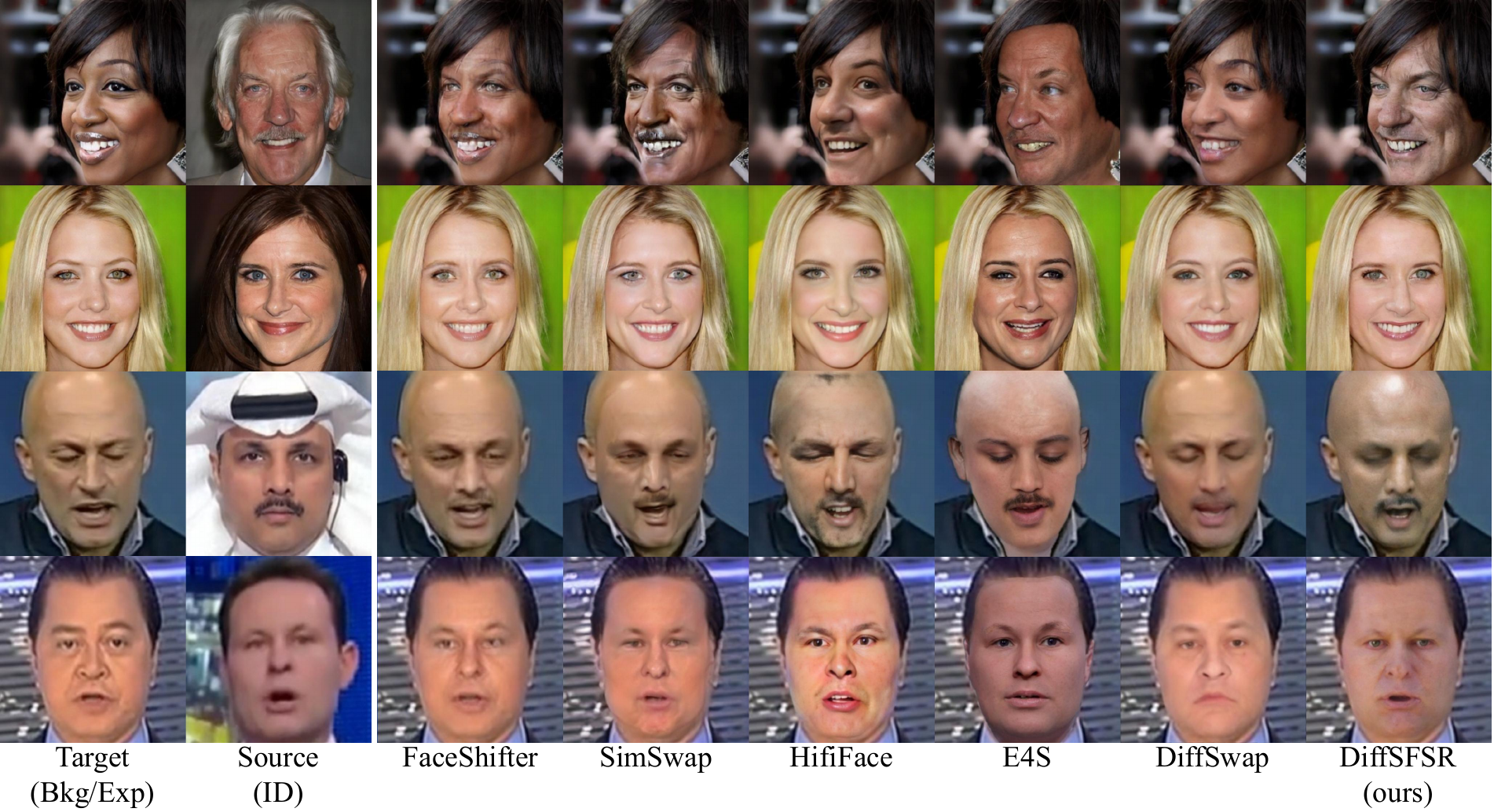} 
\vspace{-6pt}
\caption{Qualitative comparison in face swapping task.}
\vspace{-8pt}
\label{fig:swap_id}
\end{figure}

From the statistics in the part of Swap ID in Tab. \ref{tab:main},   since we focus on both ID and expression, our method outperforms all the competitors in Exp. and achieves promising performance in ID. and Pose. As observed in Fig. \ref{fig:swap_id}, except for SimSwap, our results are more similar to the source faces in terms of inner facial features, e.g., beard. 
There are obvious artifacts in the results of HifiFace. And the faces generated by E4S do not blend well into the background, leading to less natural results. In addition to expression preserving, our advantage over SimSwap is that our method can generate faces with better image quality, with less blur and artifacts, due to the powerful image generation capability of the latent diffusion model \cite{StableDiffusion}.

\begin{table*}[]
\centering
\begin{tabular}{l|llll}
\hline
Methods         & \multicolumn{1}{c}{Identity.} & \multicolumn{1}{c}{Expression.}              & \multicolumn{1}{c}{Realism.} & \multicolumn{1}{c}{Image Quality.} \\ \hline
SimSwap         & 2.84 (p=0.124)                & \multicolumn{1}{c}{3.44 (p\textless{}0.001)} & 2.71 (p\textless{}0.001)     & 2.69 (p\textless{}0.001)           \\
DiffSwap        & 2.58 (p\textless{}0.001)      & 3.59 (p\textless{}0.05)                               & 3.19 (p\textless{}0.005)               & 3.04 (p\textless{}0.001)           \\
E4S             & 2.87 (p=0.232)                & 2.72 (p\textless{}0.001)                     & 3.04 (p\textless{}0.001)     & 3.03 (p\textless{}0.001)           \\
DiffSFSR (ours) & \textbf{3.01}                 & \textbf{3.79}                                         & \textbf{3.72}                & \textbf{3.60}                      \\ \hline
\end{tabular}
\caption{Uers study in face swapping methods. The best values are highlighted in bold. The ANOVA tests are conducted, in which a p-value less than 0.05 is considered to indicate a statistically significant difference from the performance of our method.}
\label{tab:userstudy_3}
\end{table*}


\textit{User Study with Face Swapping Methods.}
To comprehensively compare our method with other face-swapping methods, we implemented another human evaluation experiment. For simplicity, we compare our DiffSFSR with the most recent methods of DiffSwap\cite{DiffSwap2023} and E4S\cite{E4S2023}, as well as SimSwap\cite{SimSwap2020} due to its superior performance in terms of ID consistency (see Tab. \ref{tab:main}). 50 participants were recruited to score the results of all methods in terms of ID consistency, expression consistency, realism, and image quality (with a minimum score of 1 and a maximum score of 5; 1 refers to the worst and 5 refers to the best). As shown in Tab. \ref{tab:userstudy_3}, our method registers the topmost scores across all measured metrics. Moreover, in terms of statistical significance, our approach is competitive with other methods in ID consistency while considerably surpassing its competitors on expression consistency, realism, and image quality under a p-value of 0.05. In summary, our method is comparable to other methods in terms of ID consistency but can produce more accurate expressions and more realistic, high-quality images.



\begin{figure}

\centering
\includegraphics[width=0.47\textwidth]{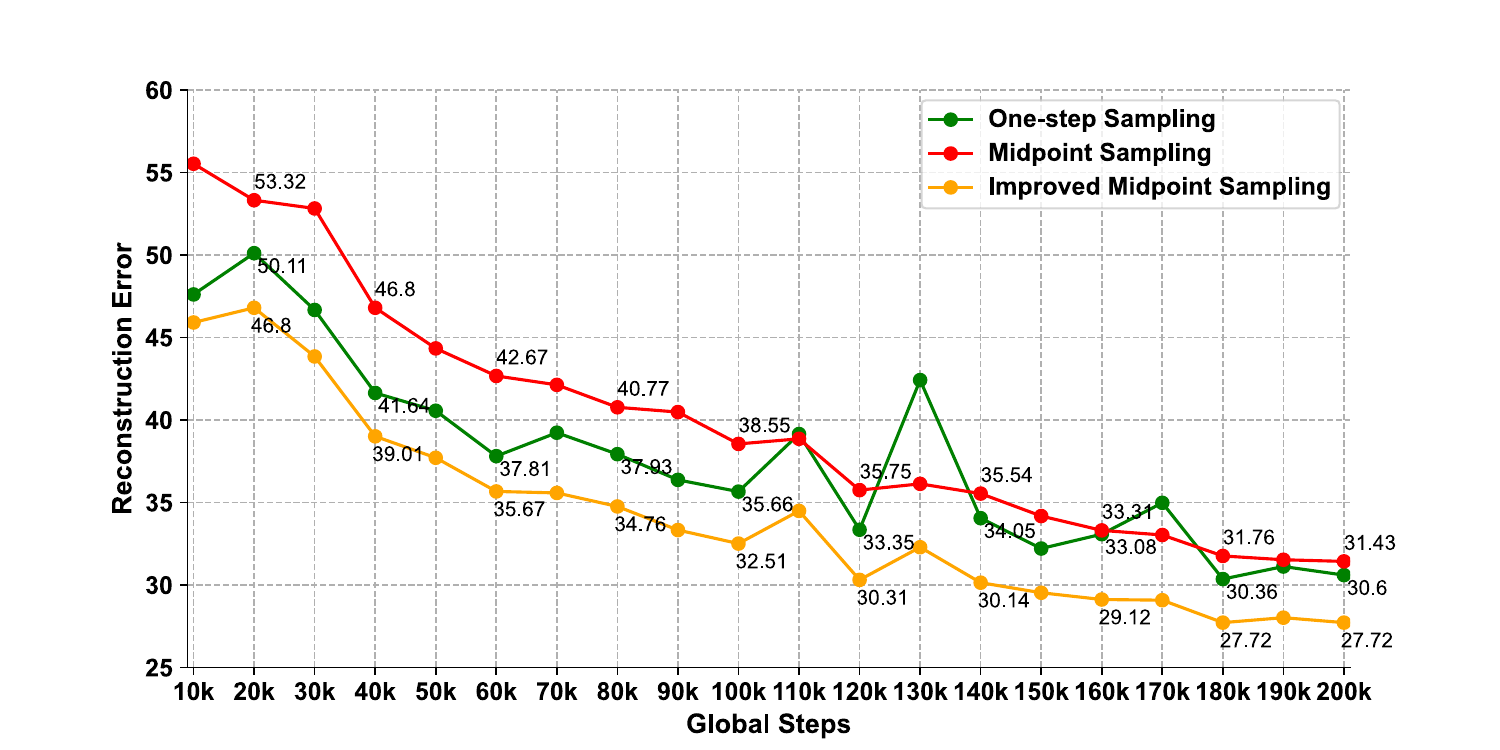} 
\vspace{-10pt}
\caption{Image reconstruction performance of different sampling methods.}
\label{fig:sampling_MSE}
\end{figure}

\begin{figure}
\centering
\includegraphics[width=0.47\textwidth]{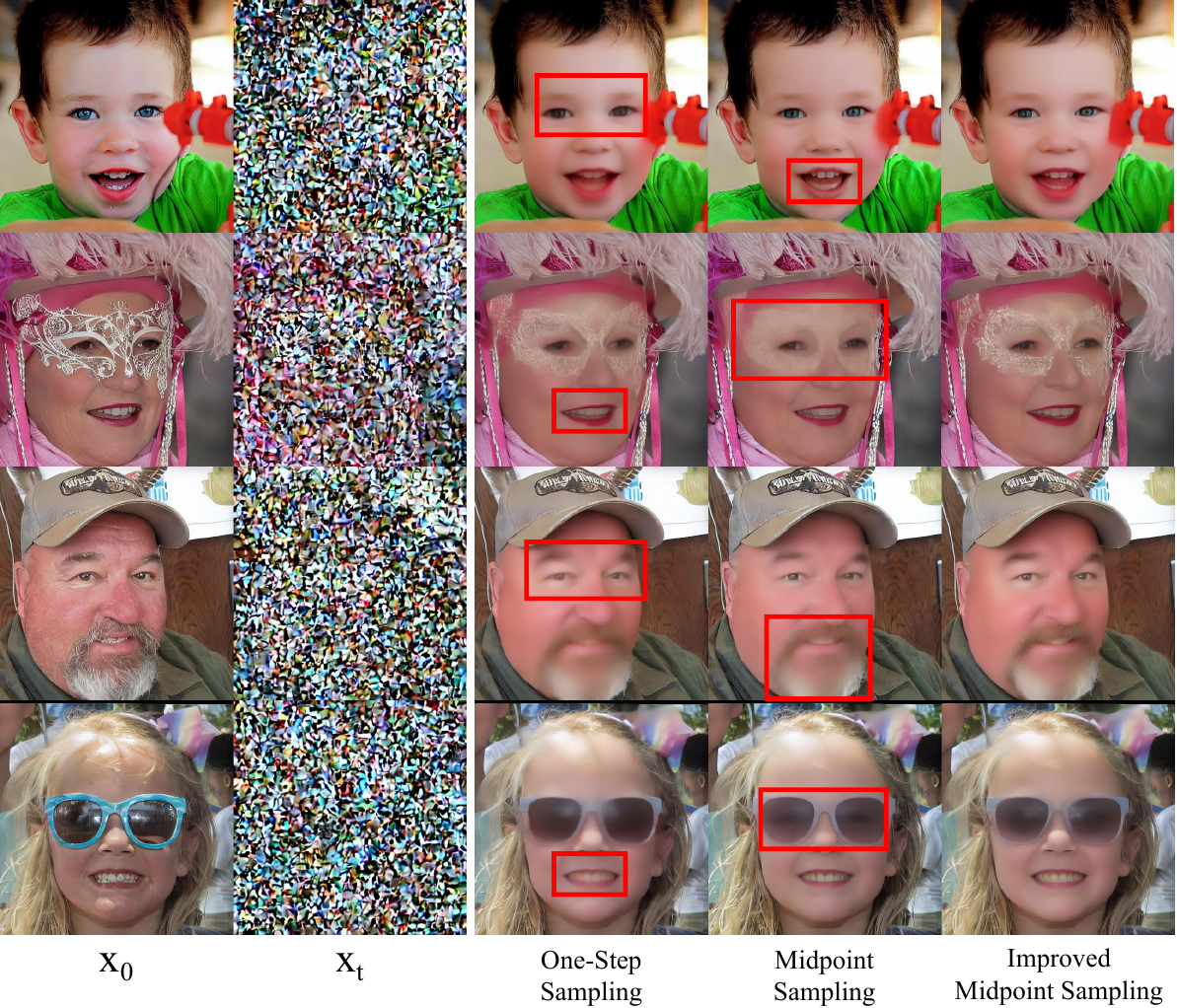} 
\vspace{-8pt}
\caption{Image reconstruction results by using different sampling methods.}
\vspace{-8pt}
\label{fig:ablation1_sampling}
\end{figure}

\textbf{Study on Improved Midpoint Sampling.}
As introduced in Sect. \ref{sect:IMS}, we need $\mathbf{x}_0^*$ during training to impose the identity and expression constraints. The more accurate estimation of $\mathbf{x}_0^*$ is, the more accurate identity and expression losses are.
We conduct an experiment on the dataset constructed by randomly selecting 500 image pairs from FFHQ, to evaluate three sampling methods: 1) one-step sampling using Eq. \ref{eq:4}, 2) midpoint sampling used in \cite{DiffSwap2023}, 3) our proposed improved midpoint sampling. MSE is used to measure the error between sampling results and ground truth, thus showing the image reconstruction performance.

As shown in Fig. \ref{fig:sampling_MSE}, all sampling methods can decrease the reconstruction errors along with the training steps increasing. Our sampling method can achieve lower MSE than others in all periods.  Fig. \ref{fig:ablation1_sampling} shows the reconstruction results by denoising $\mathbf{x}_t$ using different sampling methods. Our results are not only more faithful to ground truth $\mathbf{x}_0$, but also more realistic and clear in the regions of eyes, mouths and even the reflection of sunglasses. 

\begin{figure}
\centering
\includegraphics[width=0.47\textwidth]{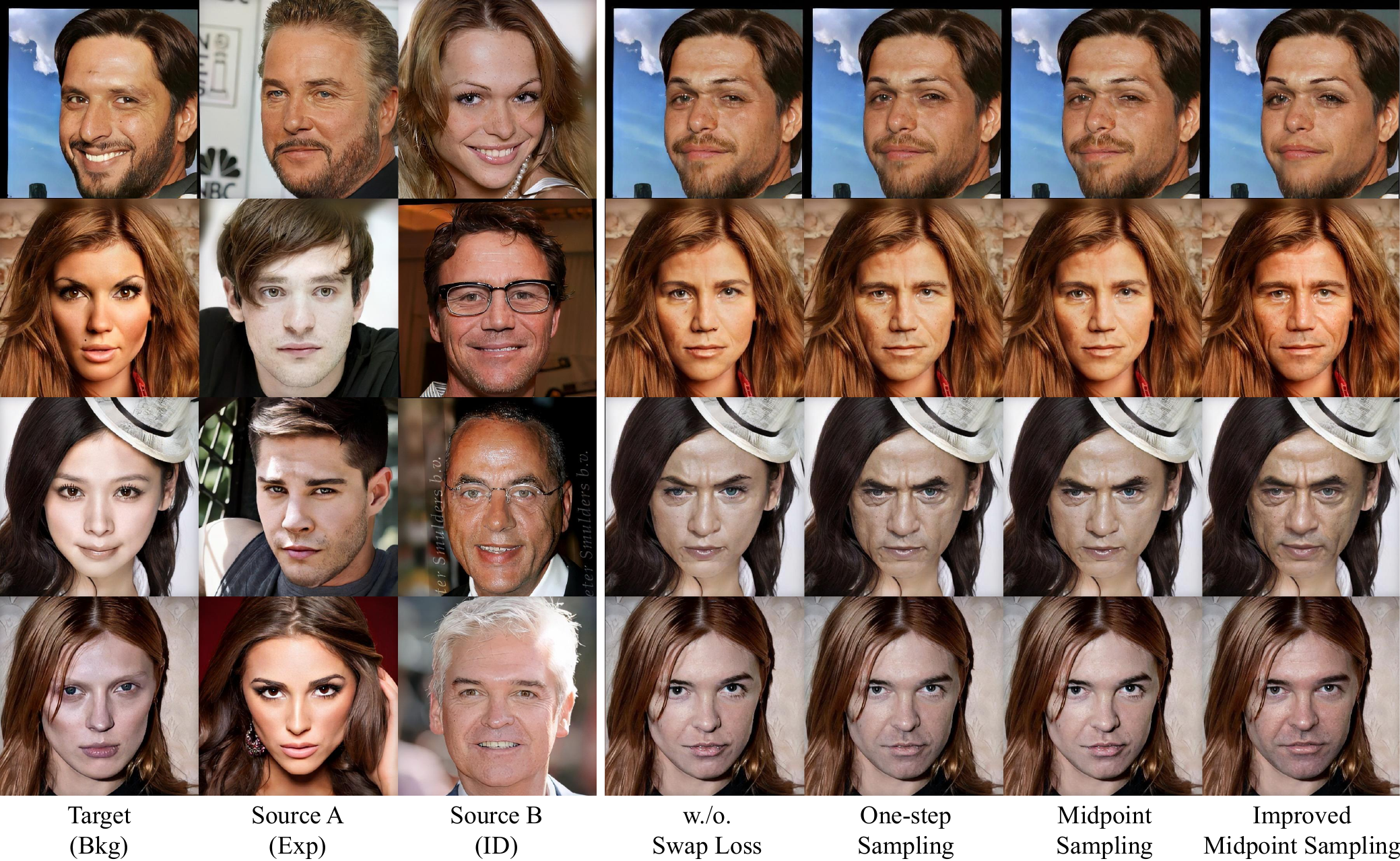} 
\vspace{-10pt}
\caption{SFSR results by using different sampling methods. }
\vspace{-2pt}
\label{fig:ablation1}
\end{figure}

Fig. \ref{fig:ablation1} shows the comparison in terms of generating final faces. In terms of identity preserving, our sampling methods can produce more facial details, e.g. wrinkles and whiskers, thus more similar to source B. In terms of expression preservation, our result in the 3rd row exhibits less angry expression, thus being more consistent with the expression in source A. In the 4th row, our result displays the expression of slightly opening mouth better than others. 

Tab. \ref{tab:ablation1} shows the quantitative comparison. When comparing ours to w./o. ID$\&$Exp Losses, the effectiveness of identity and expression losses has been validated by removing them. The improved midpoint sampling gets the best score in ID. and the second best score in Exp., which indicates it can impose more effective constraints on the training of diffusion model.

\begin{table}[]
\centering
\begin{tabular}{l|ll}
\hline
\multicolumn{1}{c|}{Methods}                  & \multicolumn{1}{c}{ID.$\uparrow$} & \multicolumn{1}{c}{Exp.$\downarrow$
                  } \\ \hline
w./o. ID$\&$Exp Losses   & 67.5                  & 0.71                   \\
One-step Sampling & 83.1                  & \textbf{0.62}                   \\
Midpoint Sampling \cite{DiffSwap2023} & 74.0                  & 0.70                   \\
Improved Midpoint Sampling (ours)          & \textbf{87.0}         & 0.63          \\ \hline
\end{tabular}
\vspace{-7pt}
\caption{Quantitative results of using different sampling methods. All values are
scaled up by a factor of 100 for simplicity.}
\vspace{-2pt}
\label{tab:ablation1}
\end{table}

\subsection{Ablation Study} \label{sec:ablation}
\begin{table}[]
\centering
\begin{tabular}{l|llll}
\hline
\multicolumn{1}{c|}{Methods} & \multicolumn{1}{c}{ID.$\uparrow$} & \multicolumn{1}{c}{Exp.$\downarrow$} & \multicolumn{1}{c}{Pose.$\downarrow$} & \multicolumn{1}{c}{MSE.$\downarrow$} \\ \hline
w./o. Bkg Condi.             & 81.5                   & 0.66                     & 8.97                     & 51.40                   \\
w./o. ID Emb.3               & 74.9                   & 0.63                     & 7.54                     & 52.09                   \\
w./o. ID Emb.2               & 37.9                   & \textbf{0.56}            & 7.90                     & 52.98                   \\
Full Model                         & \textbf{87.0}          & 0.63                     & \textbf{7.36}            & \textbf{51.23}          \\ \hline
\end{tabular}
\vspace{-7pt}
\caption{Quantitative result of ablation studies. The score in ID. and Exp. are scaled up by a factor of 100 for simplicity.}
\vspace{-9pt}
\label{tab:ablation2}
\end{table}

\begin{figure}
\centering
\includegraphics[width=0.48\textwidth]{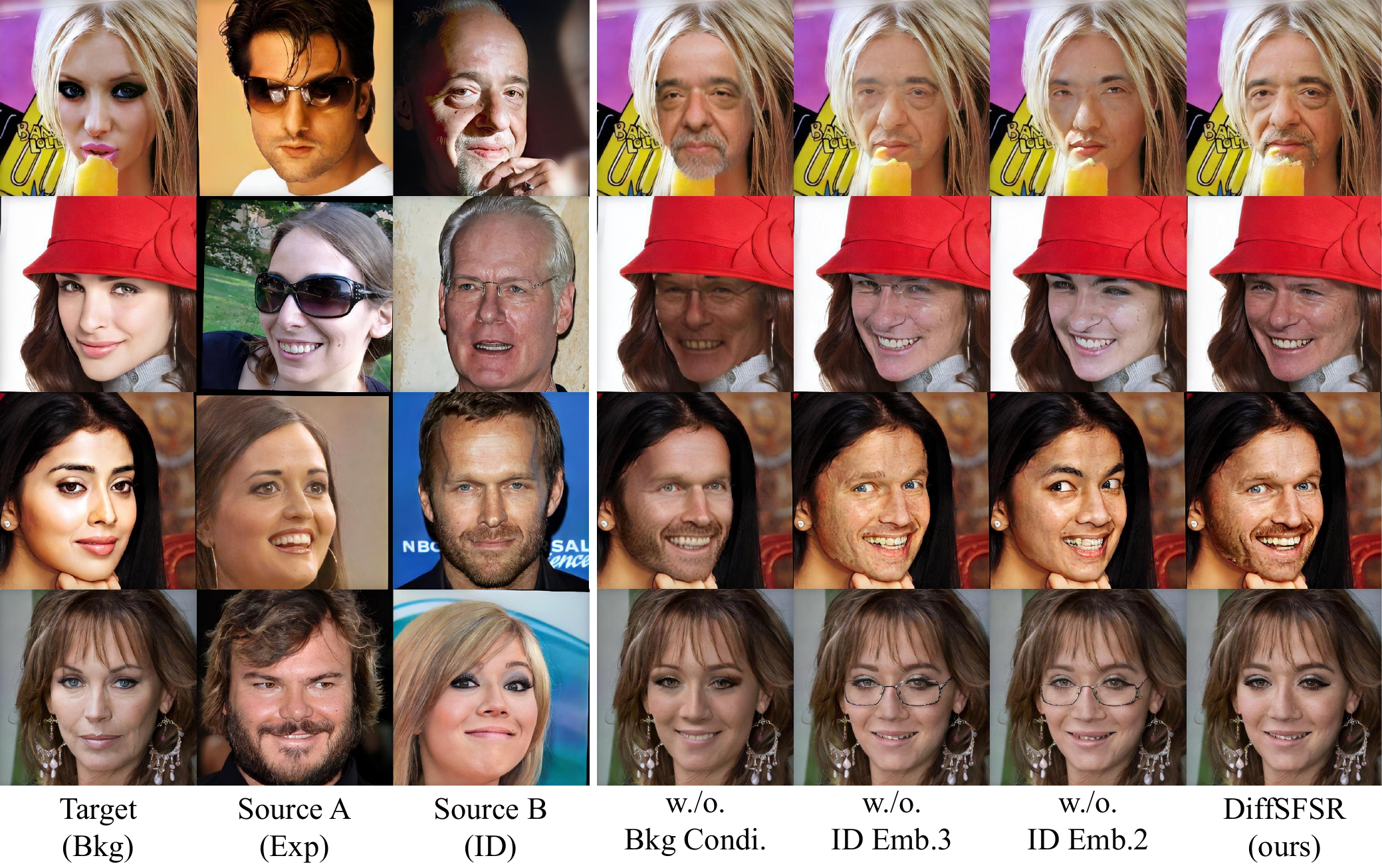} 
\vspace{-22pt}
\caption{Qualitative results in ablation study.}
\vspace{-12pt}
\label{fig:ablation2}
\end{figure}

We conduct ablation studies to demonstrate the effectiveness of the background conditioning and the compound identity embedding, by removing them individually during training. In the method without background conditioning, we compensate with a segmentation map to specify face region in training and provide background only in inference. Swapping task (mismatch conditions) and reconstruction task (match conditions) are performed in CelebA-HQ. 

\textbf{Effect of background conditioning.} 
Without background conditioning, the diffusion model can't learn the accurate lighting and face pose, and can't generate faces of higher image quality and being consistent with the background. These arguments are well supported by the results shown in Fig. \ref{fig:ablation2}. The generated faces without background conditioning suffer from inaccurate lighting, exhibit more face pose errors, lack seamless blending with the background, and are comparatively more blurry.

\textbf{Effect of compound identity embedding.} 
Compound identity embedding can significantly improve identity similarity. 
As shown in the first, second and third rows in Fig. \ref{fig:ablation2}, each time we remove an identity embedding, the generated faces become more similar to source A in terms of identity, and less similar to source B which provides the expected identity. This phenomenon indicates that the residual identity attribute in the expression embedding will be unexpectedly transferred to the result when the identity embedding is weaker than the expression embedding.
Notably in the fourth row, the inadequate identity representation even incorrectly puts glasses on the results. As shown in Tab. \ref{tab:ablation2}, the identity similarity in our results surpasses all ablation methods. 

\section{Conclusion}
Given a text prompt, an expression label and a selfie photo, our personalized face generation framework can produce high-fidelity and identity-expression preserving portraits. To realize the framework, we propose a new diffusion model that can conduct simultaneous face swapping and reenactment task. Extensive experiments have demonstrated the controllability and scalability of the proposed framework. We hope our efforts can inspire future work in personalized generation framework
to explore the use of more modalities as conditioning to achieve higher controllability and image quality.

\section{Limitations} As mentioned above, although the facial expressions between reference and synthesized images are close to each other in our framework, it can be found that their facial expressions could not fully reflect the semantic information by the text label. For example, the text label ``agitation" is not consistent with the reference image in the 4th row of Fig. \ref{fig:p1}. This can be attributed to the flaws from the dataset \cite{EmoFace135} that cannot guarantee that all images can fully display their corresponding expression labels. Additionally, these expression labels have more or less ambiguity among them, leading to the overlapping of their semantics.

\section{Appendix}
\label{sec:allresults}
\begin{figure*}
\centering
\includegraphics[width=0.95\textwidth]{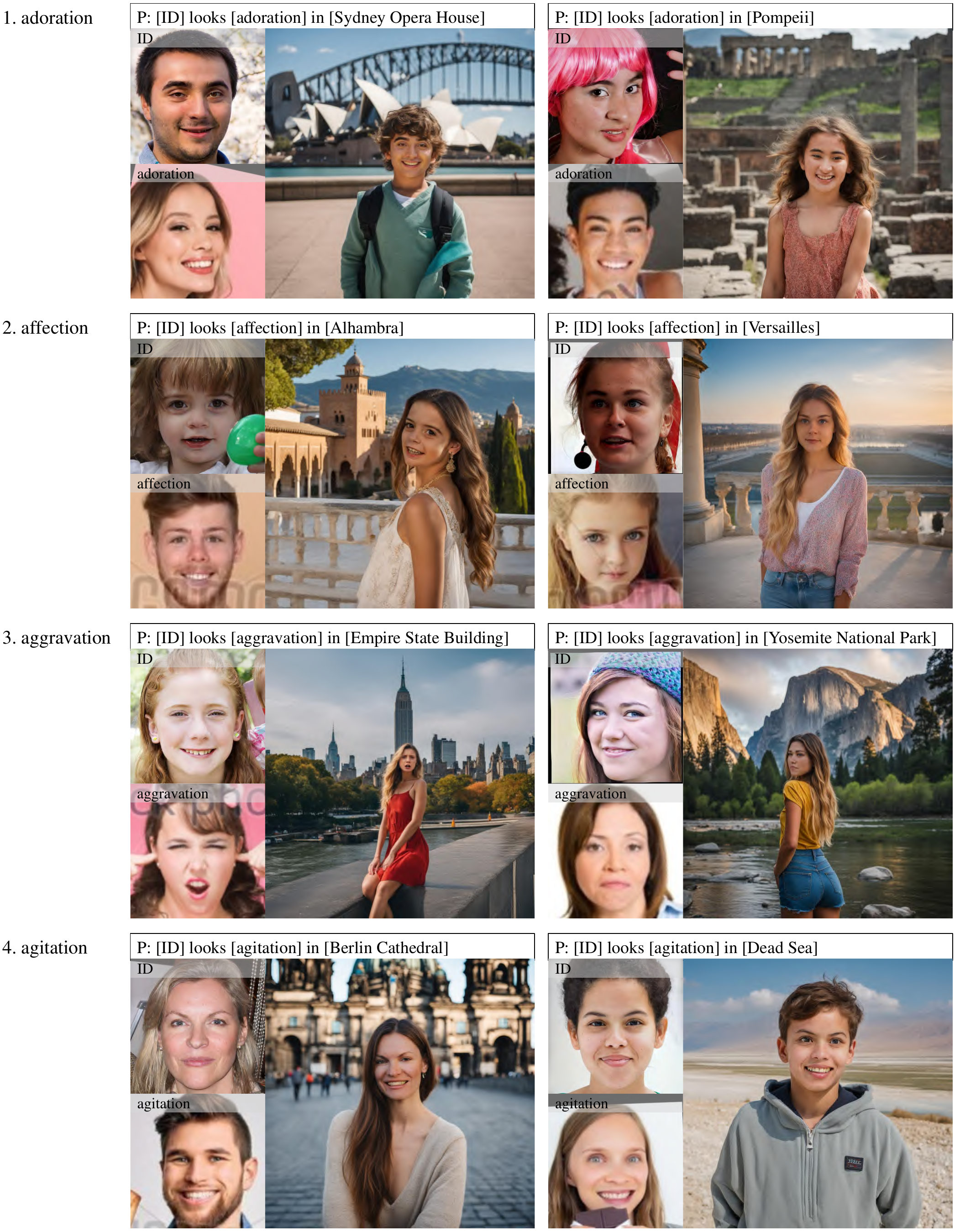} 
\vspace{-10pt}
\caption{Resulting samples of the full set of 135 expression labels. The input text prompt is shown at the top. The image in the top right corner refers to the ID image and the image in the bottom right corner refers to the expression reference image. The image on the right showcases the resulting image according to the inputs of the text prompt and ID image. Please zoom in for more details.}
\label{fig:p1}
\end{figure*}

\begin{figure*}
\centering
\includegraphics[width=0.95\textwidth]{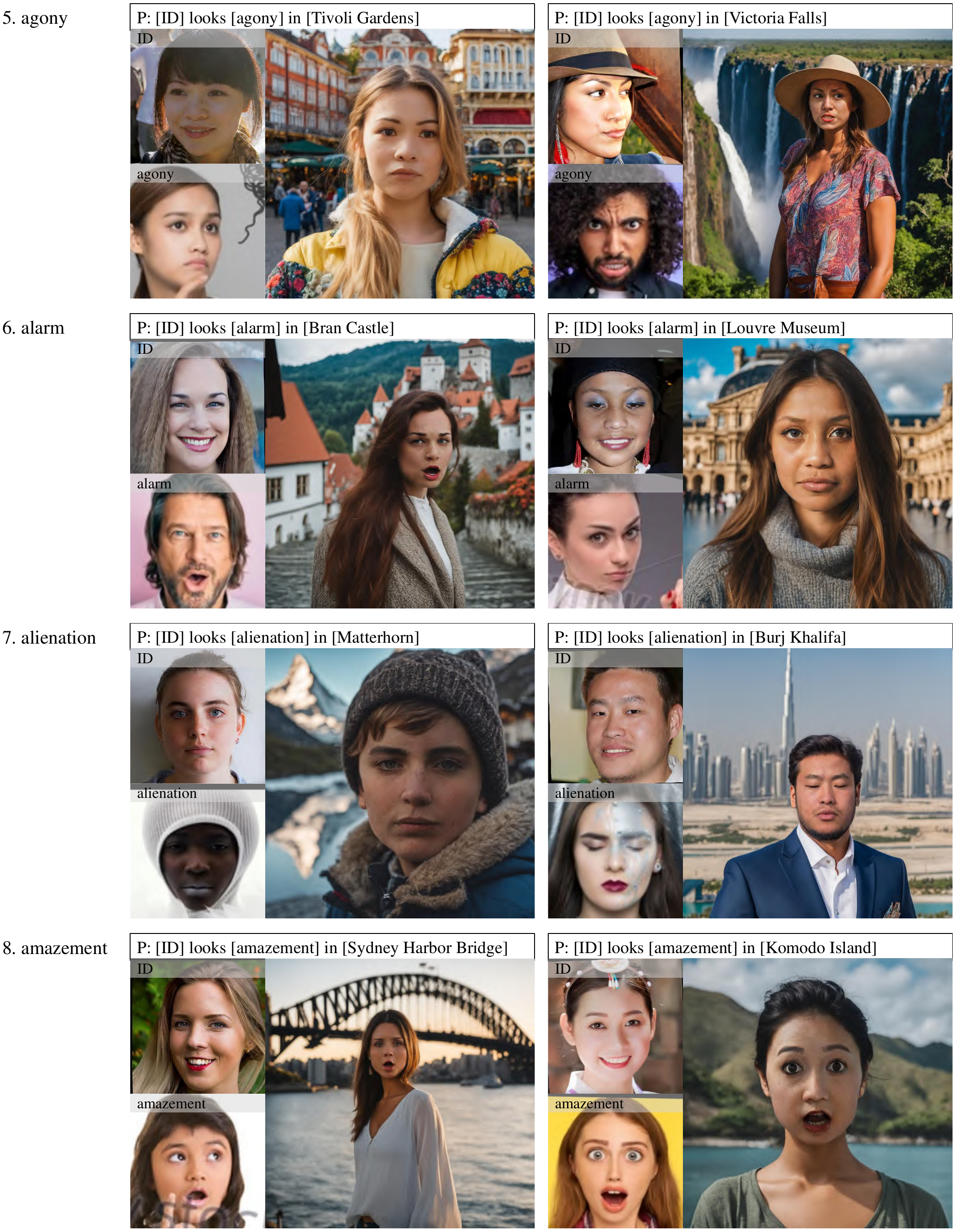} 
\vspace{-10pt}
\caption{Continues from Figures \ref{fig:p1}. The input text prompt is shown at the top. The image in the top right corner refers to the ID image and the image in the bottom right corner refers to the expression reference image. The image on the right showcases the resulting image according to the inputs of the text prompt and ID image. Please zoom in for more details.}
\label{fig:p2}
\end{figure*}

\begin{figure*}
\centering
\includegraphics[width=0.95\textwidth]{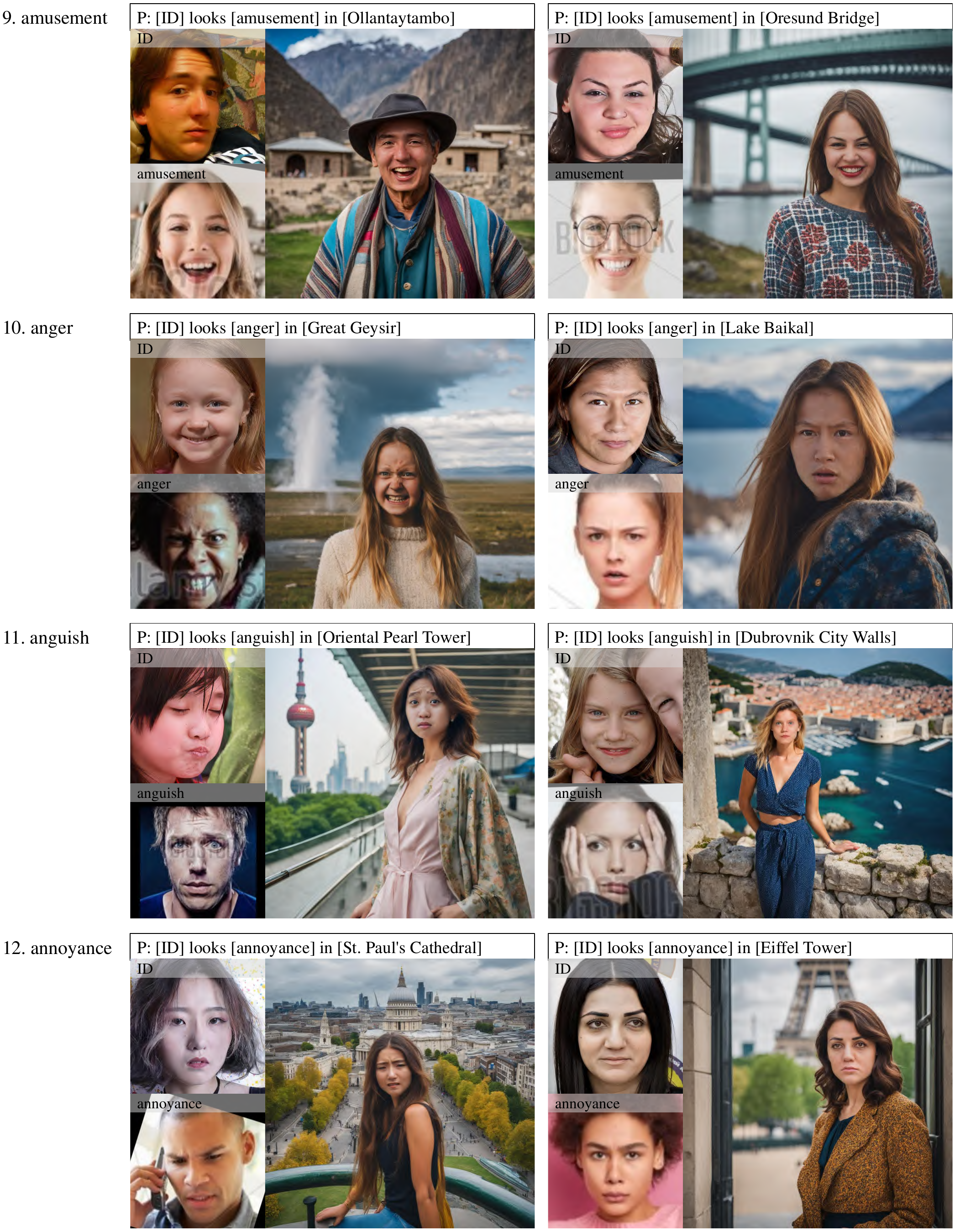} 
\vspace{-10pt}
\caption{Continues from Figures \ref{fig:p1}-\ref{fig:p2}. The input text prompt is shown at the top. The image in the top right corner refers to the ID image and the image in the bottom right corner refers to the expression reference image. The image on the right showcases the resulting image according to the inputs of the text prompt and ID image. Please zoom in for more details.}
\label{fig:p3}
\end{figure*}

\begin{figure*}
\centering
\includegraphics[width=0.95\textwidth]{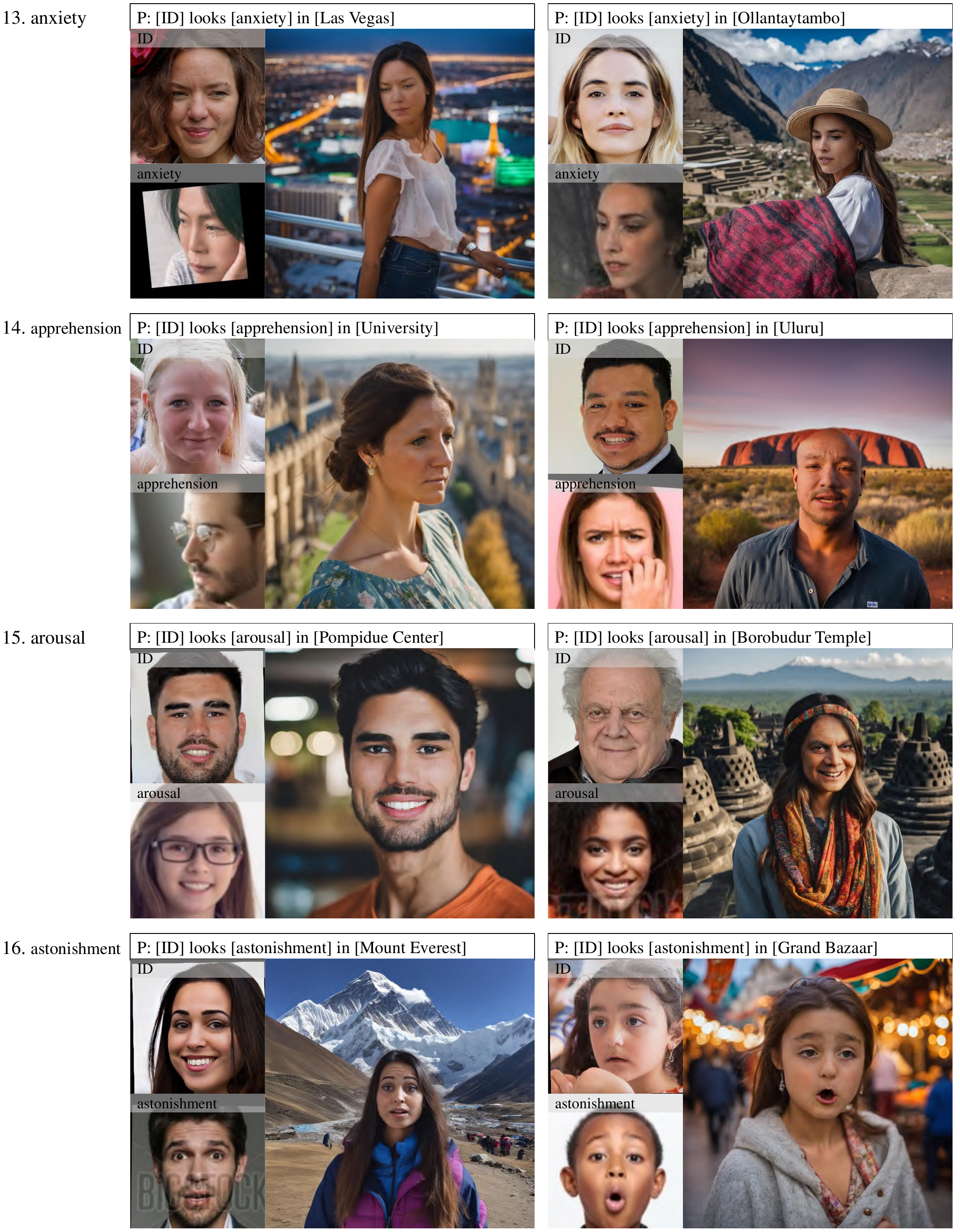} 
\vspace{-10pt}
\caption{Continues from Figures \ref{fig:p1}-\ref{fig:p3}. The input text prompt is shown at the top. The image in the top right corner refers to the ID image and the image in the bottom right corner refers to the expression reference image. The image on the right showcases the resulting image according to the inputs of the text prompt and ID image. Please zoom in for more details.}
\label{fig:p4}
\end{figure*}

\begin{figure*}
\centering
\includegraphics[width=0.95\textwidth]{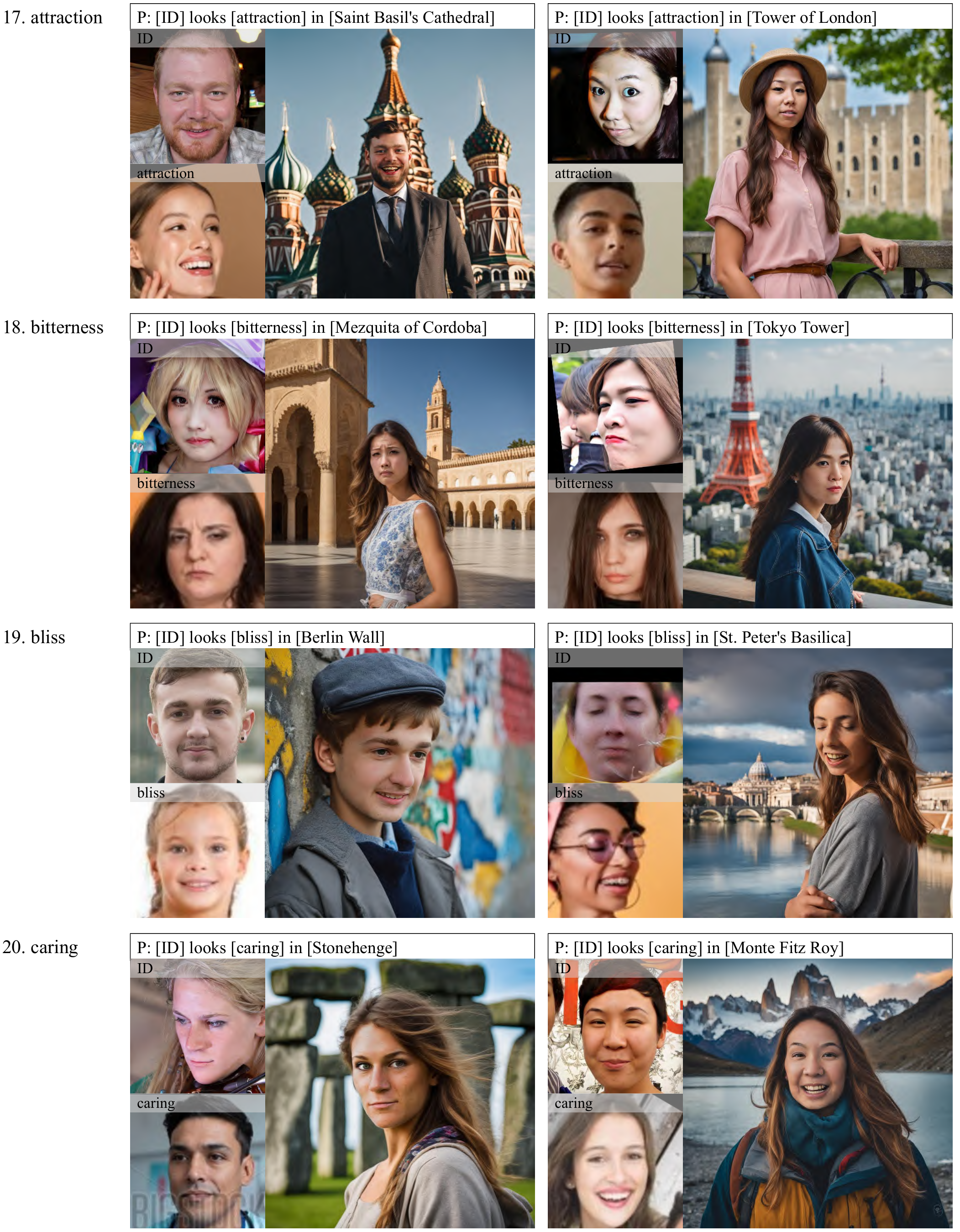} 
\vspace{-10pt}
\caption{Continues from Figures \ref{fig:p1}-\ref{fig:p4}. The input text prompt is shown at the top. The image in the top right corner refers to the ID image and the image in the bottom right corner refers to the expression reference image. The image on the right showcases the resulting image according to the inputs of the text prompt and ID image. Please zoom in for more details.}
\label{fig:p5}
\end{figure*}

\begin{figure*}
\centering
\includegraphics[width=0.95\textwidth]{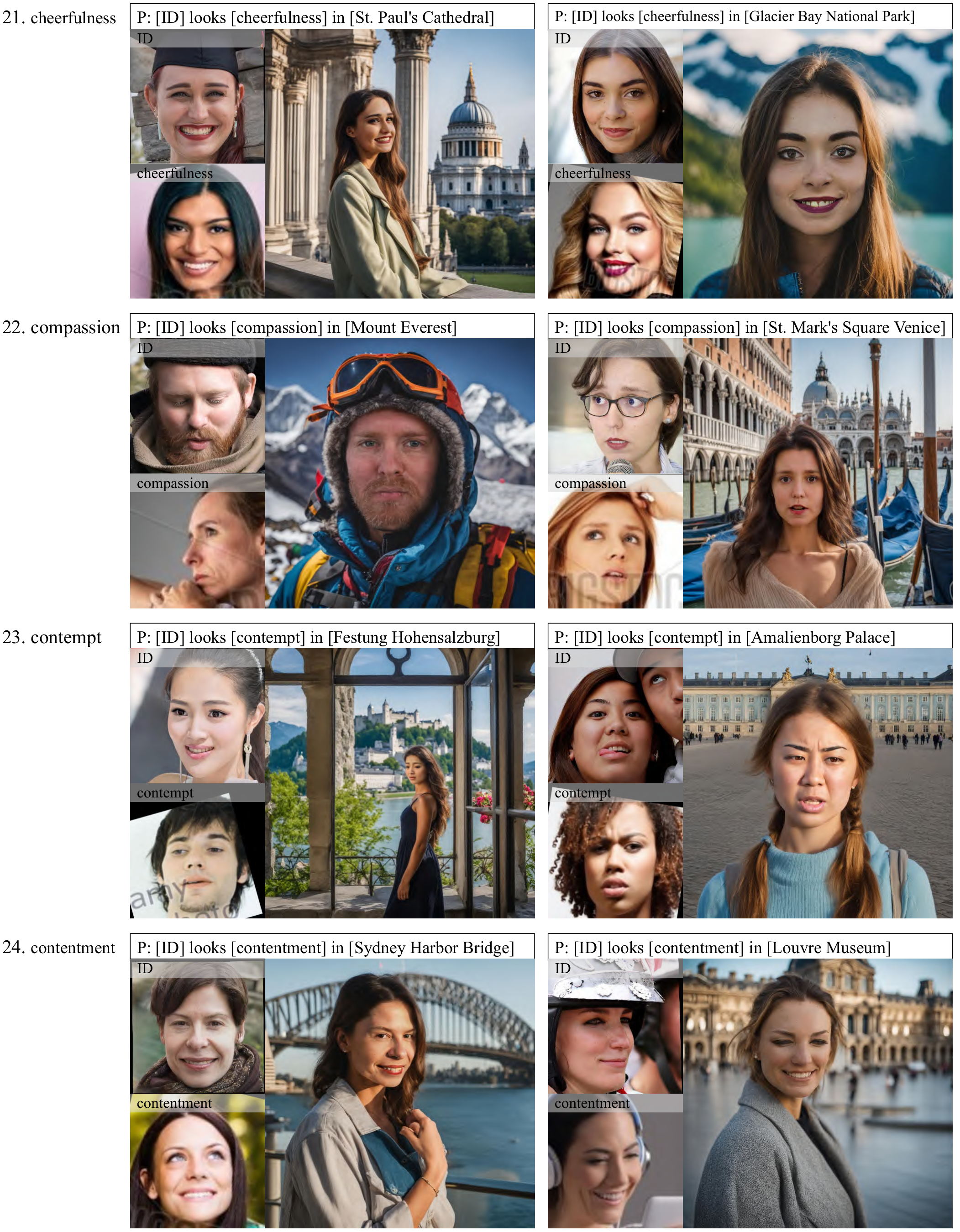} 
\vspace{-10pt}
\caption{Continues from Figures \ref{fig:p1}-\ref{fig:p5}. The input text prompt is shown at the top. The image in the top right corner refers to the ID image and the image in the bottom right corner refers to the expression reference image. The image on the right showcases the resulting image according to the inputs of the text prompt and ID image. Please zoom in for more details.}
\label{fig:p6}
\end{figure*}

\begin{figure*}
\centering
\includegraphics[width=0.95\textwidth]{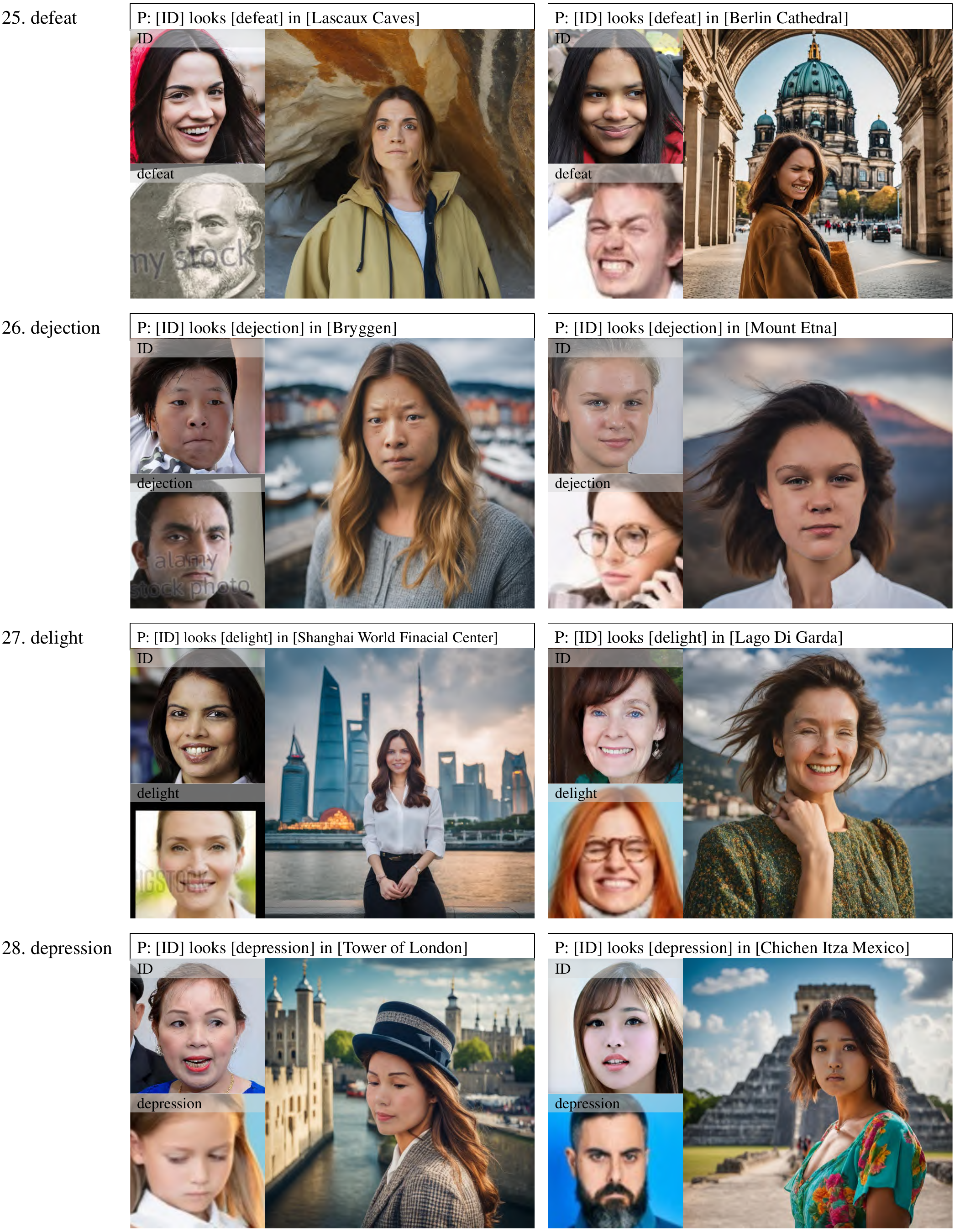} 
\vspace{-10pt}
\caption{Continues from Figures \ref{fig:p1}-\ref{fig:p6}. The input text prompt is shown at the top. The image in the top right corner refers to the ID image and the image in the bottom right corner refers to the expression reference image. The image on the right showcases the resulting image according to the inputs of the text prompt and ID image. Please zoom in for more details.}
\label{fig:p7}
\end{figure*}

\begin{figure*}
\centering
\includegraphics[width=0.95\textwidth]{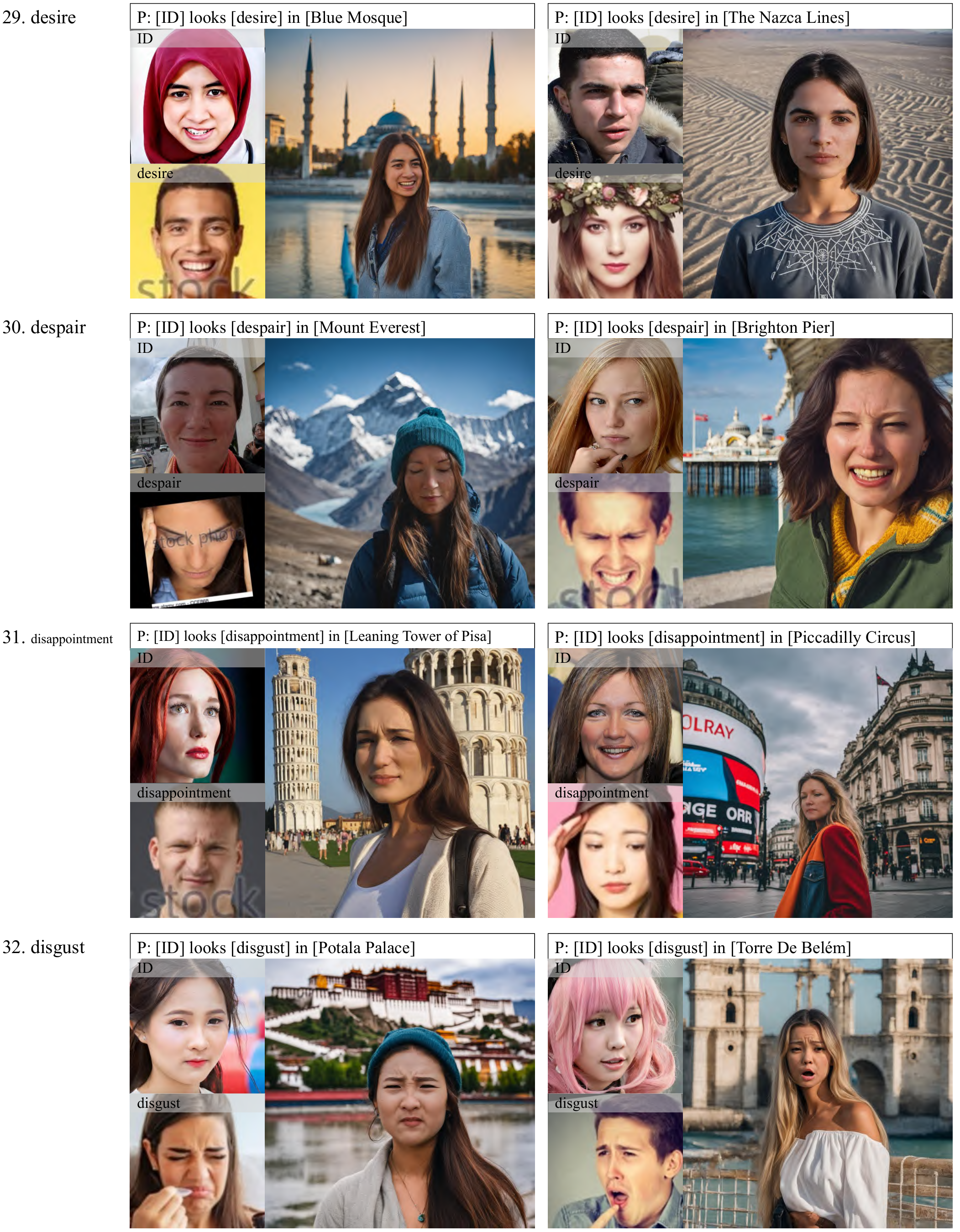} 
\vspace{-10pt}
\caption{Continues from Figures \ref{fig:p1}-\ref{fig:p7}. The input text prompt is shown at the top. The image in the top right corner refers to the ID image and the image in the bottom right corner refers to the expression reference image. The image on the right showcases the resulting image according to the inputs of the text prompt and ID image. Please zoom in for more details.}
\label{fig:p8}
\end{figure*}

\begin{figure*}
\centering
\includegraphics[width=0.95\textwidth]{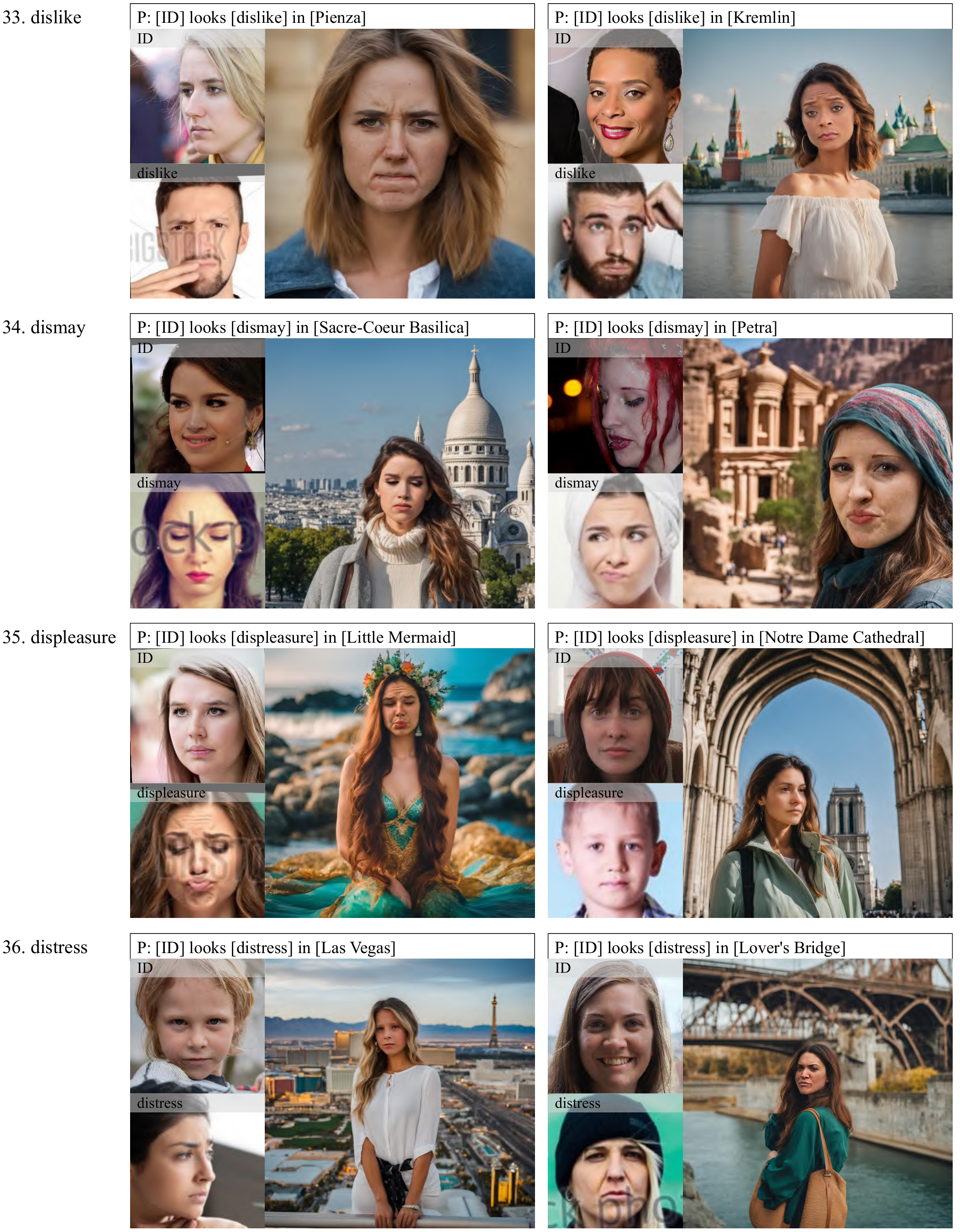} 
\vspace{-10pt}
\caption{Continues from Figures \ref{fig:p1}-\ref{fig:p8}. The input text prompt is shown at the top. The image in the top right corner refers to the ID image and the image in the bottom right corner refers to the expression reference image. The image on the right showcases the resulting image according to the inputs of the text prompt and ID image. Please zoom in for more details.}
\label{fig:p9}
\end{figure*}

\begin{figure*}
\centering
\includegraphics[width=0.95\textwidth]{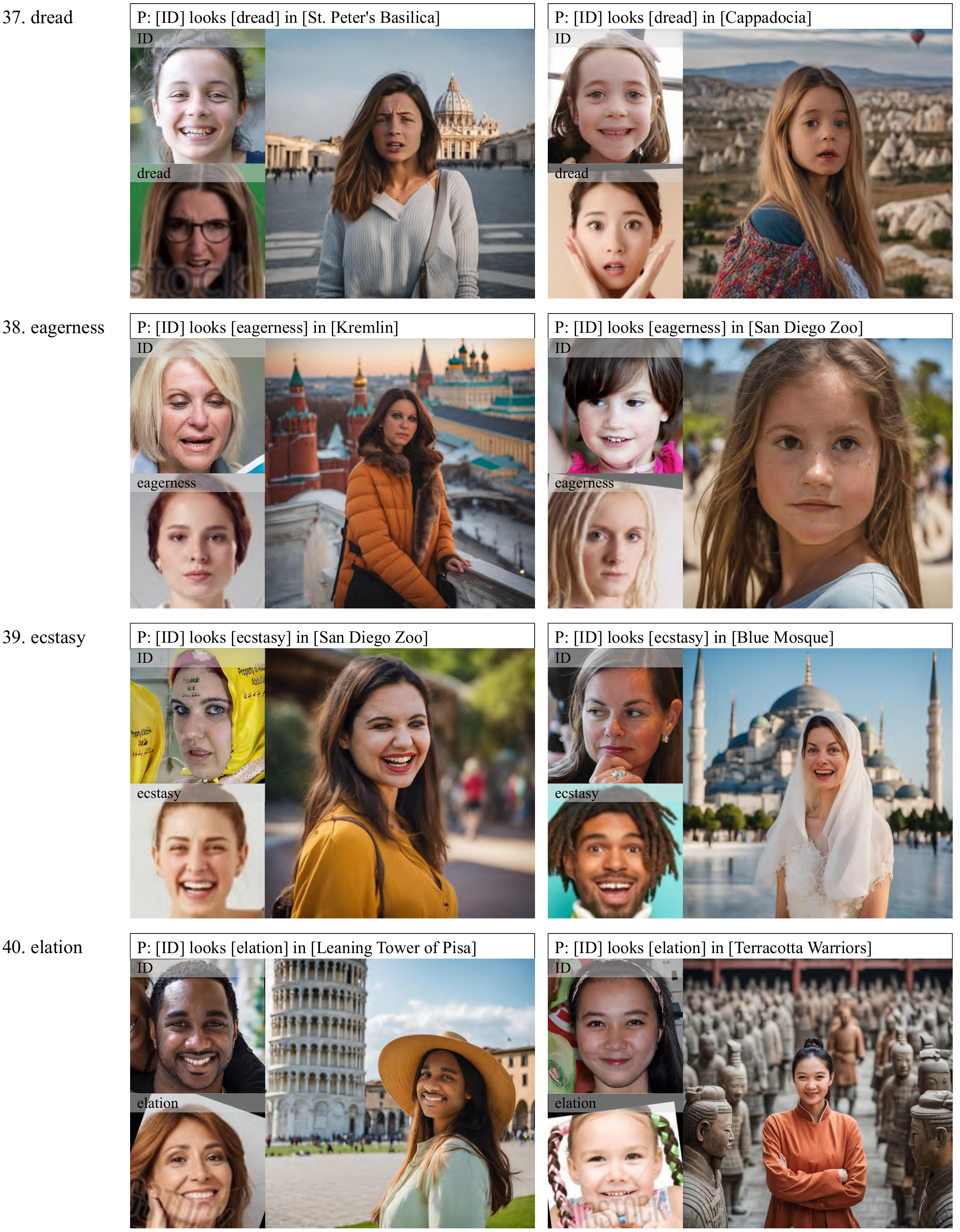} 
\vspace{-10pt}
\caption{Continues from Figures \ref{fig:p1}-\ref{fig:p9}. The input text prompt is shown at the top. The image in the top right corner refers to the ID image and the image in the bottom right corner refers to the expression reference image. The image on the right showcases the resulting image according to the inputs of the text prompt and ID image. Please zoom in for more details.}
\label{fig:p10}
\end{figure*}

\begin{figure*}
\centering
\includegraphics[width=0.95\textwidth]{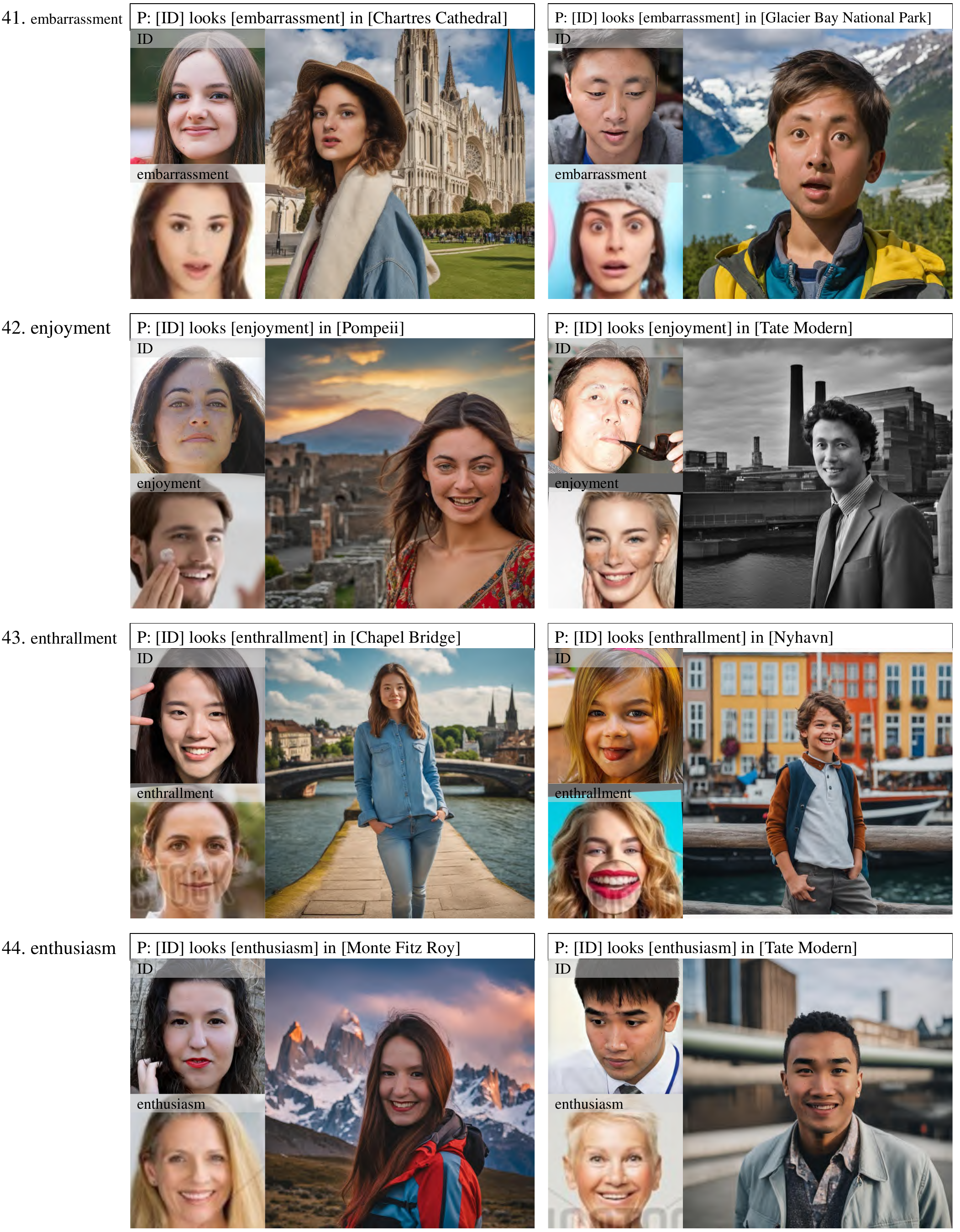} 
\vspace{-10pt}
\caption{Continues from Figures \ref{fig:p1}-\ref{fig:p10}. The input text prompt is shown at the top. The image in the top right corner refers to the ID image and the image in the bottom right corner refers to the expression reference image. The image on the right showcases the resulting image according to the inputs of the text prompt and ID image. Please zoom in for more details.}
\label{fig:p11}
\end{figure*}

\begin{figure*}
\centering
\includegraphics[width=0.95\textwidth]{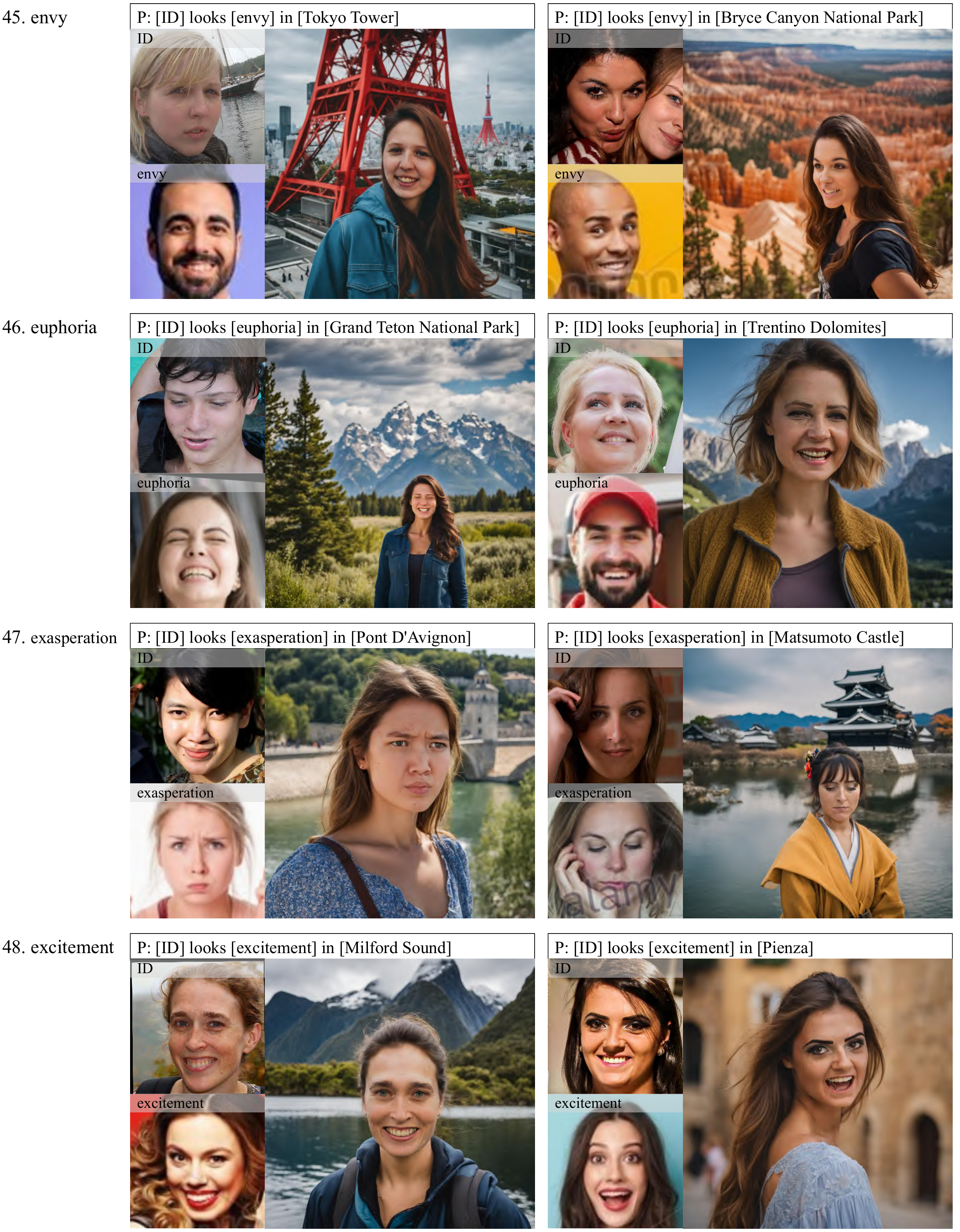} 
\vspace{-10pt}
\caption{Continues from Figures \ref{fig:p1}-\ref{fig:p11}. The input text prompt is shown at the top. The image in the top right corner refers to the ID image and the image in the bottom right corner refers to the expression reference image. The image on the right showcases the resulting image according to the inputs of the text prompt and ID image. Please zoom in for more details.}
\label{fig:p12}
\end{figure*}

\begin{figure*}
\centering
\includegraphics[width=0.95\textwidth]{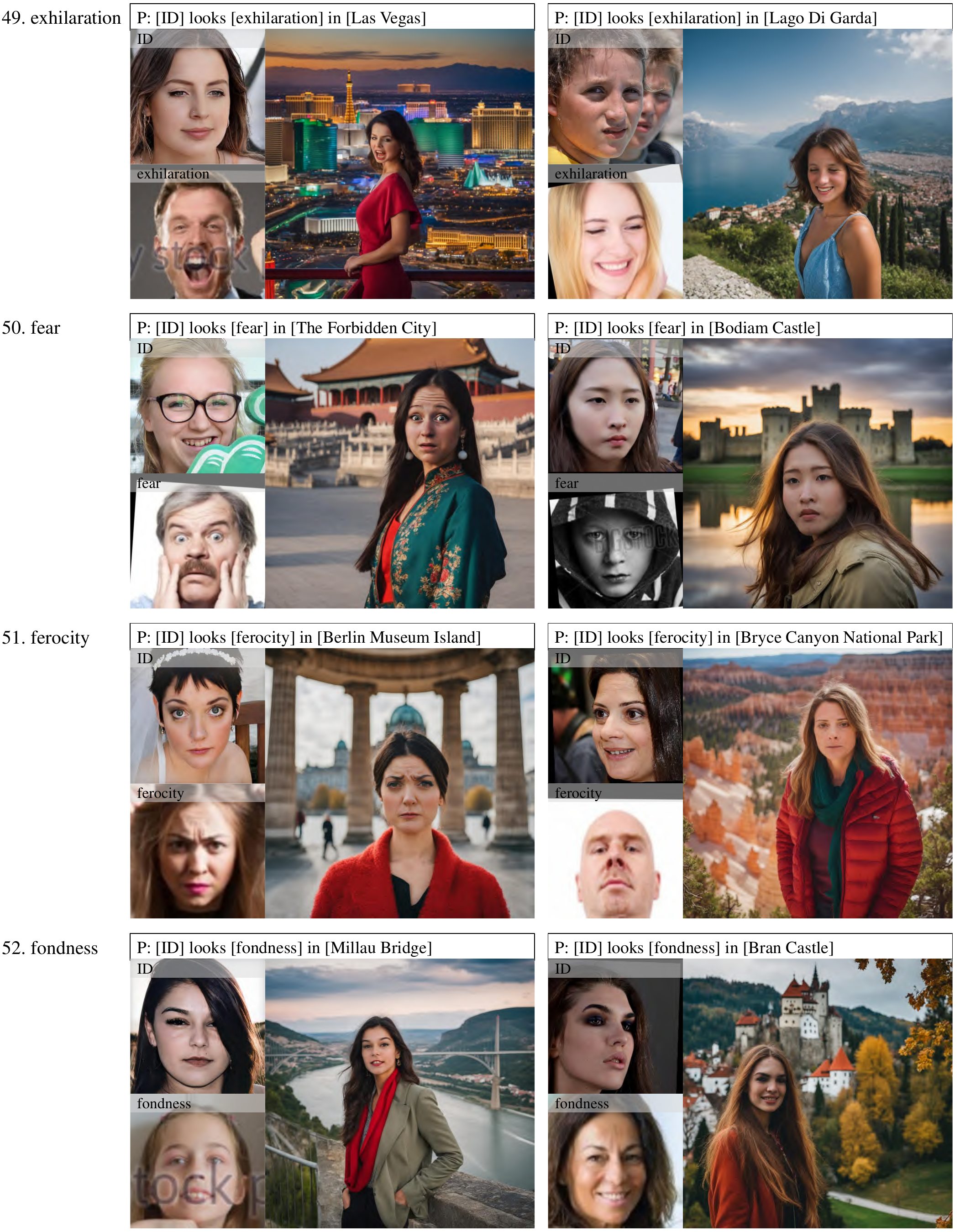} 
\vspace{-10pt}
\caption{Continues from Figures \ref{fig:p1}-\ref{fig:p12}. The input text prompt is shown at the top. The image in the top right corner refers to the ID image and the image in the bottom right corner refers to the expression reference image. The image on the right showcases the resulting image according to the inputs of the text prompt and ID image. Please zoom in for more details.}
\label{fig:p13}
\end{figure*}

\begin{figure*}
\centering
\includegraphics[width=0.95\textwidth]{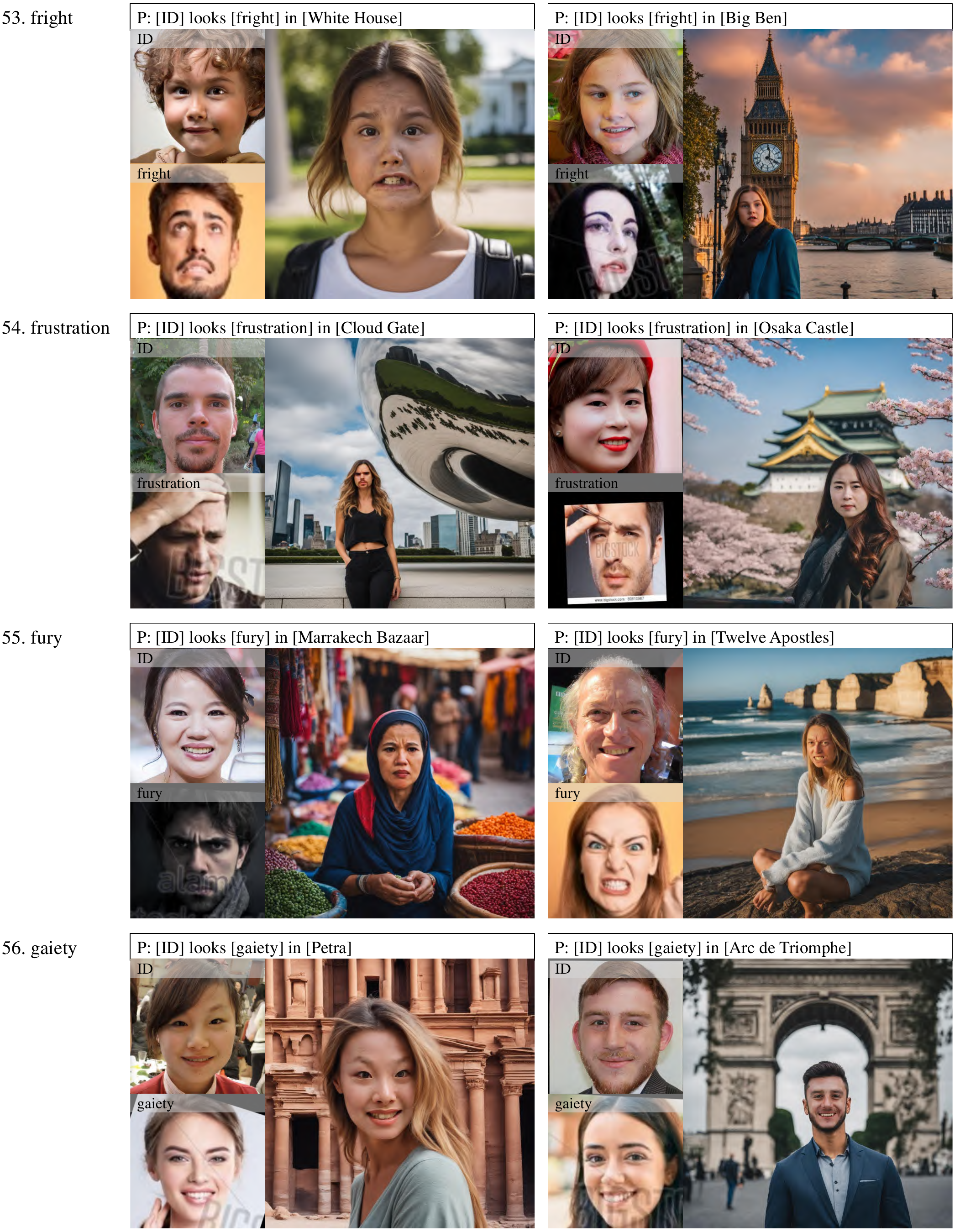} 
\vspace{-10pt}
\caption{Continues from Figures \ref{fig:p1}-\ref{fig:p13}. The input text prompt is shown at the top. The image in the top right corner refers to the ID image and the image in the bottom right corner refers to the expression reference image. The image on the right showcases the resulting image according to the inputs of the text prompt and ID image. Please zoom in for more details.}
\label{fig:p14}
\end{figure*}

\begin{figure*}
\centering
\includegraphics[width=0.95\textwidth]{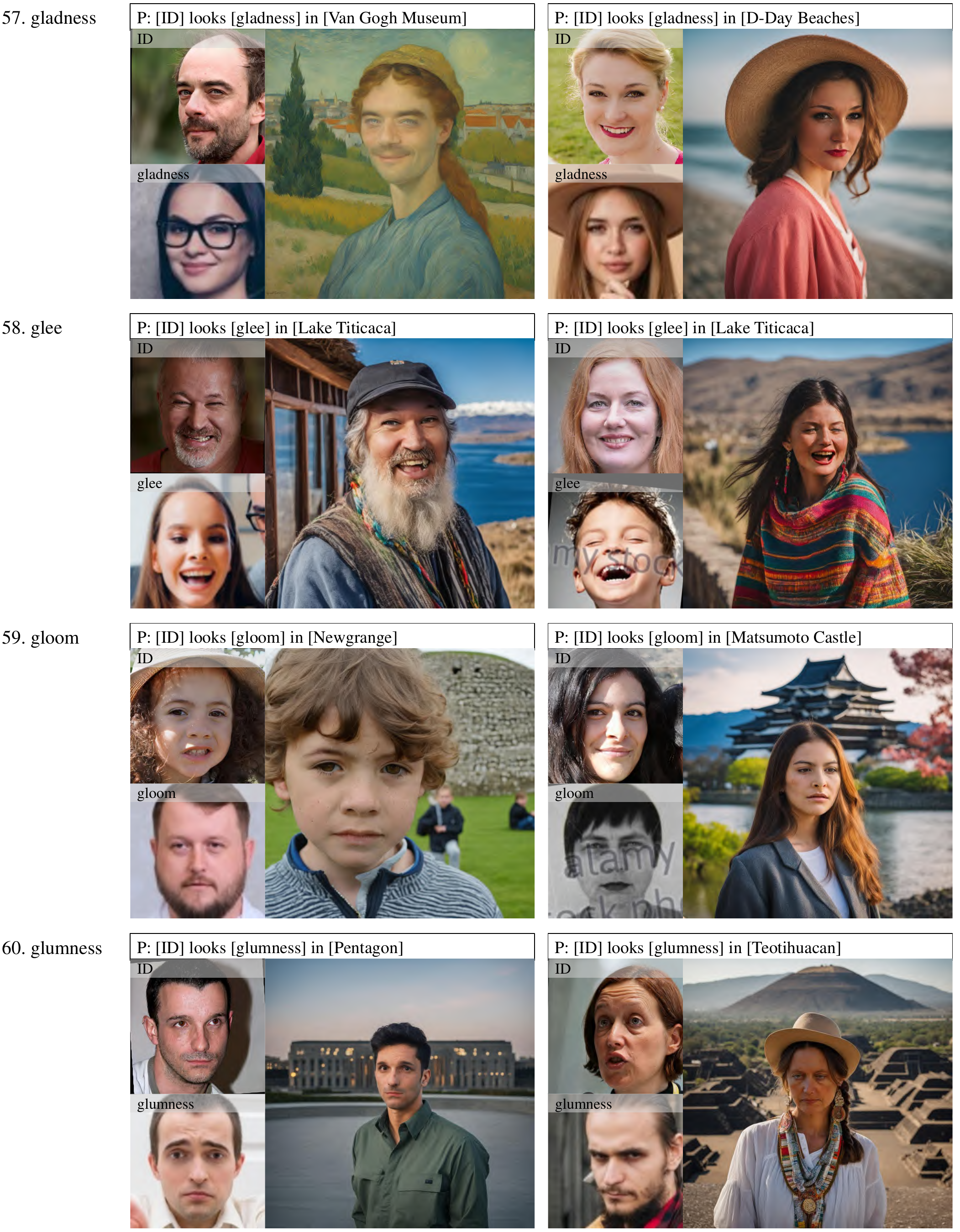} 
\vspace{-10pt}
\caption{Continues from Figures \ref{fig:p1}-\ref{fig:p14}. The input text prompt is shown at the top. The image in the top right corner refers to the ID image and the image in the bottom right corner refers to the expression reference image. The image on the right showcases the resulting image according to the inputs of the text prompt and ID image. Please zoom in for more details.}
\label{fig:p15}
\end{figure*}

\begin{figure*}
\centering
\includegraphics[width=0.95\textwidth]{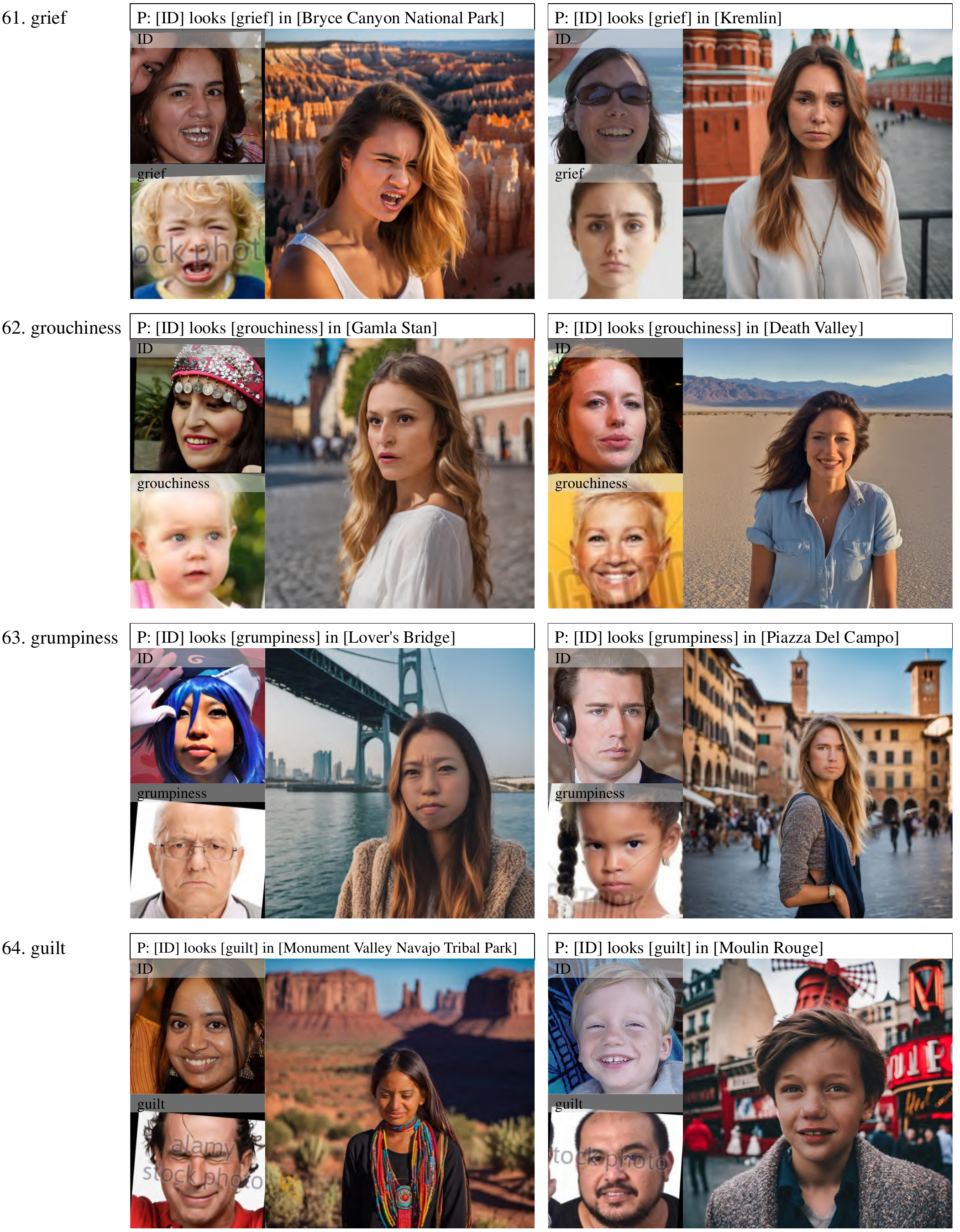} 
\vspace{-10pt}
\caption{Continues from Figures \ref{fig:p1}-\ref{fig:p15}. The input text prompt is shown at the top. The image in the top right corner refers to the ID image and the image in the bottom right corner refers to the expression reference image. The image on the right showcases the resulting image according to the inputs of the text prompt and ID image. Please zoom in for more details.}
\label{fig:p16}
\end{figure*}

\begin{figure*}
\centering
\includegraphics[width=0.95\textwidth]{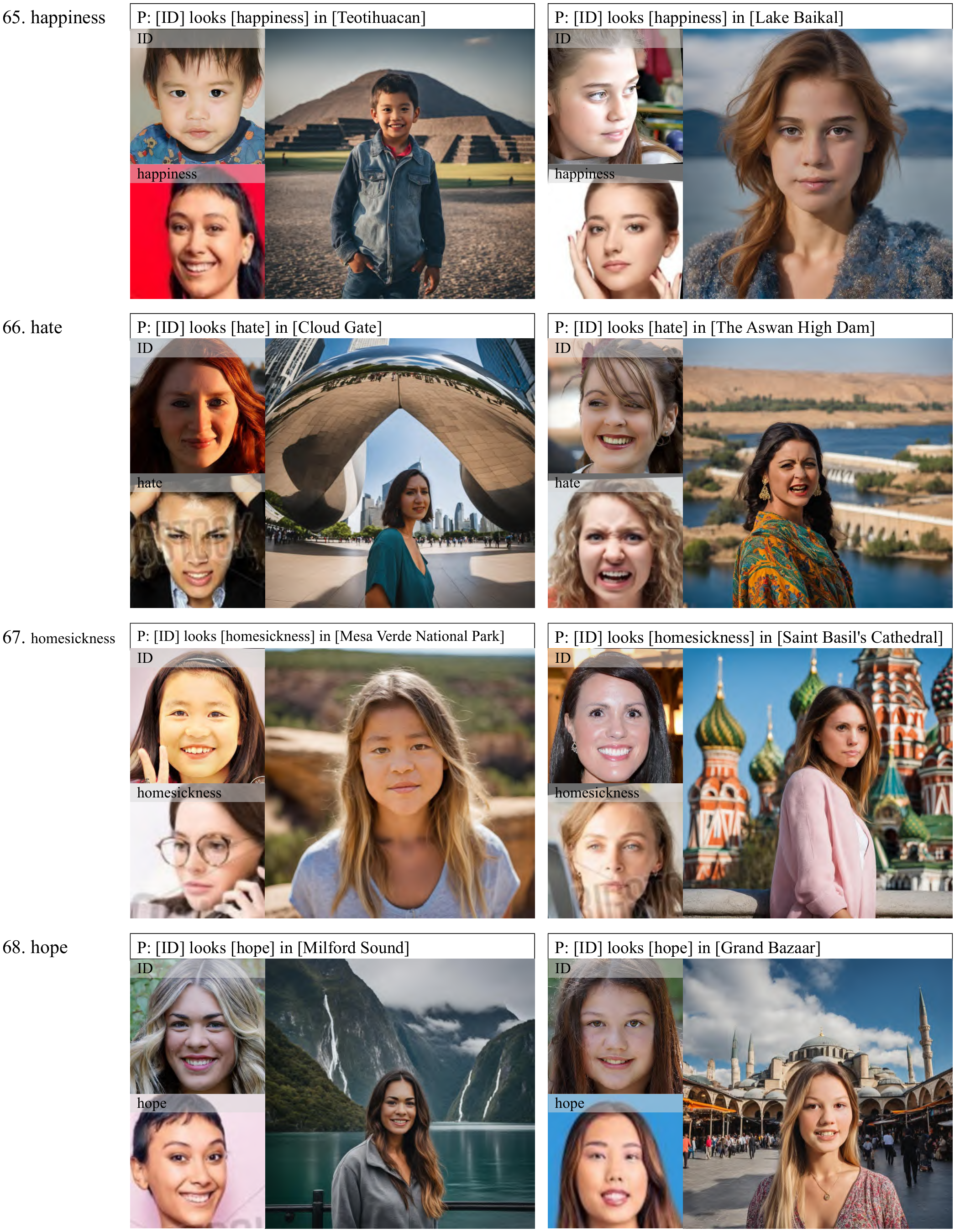} 
\vspace{-10pt}
\caption{Continues from Figures \ref{fig:p1}-\ref{fig:p16}. The input text prompt is shown at the top. The image in the top right corner refers to the ID image and the image in the bottom right corner refers to the expression reference image. The image on the right showcases the resulting image according to the inputs of the text prompt and ID image. Please zoom in for more details.}
\label{fig:p17}
\end{figure*}

\begin{figure*}
\centering
\includegraphics[width=0.95\textwidth]{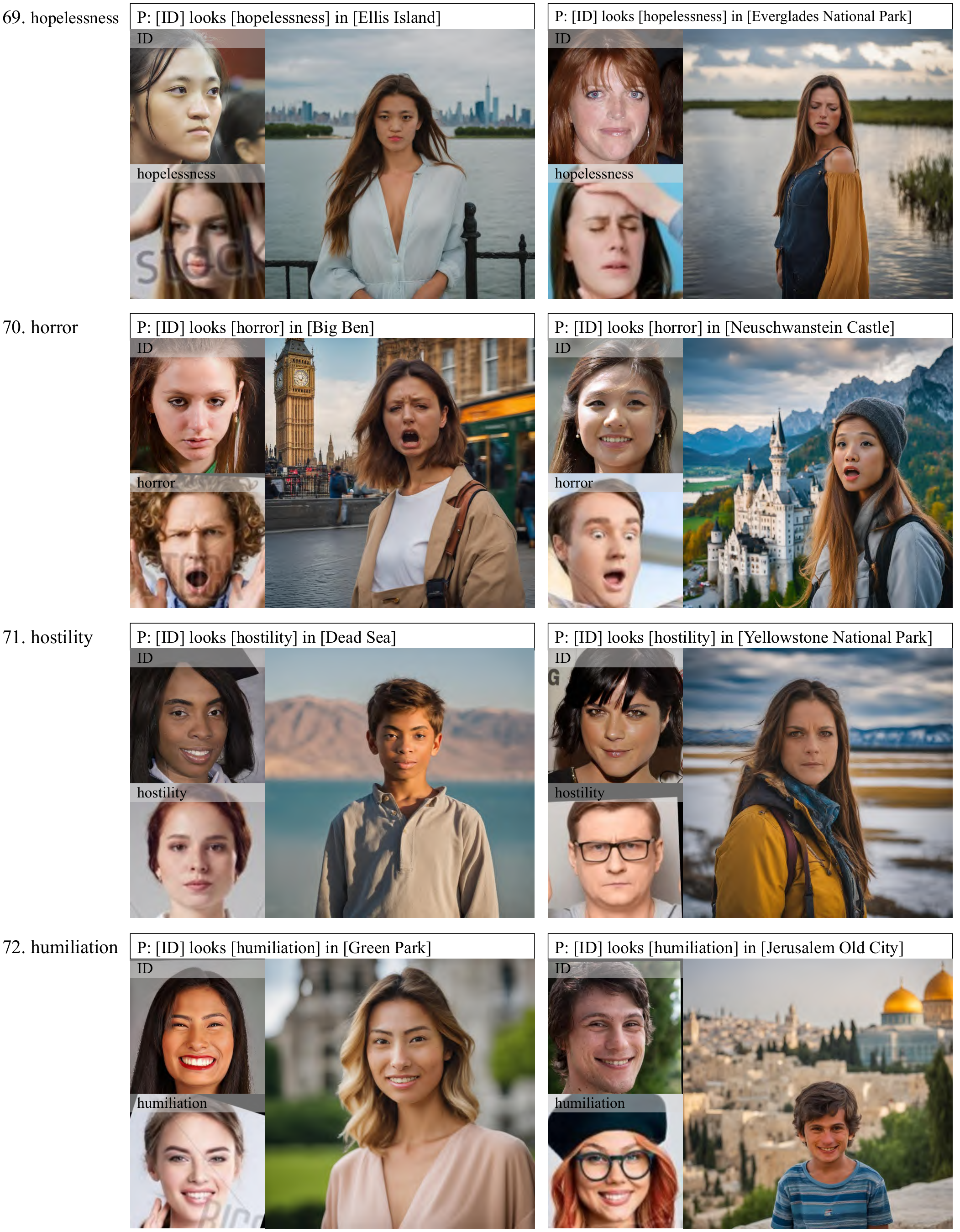} 
\vspace{-10pt}
\caption{Continues from Figures \ref{fig:p1}-\ref{fig:p17}. The input text prompt is shown at the top. The image in the top right corner refers to the ID image and the image in the bottom right corner refers to the expression reference image. The image on the right showcases the resulting image according to the inputs of the text prompt and ID image. Please zoom in for more details.}
\label{fig:p18}
\end{figure*}

\begin{figure*}
\centering
\includegraphics[width=0.95\textwidth]{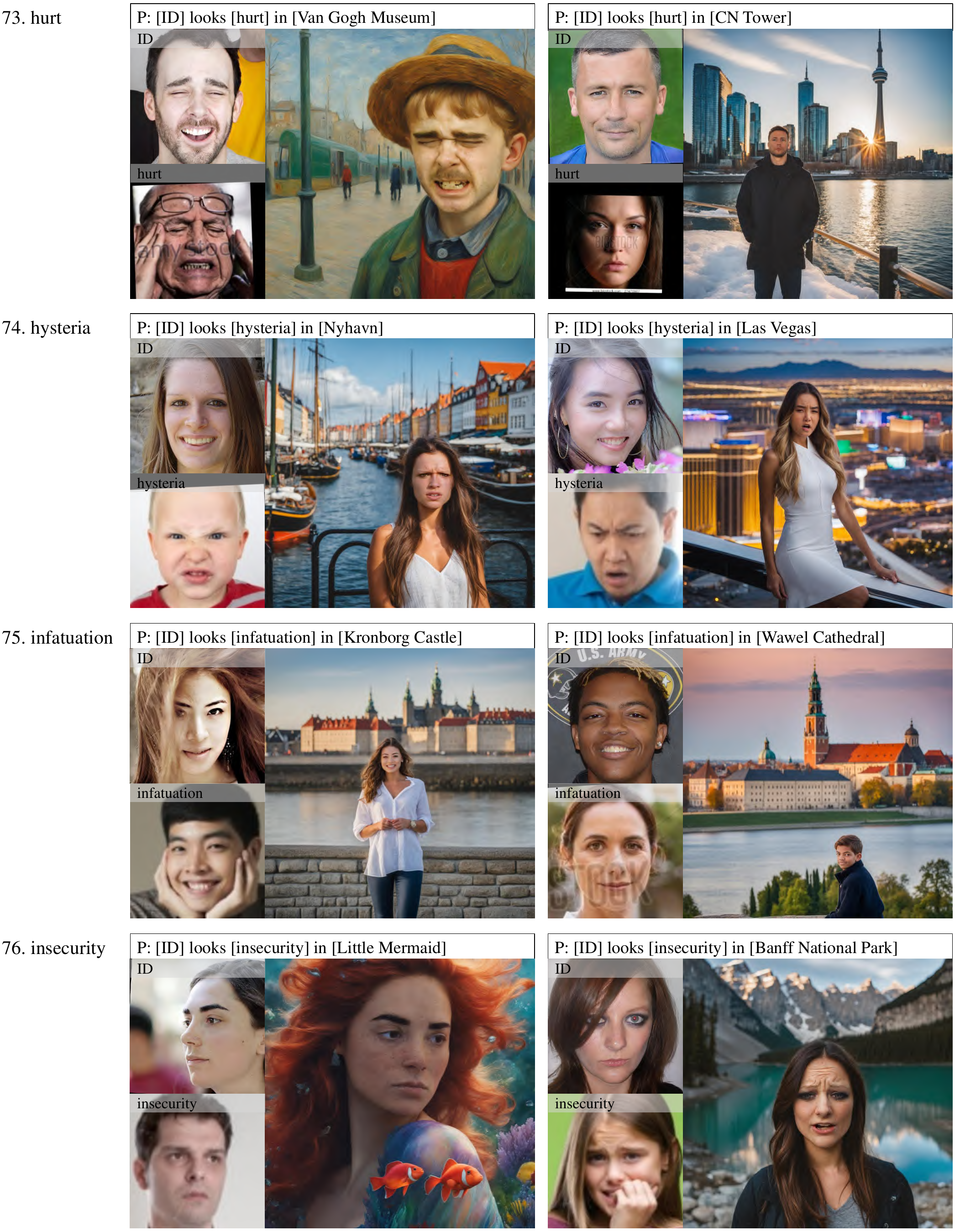} 
\vspace{-10pt}
\caption{Continues from Figures \ref{fig:p1}-\ref{fig:p18}. The input text prompt is shown at the top. The image in the top right corner refers to the ID image and the image in the bottom right corner refers to the expression reference image. The image on the right showcases the resulting image according to the inputs of the text prompt and ID image. Please zoom in for more details.}
\label{fig:p19}
\end{figure*}

\begin{figure*}
\centering
\includegraphics[width=0.95\textwidth]{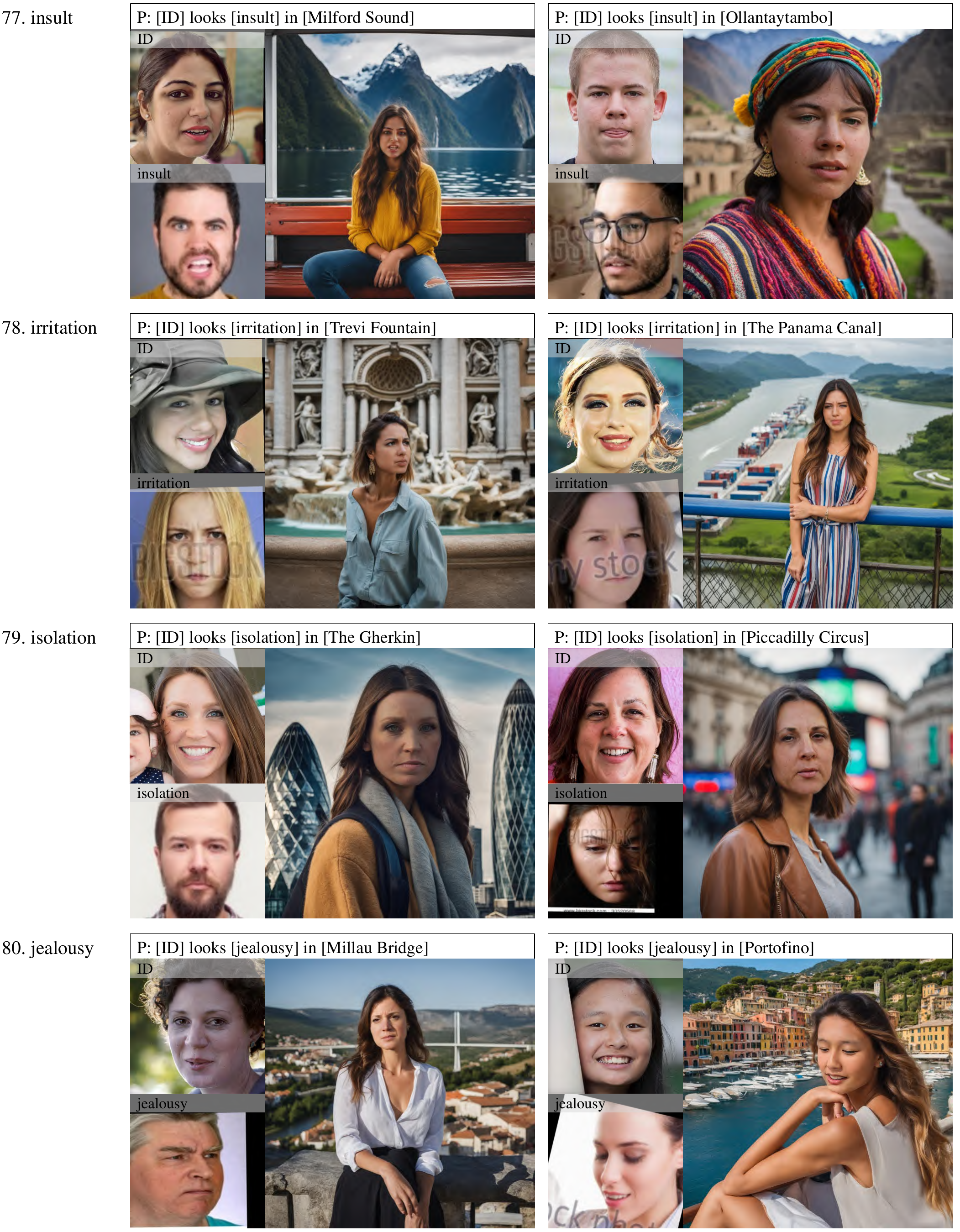} 
\vspace{-10pt}
\caption{Continues from Figures \ref{fig:p1}-\ref{fig:p19}. The input text prompt is shown at the top. The image in the top right corner refers to the ID image and the image in the bottom right corner refers to the expression reference image. The image on the right showcases the resulting image according to the inputs of the text prompt and ID image. Please zoom in for more details.}
\label{fig:p20}
\end{figure*}

\begin{figure*}
\centering
\includegraphics[width=0.95\textwidth]{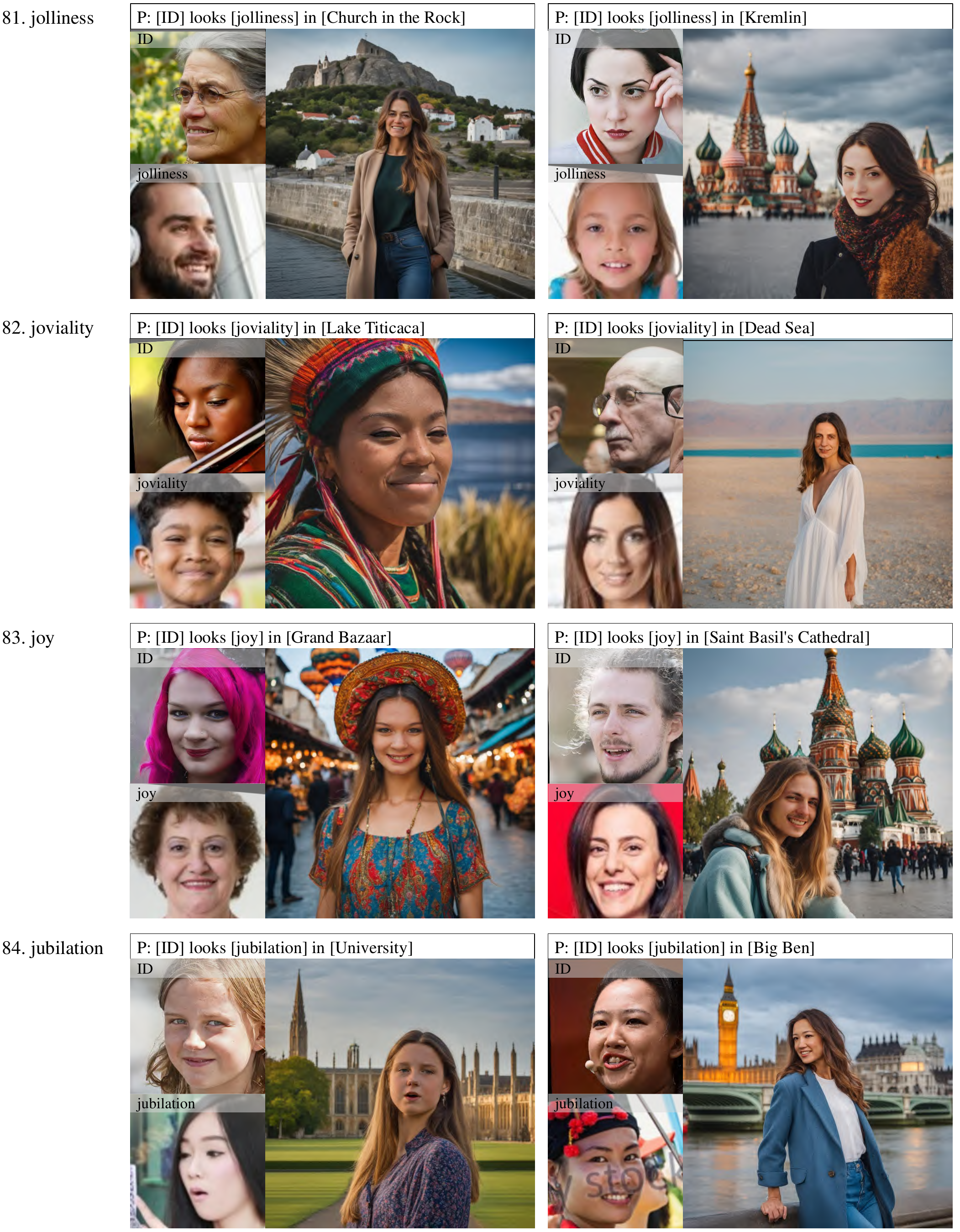} 
\vspace{-10pt}
\caption{Continues from Figures \ref{fig:p1}-\ref{fig:p20}. The input text prompt is shown at the top. The image in the top right corner refers to the ID image and the image in the bottom right corner refers to the expression reference image. The image on the right showcases the resulting image according to the inputs of the text prompt and ID image. Please zoom in for more details.}
\label{fig:p21}
\end{figure*}

\begin{figure*}
\centering
\includegraphics[width=0.95\textwidth]{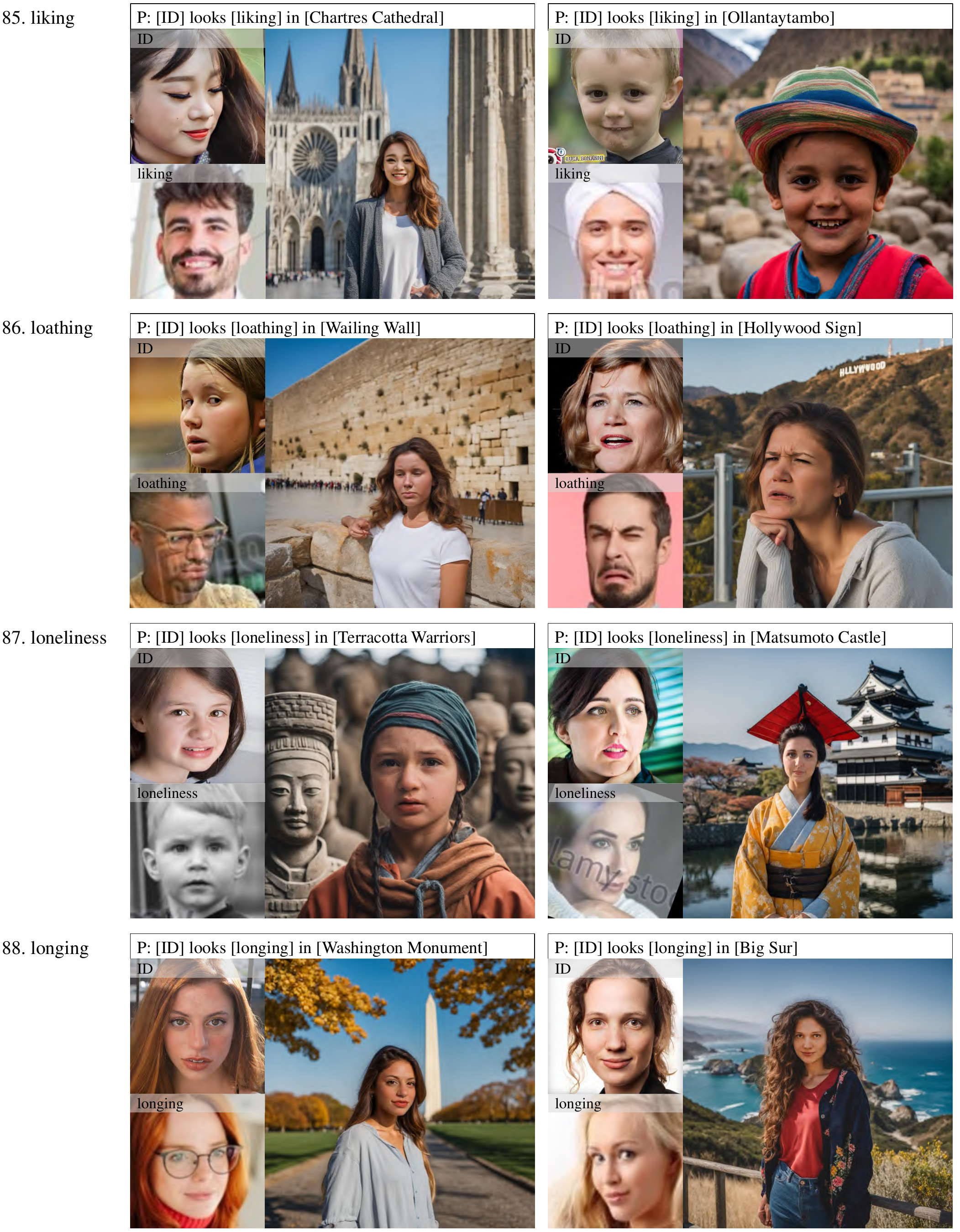} 
\vspace{-10pt}
\caption{Continues from Figures \ref{fig:p1}-\ref{fig:p21}. The input text prompt is shown at the top. The image in the top right corner refers to the ID image and the image in the bottom right corner refers to the expression reference image. The image on the right showcases the resulting image according to the inputs of the text prompt and ID image. Please zoom in for more details.}
\label{fig:p22}
\end{figure*}

\begin{figure*}
\centering
\includegraphics[width=0.95\textwidth]{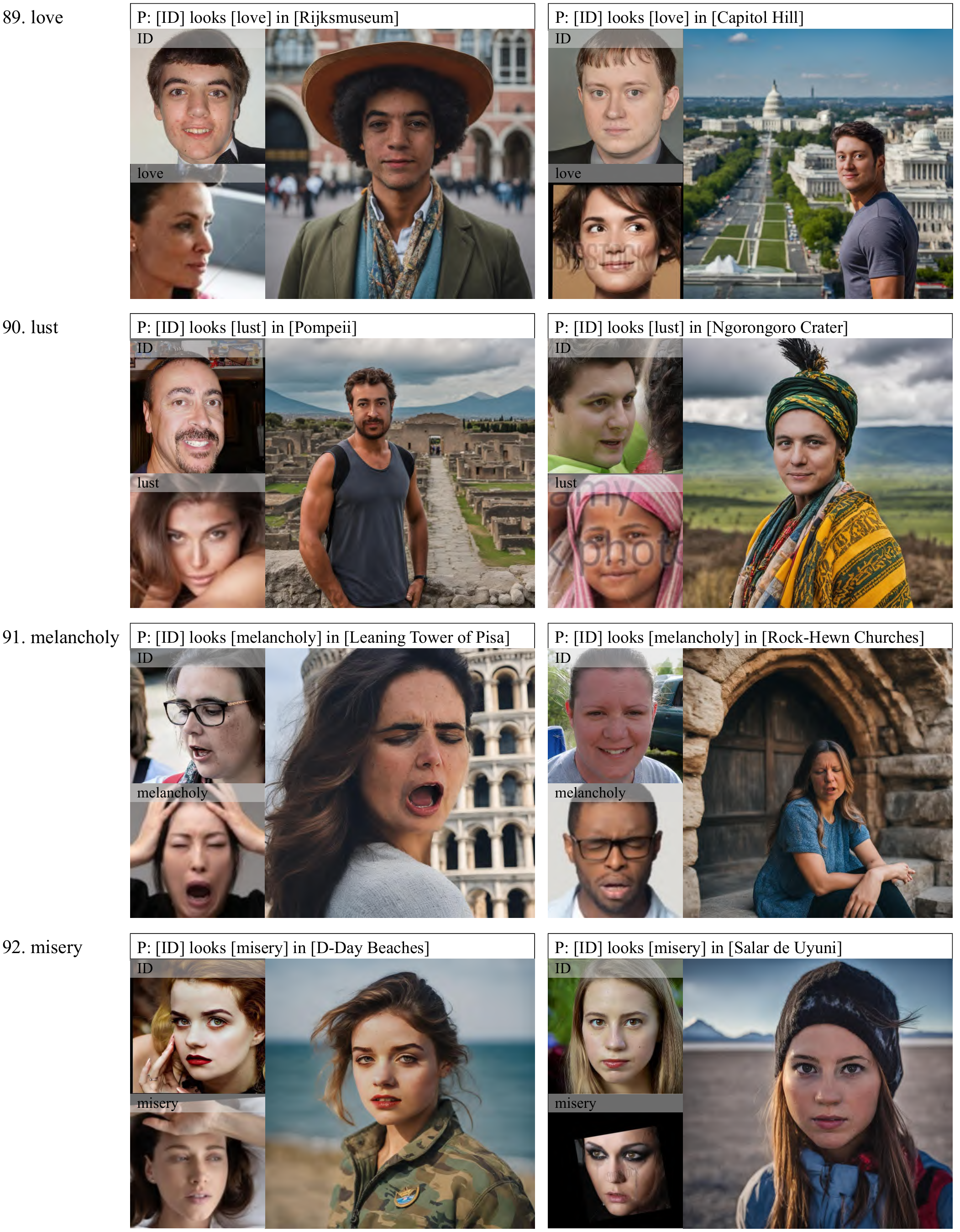} 
\vspace{-10pt}
\caption{Continues from Figures \ref{fig:p1}-\ref{fig:p22}. The input text prompt is shown at the top. The image in the top right corner refers to the ID image and the image in the bottom right corner refers to the expression reference image. The image on the right showcases the resulting image according to the inputs of the text prompt and ID image. Please zoom in for more details.}
\label{fig:p23}
\end{figure*}

\begin{figure*}
\centering
\includegraphics[width=0.95\textwidth]{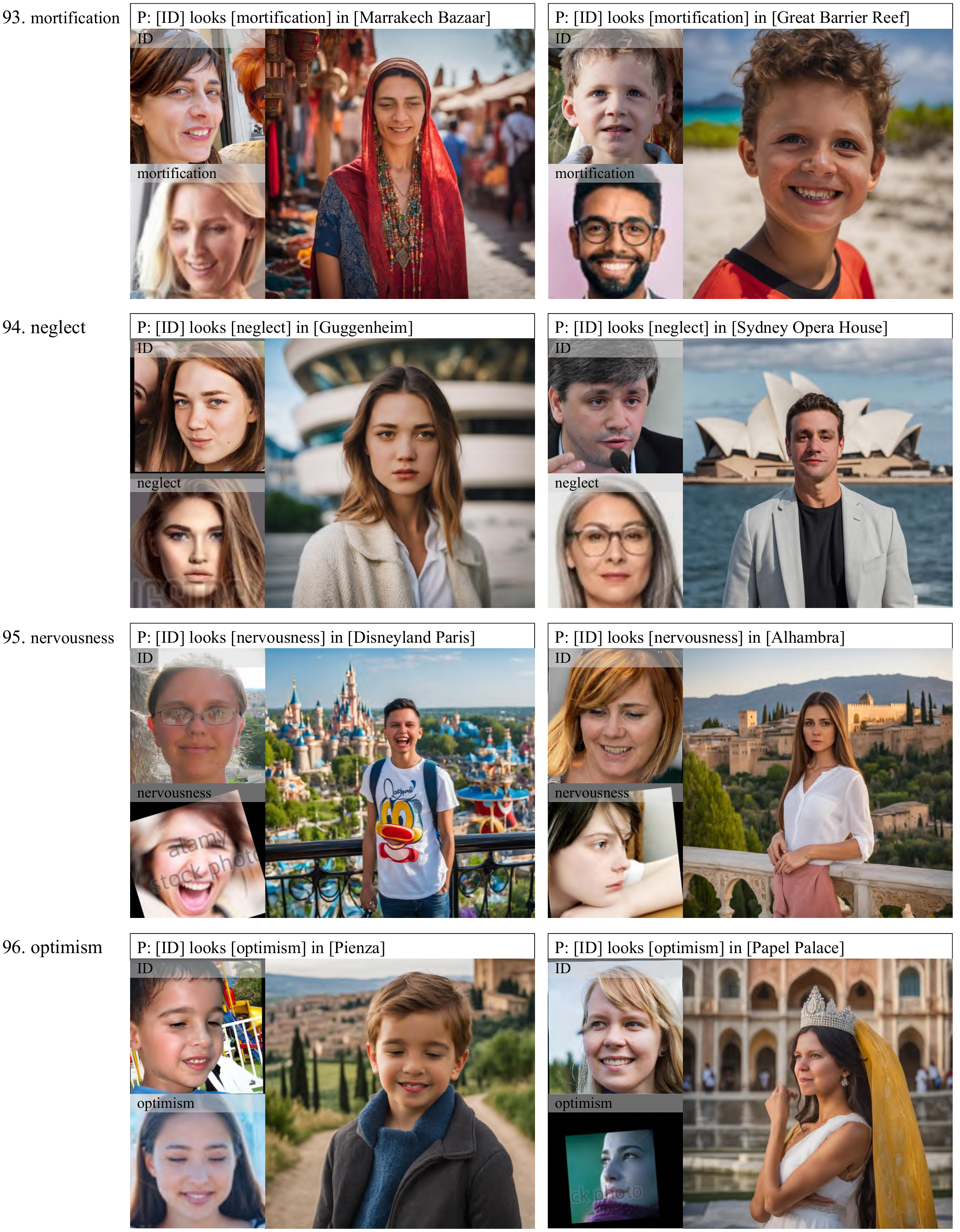} 
\vspace{-10pt}
\caption{Continues from Figures \ref{fig:p1}-\ref{fig:p23}. The input text prompt is shown at the top. The image in the top right corner refers to the ID image and the image in the bottom right corner refers to the expression reference image. The image on the right showcases the resulting image according to the inputs of the text prompt and ID image. Please zoom in for more details.}
\label{fig:p24}
\end{figure*}

\begin{figure*}
\centering
\includegraphics[width=0.95\textwidth]{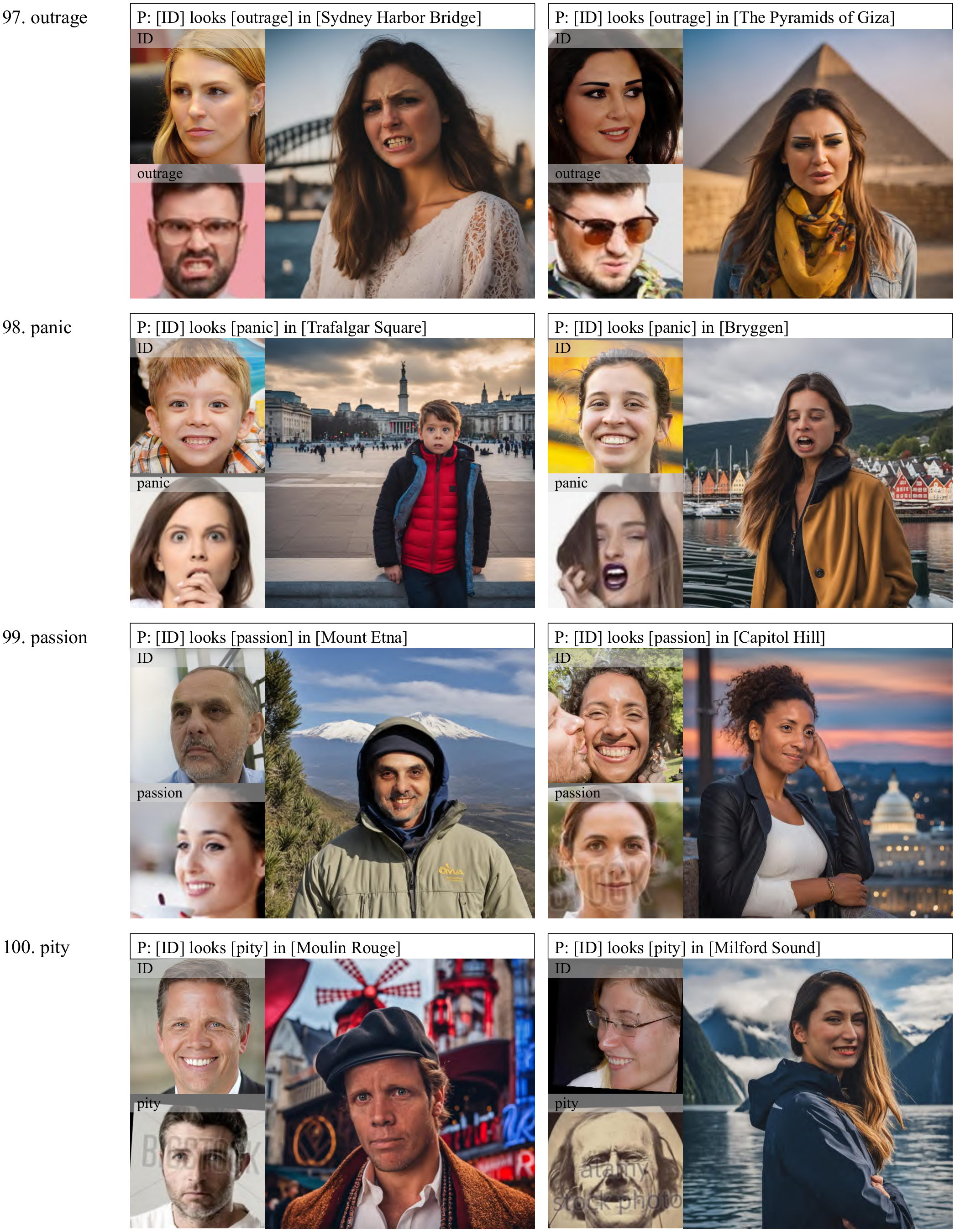} 
\vspace{-10pt}
\caption{Continues from Figures \ref{fig:p1}-\ref{fig:p11}. The input text prompt is shown at the top. The image in the top right corner refers to the ID image and the image in the bottom right corner refers to the expression reference image. The image on the right showcases the resulting image according to the inputs of the text prompt and ID image. Please zoom in for more details.}
\label{fig:p25}
\end{figure*}

\begin{figure*}
\centering
\includegraphics[width=0.95\textwidth]{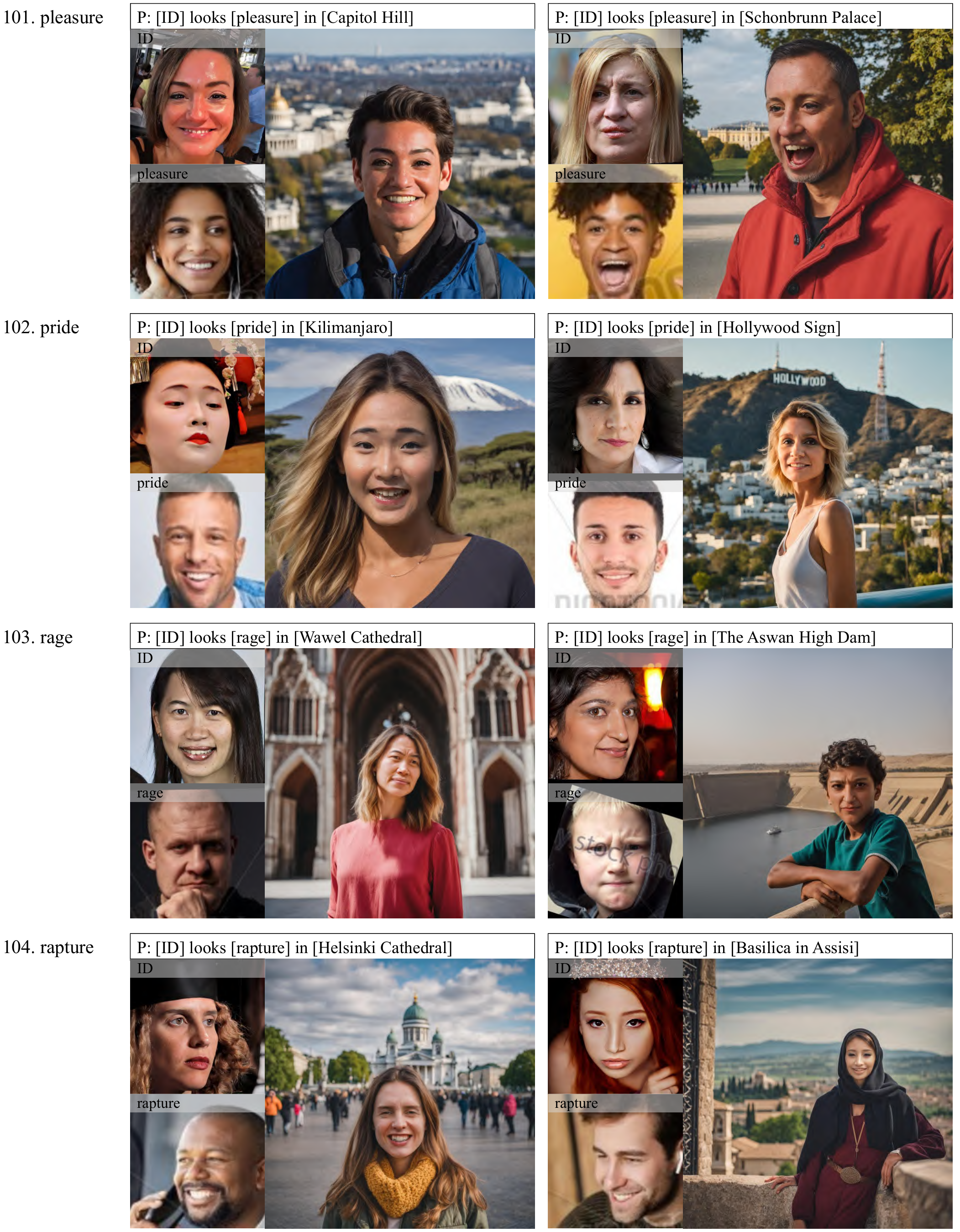} 
\vspace{-10pt}
\caption{Continues from Figures \ref{fig:p1}-\ref{fig:p25}. The input text prompt is shown at the top. The image in the top right corner refers to the ID image and the image in the bottom right corner refers to the expression reference image. The image on the right showcases the resulting image according to the inputs of the text prompt and ID image. Please zoom in for more details.}
\label{fig:p26}
\end{figure*}

\begin{figure*}
\centering
\includegraphics[width=0.95\textwidth]{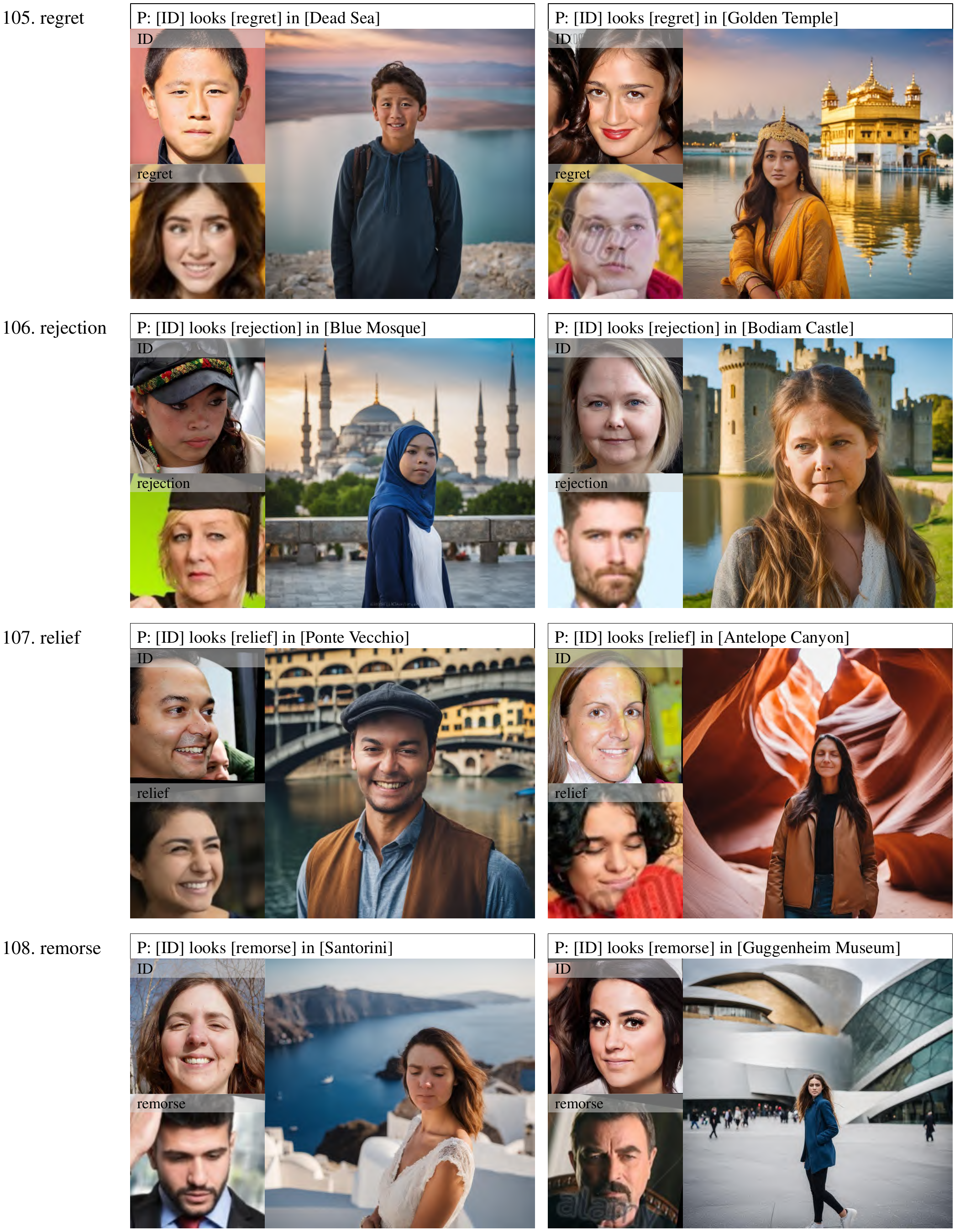} 
\vspace{-10pt}
\caption{Continues from Figures \ref{fig:p1}-\ref{fig:p26}. The input text prompt is shown at the top. The image in the top right corner refers to the ID image and the image in the bottom right corner refers to the expression reference image. The image on the right showcases the resulting image according to the inputs of the text prompt and ID image. Please zoom in for more details.}
\label{fig:p27}
\end{figure*}

\begin{figure*}
\centering
\includegraphics[width=0.95\textwidth]{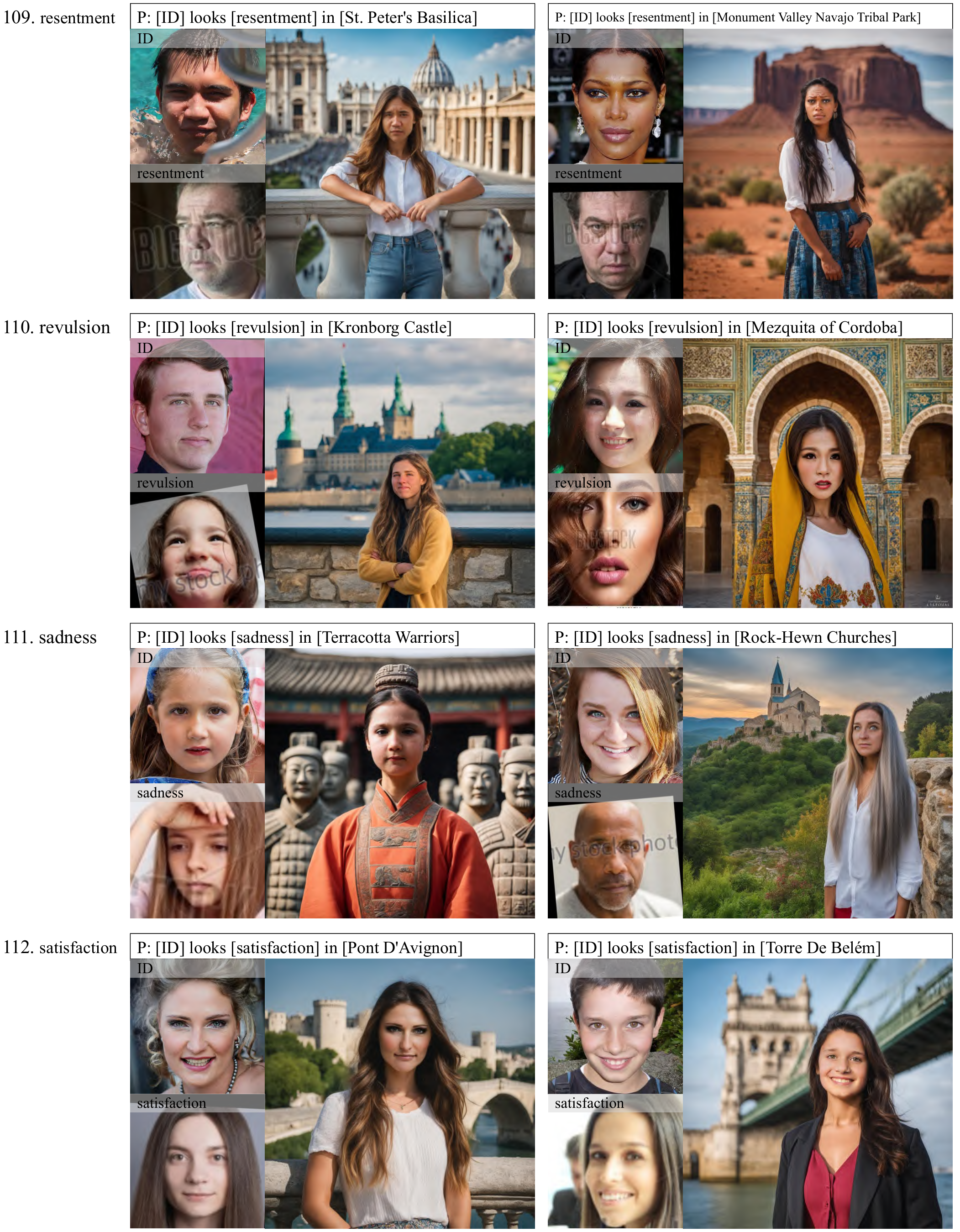} 
\vspace{-10pt}
\caption{Continues from Figures \ref{fig:p1}-\ref{fig:p27}. The input text prompt is shown at the top. The image in the top right corner refers to the ID image and the image in the bottom right corner refers to the expression reference image. The image on the right showcases the resulting image according to the inputs of the text prompt and ID image. Please zoom in for more details.}
\label{fig:p28}
\end{figure*}

\begin{figure*}
\centering
\includegraphics[width=0.95\textwidth]{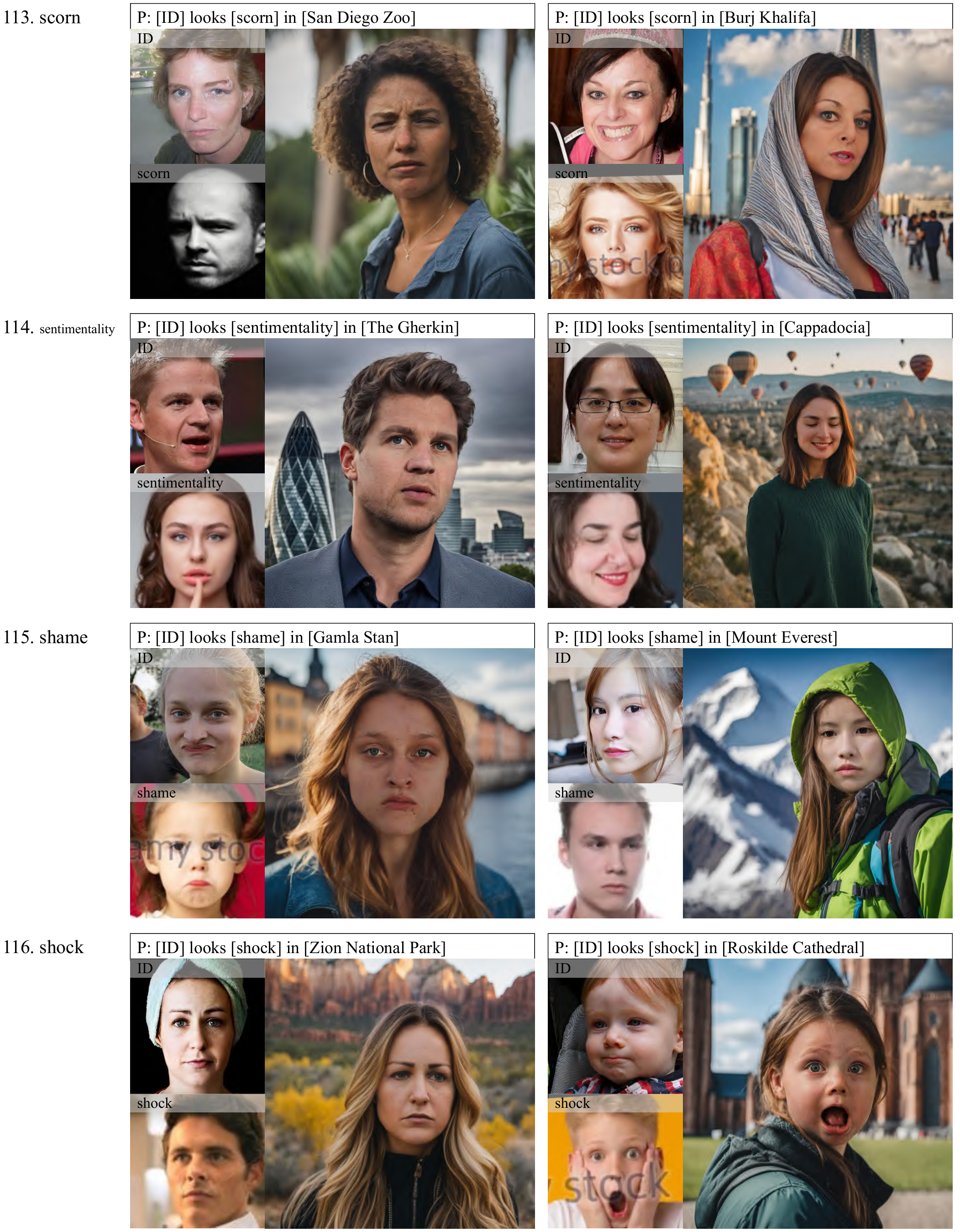} 
\vspace{-10pt}
\caption{Continues from Figures \ref{fig:p1}-\ref{fig:p28}. The input text prompt is shown at the top. The image in the top right corner refers to the ID image and the image in the bottom right corner refers to the expression reference image. The image on the right showcases the resulting image according to the inputs of the text prompt and ID image. Please zoom in for more details.}
\label{fig:p29}
\end{figure*}

\begin{figure*}
\centering
\includegraphics[width=0.95\textwidth]{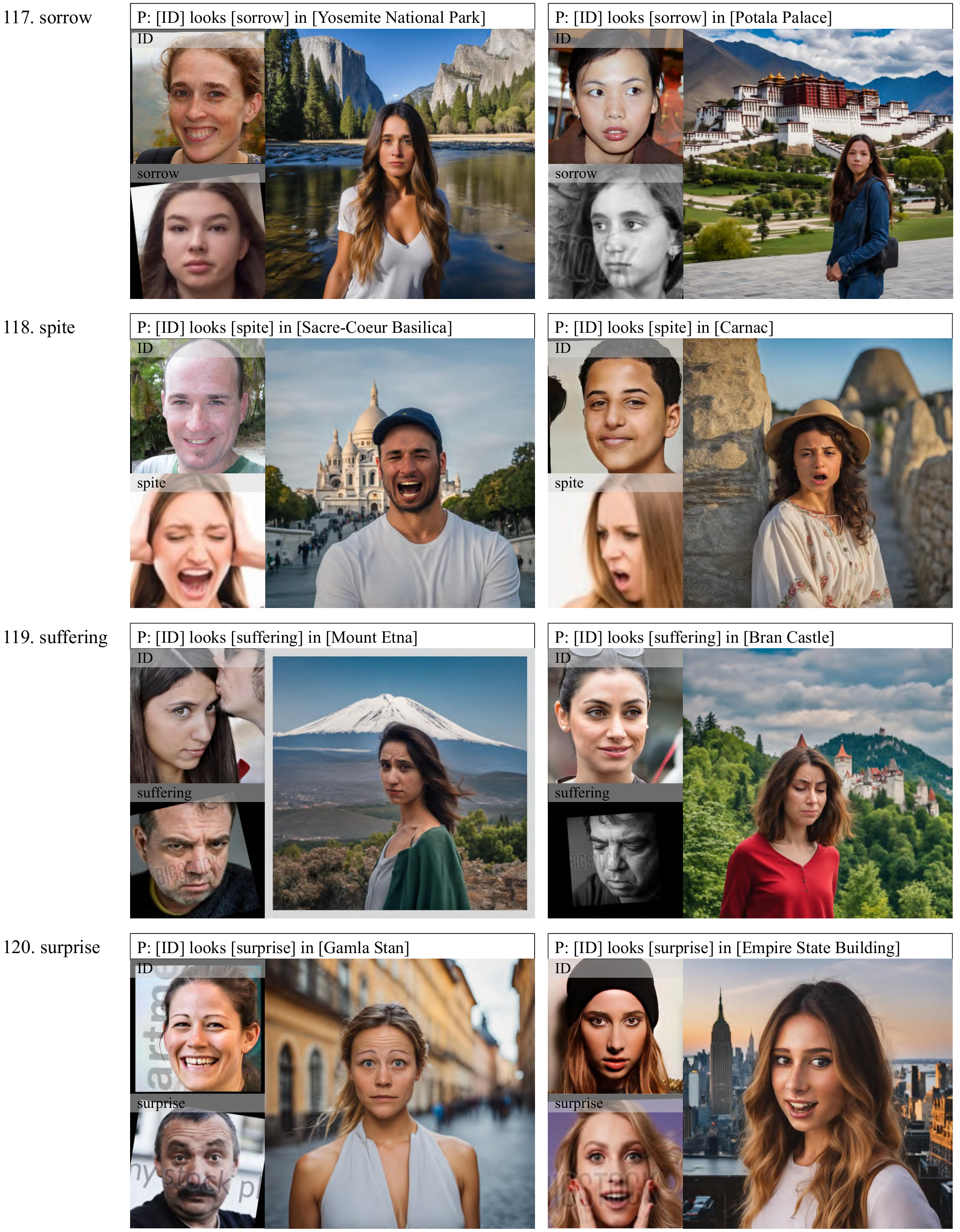} 
\vspace{-10pt}
\caption{Continues from Figures \ref{fig:p1}-\ref{fig:p29}. The input text prompt is shown at the top. The image in the top right corner refers to the ID image and the image in the bottom right corner refers to the expression reference image. The image on the right showcases the resulting image according to the inputs of the text prompt and ID image. Please zoom in for more details.}
\label{fig:p30}
\end{figure*}

\begin{figure*}
\centering
\includegraphics[width=0.95\textwidth]{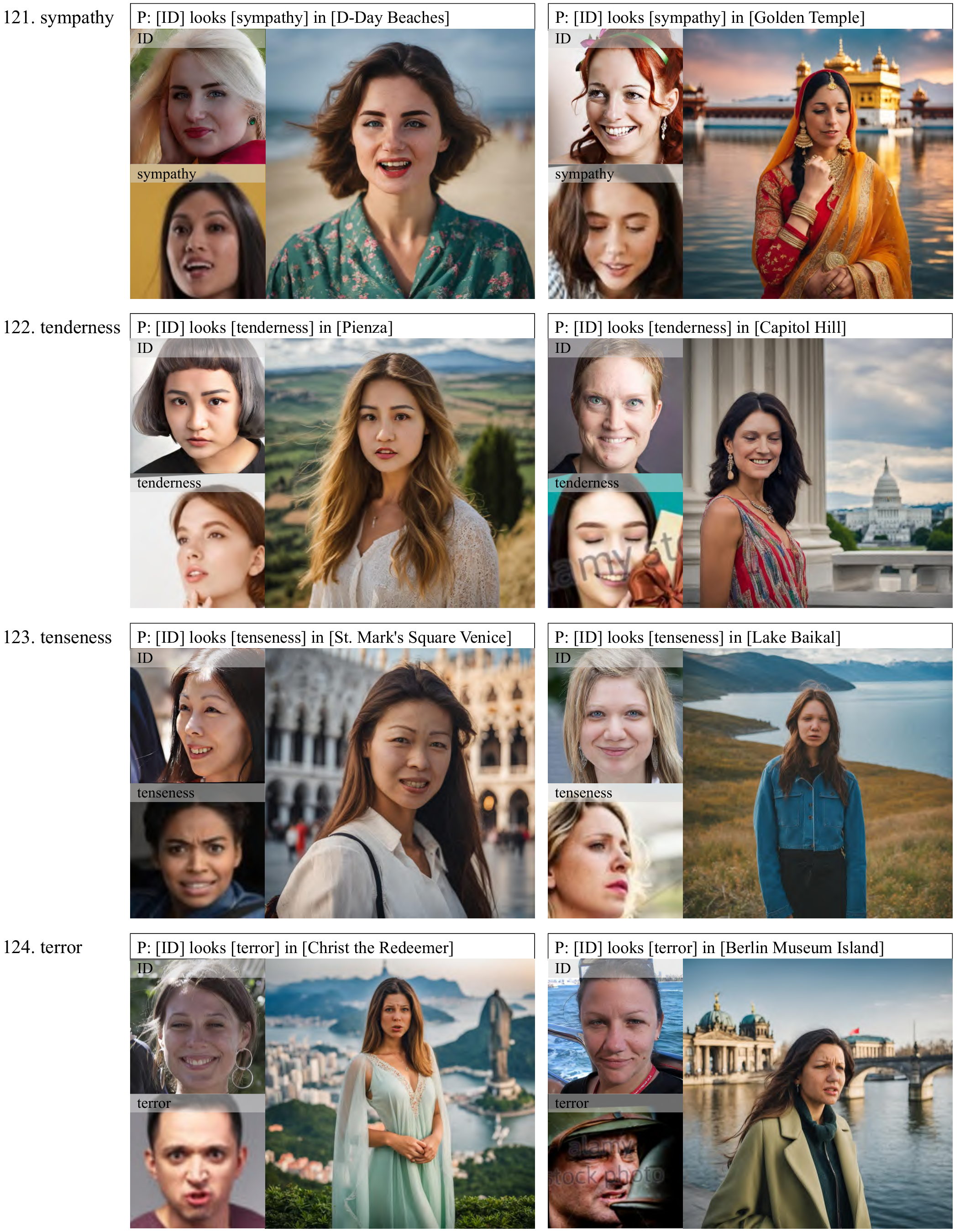} 
\vspace{-10pt}
\caption{Continues from Figures \ref{fig:p1}-\ref{fig:p30}. The input text prompt is shown at the top. The image in the top right corner refers to the ID image and the image in the bottom right corner refers to the expression reference image. The image on the right showcases the resulting image according to the inputs of the text prompt and ID image. Please zoom in for more details.}
\label{fig:p31}
\end{figure*}

\begin{figure*}
\centering
\includegraphics[width=0.95\textwidth]{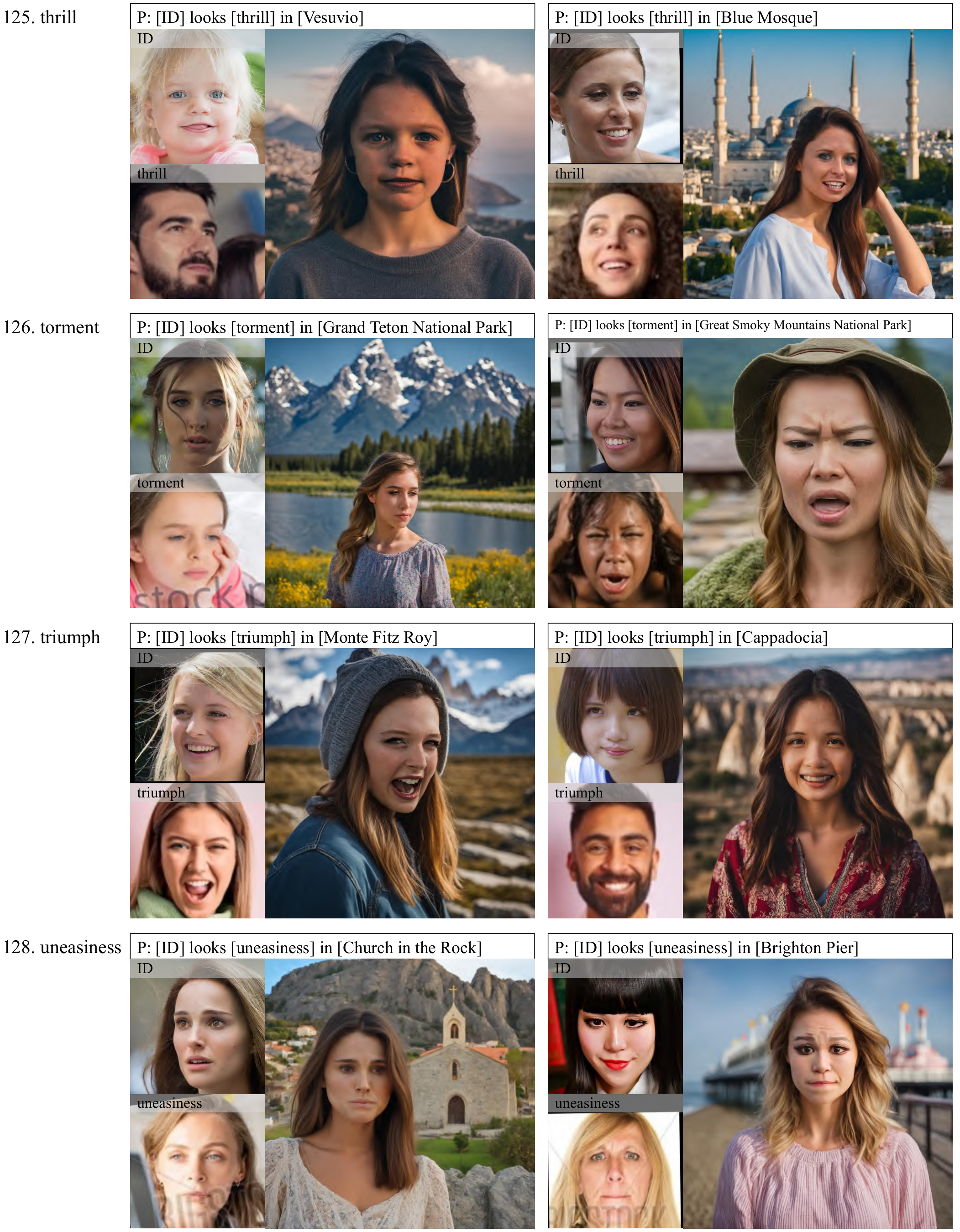} 
\vspace{-10pt}
\caption{Continues from Figures \ref{fig:p1}-\ref{fig:p31}. The input text prompt is shown at the top. The image in the top right corner refers to the ID image and the image in the bottom right corner refers to the expression reference image. The image on the right showcases the resulting image according to the inputs of the text prompt and ID image. Please zoom in for more details.}
\label{fig:p32}
\end{figure*}

\begin{figure*}
\centering
\includegraphics[width=0.95\textwidth]{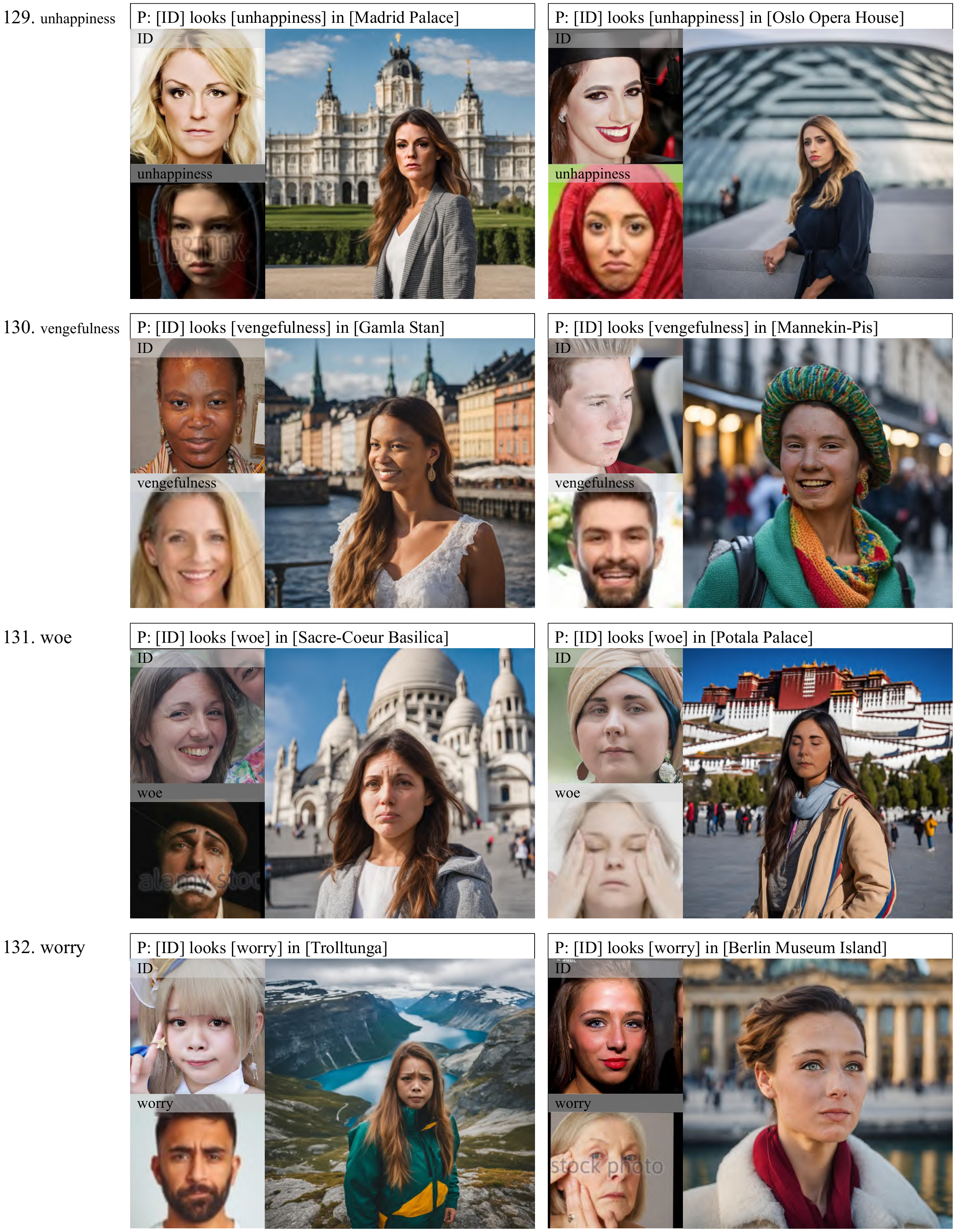} 
\vspace{-10pt}
\caption{Continues from Figures \ref{fig:p1}-\ref{fig:p32}. The input text prompt is shown at the top. The image in the top right corner refers to the ID image and the image in the bottom right corner refers to the expression reference image. The image on the right showcases the resulting image according to the inputs of the text prompt and ID image. Please zoom in for more details.}
\label{fig:p33}
\end{figure*}

\begin{figure*}
\centering
\includegraphics[width=0.95\textwidth]{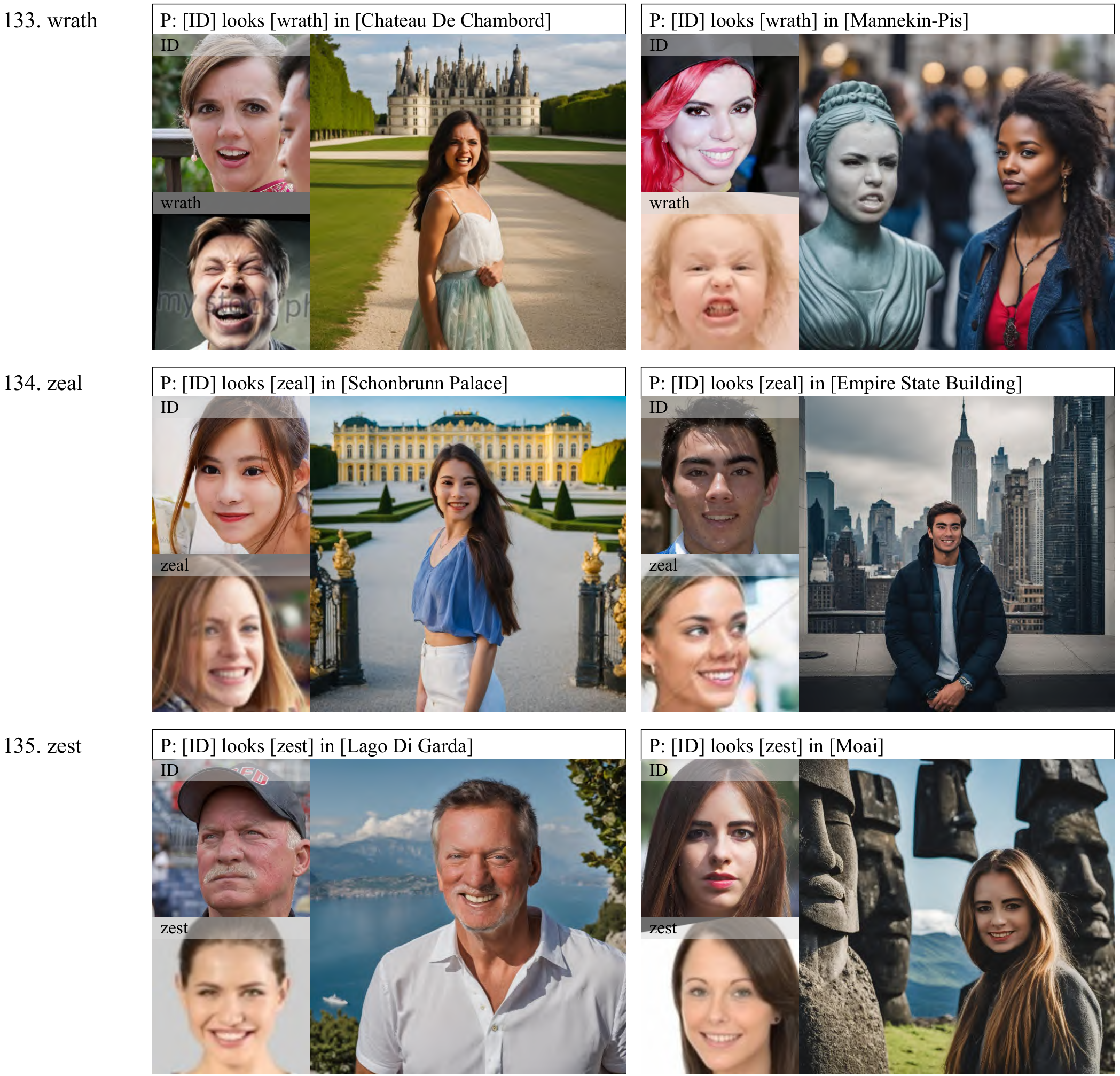} 
\vspace{-10pt}
\caption{Continues from Figures \ref{fig:p1}-\ref{fig:p33}. The input text prompt is shown at the top. The image in the top right corner refers to the ID image and the image in the bottom right corner refers to the expression reference image. The image on the right showcases the resulting image according to the inputs of the text prompt and ID image. Please zoom in for more details.}
\label{fig:p34}
\end{figure*}

Fig. \ref{fig:p1}-\ref{fig:p34} shows all of the fine-grained expression synthesis results with 135 labels of expression text \cite{EmoFace135}. Readers can zoom in for more details. However, these results are highly compressed due to the size limitations of the submitted files so some areas are distorted. Please refer to our project homepage for the original results: \url{https://diffsfsr.github.io/}

{
    \small
    \bibliographystyle{ieeenat_fullname}
    \bibliography{main}
}


\end{document}